\documentclass{elsarticle}
\usepackage{lineno,hyperref}

\usepackage[cmex10]{amsmath}
\usepackage{array}
\usepackage{graphicx}
\usepackage{array}
\usepackage{eqparbox}
\usepackage{url}
\usepackage{gensymb}
\usepackage{rotating}
\usepackage[usenames]{color}
\usepackage{colortbl}
\usepackage{dcolumn}
\usepackage{xspace}
\usepackage{algpseudocode}
\usepackage{algorithm}
\usepackage{tabu}
\usepackage{setspace}
\usepackage{appendix}
\usepackage{changepage}
\usepackage{natbib}
\usepackage{afterpage}
\usepackage{calc}
\usepackage[paperheight=\textheight+15ex,paperwidth=\textwidth+10ex,
 top=5ex, bottom=10ex, left=5ex, right=5ex]{geometry}

\makeatletter
\newcommand*{\@rowstyle}{}
\newcommand*{\rowstyle}[1]{
  \gdef\@rowstyle{#1}%
  \@rowstyle\ignorespaces%
}
\newcolumntype{=}{
  >{\gdef\@rowstyle{}}%
}
\newcolumntype{+}{
  >{\@rowstyle}%
}
\newcolumntype{B}[3]{>{\@rowstyle\DC@{#1}{#2}{#3}}c<{\DC@end}}
\newcommand{\algrule}[1][.2pt]{\par\vskip.5\baselineskip\hrule height #1\par\vskip.5\baselineskip}
\newcommand{\sqbox}[1]{%
    \@begin@tempboxa\hbox{#1}%
    \@tempdima=\dimexpr\width-\totalheight\relax
    \ifdim\@tempdima<\z@
        \fbox{\hbox{\hspace{-.5\@tempdima}\usebox\@tempboxa\hspace{-.5\@tempdima}}}%
    \else
        \ht\@tempboxa=\dimexpr\ht\@tempboxa+.5\@tempdima\relax
        \dp\@tempboxa=\dimexpr\dp\@tempboxa+.5\@tempdima\relax
        \fbox{\usebox\@tempboxa}%
    \fi
    \@end@tempboxa
}
\makeatother

\newcolumntype{.}{B{.}{.}{1}}
\newcolumntype{,}{D{.}{.}{1}}

\journal{Computer Vision and Image Understanding}

\begin{document}

\begin{frontmatter}

\title{MODS: Fast and Robust Method for Two-View Matching}
\author{Dmytro Mishkin, Jiri Matas, Michal Perdoch}
\address{Center for Machine Perception, Faculty of Electrical Engineering,\\ 
Czech Technical University in Prague.  Karlovo namesti, 13. Prague 2, 12135\\
}





\begin{abstract}

A novel algorithm for wide-baseline matching called MODS --  Matching On Demand with view Synthesis -- is presented. The MODS algorithm is experimentally shown to solve a broader range of wide-baseline problems than the state of the art while being nearly as fast as standard matchers on simple problems. The apparent robustness vs. speed trade-off is finessed by the use of progressively more time-consuming feature detectors and by on-demand generation of synthesized images that is performed until a reliable estimate of geometry is obtained.

We introduce an improved method for tentative correspondence selection, applicable both with and without view synthesis. A modification of the standard first to second nearest distance rule  increases the number of correct matches by 5-20\% at no additional computational cost.

Performance of the MODS algorithm is evaluated on several standard publicly available datasets, and on a new set of geometrically challenging wide baseline problems that is made public together with the ground truth. Experiments show that the MODS outperforms the state-of-the-art in robustness and speed. Moreover, MODS performs well on other classes of difficult two-view problems like matching of images from different modalities, with wide temporal baseline or with  significant lighting changes.

\end{abstract}

\end{frontmatter}


\section{Introduction}
\label{sec:introduction}
The wide baseline stereo \cite{Pritchett1998} problem -- the automatic estimation of a geometric transformation and
the selection of consistent correspondences between view pairs separated by a wide baseline -- has received significant attention in the last 15 years \cite{Mikolajczyk2005,Tuytelaars2008}. 
State-of-art local feature detectors \cite{Matas2002}, \cite{Mikolajczyk2004}, \cite{Lowe2004}, \cite{Bay2006} and descriptors \cite{Lowe2004}, \cite{Bay2006}, \cite{Belongie2002} allow to match images of a scene with a viewing angle difference up to $60^\circ$ for planar objects \cite{Mikolajczyk2005a} and $30^\circ$ for non-planar 3D objects \cite{Moreels2007}. 
Fast detectors \cite{Rosten2006}, \cite{Mair2010} and binary descriptors \cite{Rublee2011}, \cite{Leutenegger2011}, \cite{Alahi2012}  make matching significantly faster at the cost of decreasing tolerance to scale, rotation and affine changes. At the other end of the spectrum of wide baseline problems, the ASIFT matching scheme \cite{Morel2009,GuoshenYu2011}, increased the range of handled viewing angle differences up to the $80^\circ$ at the cost of a significant slow-down.

We propose a novel two-view matching algorithm called MODS -- matching with on-demand view synthesis -- that handles viewing angle difference even larger than the state-of-the-art ASIFT algorithm, without a significant increase of computational costs over  ``standard'' wide and narrow baseline approaches. 
The performance gain is achieved by introducing  a number of  improvements to the wide-baseline matching process.

First,  MODS employs a combination of different detectors. It is known that different detectors are suitable for different types of images \cite{Mikolajczyk2005a} and that some detectors are complementary in the type of structures in the image they respond to \cite{Aanaes2012}. Moreover, we show that the combination of the different detectors allows increasing the average speed of the matching and to match pairs of images which can not be solved by any of the detectors alone. The results indicate that searching for the ``best'' detector leads to  data- and 
problem-specific outcomes and that exploiting multiple detectors is superior to any single one.
 
Second, we introduce an iterative scheme which follows  the ``do only as much as needed'' principle. Progressively more powerful yet slower detectors and descriptors are applied, together with more images synthesized on-demand, until
sufficient support for a two-view geometry estimate is obtained.  Such on demand approach finesses an apparent robustness vs. speed trade-off, avoiding the slowdown for easy wide baseline problems  brought by the time-consuming operations needed for solving the most challenging pairs.

Third, a novel tentative correspondences generation strategy is presented which generalizes the standard first to second closest distance ratio \cite{Lowe2004}.
The selection strategy which shows performance superior to the standard method is applicable to any vector descriptor like SIFT~\cite{Lowe2004}, LIOP~\cite{Wang2011} and MROGH~\cite{Fan2012}.
  
The parameters of the MODS algorithm were optimized and its performance thoroughly evaluated. The optimization included the selection of the particular sequence of feature detectors, the choice of the number and parameters of images synthesized to facilitate matching and the parameter setting of the individual detectors.

The performance of the MODS algorithm was validated on several publicly available datasets and it was compared to the state-of-the-art in both speed and robustness, i.e. the ability to recover the two-view geometry reliably. We show that MODS significantly outperforms prior approaches in both robustness and speed.
We have collected a set of image pairs for evaluating MODS on wide baseline problems with very large angular difference between views. These form the Extreme View Dataset. The dataset with the ground truth  and the source code of the MODS algorithm is available on the authors web-page\footnote{Available at http://cmp.felk.cvut.cz/wbs/index.html}.

\section{Related work}
The standard wide baseline matching pipeline (see Alg.~\ref{alg:standard}) begins with the detection of local features, computation of descriptors, generation of tentative correspondences and ends with geometric verification using the homography or epipolar constraint.  Matching images of a scene with viewpoint difference up to $60^\circ$ for planar objects \cite{Mikolajczyk2005} and $30^\circ$ for non-planar 3D objects \cite{Moreels2007} was reported. The execution time varies from a fraction of a second to seconds for 800x600 images~\cite{Mikolajczyk2005a}.

\begin{algorithm}[tbp]
\small
\caption{{\bf The standard two view matching scheme}}
\label{alg:standard}
\begin{algorithmic}
\Require{$I_1$, $I_2$ -- two images.}
\Ensure{Fundamental or homography matrix F or H respectively;\\ a list of corresponding features.}\\
\algrule
\For {$I_1$ and $I_2$ independently}
\State \textbf{1.1 Detect and describe local features}.
\EndFor
\State \textbf{2 Generate tentative correspondences} for 
$I_1$ and $I_2$ using the 2nd closest ratio. 
\State \textbf{3 Geometrically verify tentative correspondences} using RANSAC\\ 
\hskip1em while estimating H or F.
\end{algorithmic}
\end{algorithm}

The idea of generating synthetic views to improve a local feature based wide baseline  matching pipeline was first explored by Lepetit and Fua~\cite{Lepetit2006}. They synthesized views to find distinctive keypoints repeatedly detectable under affine deformations. Synthetic views provided a training set for learning a random forest classifier that labeled individual feature points. Feature points in different images with the same label were assumed to be in correspondence. The simple keypoint detector of Lepetit and Fua is very fast, but invariant only to translation and rotation and thus the number of views necessary to achieve acceptable repeatability was high. The method was tested on pairs undergoing significant affine transformations, but the final representation did not scale and can not be easily used for indexing.

\begin{algorithm}[tbp]
\small
\caption{{\bf ASIFT}}
\label{alg:ASIFT}
\begin{algorithmic}
\Require{$I_1$, $I_2$ -- two images.}
\Ensure{List of corresponding points; fundamental matrix F.}\\
\algrule
\For {$I_1$ and $I_2$ independently}
\State \textbf{1.1 Generate synthetic views} according to the tilt-rotation-detector setup.
\State \textbf{1.2 Detect and describe local features}.
\EndFor
\State \textbf{2 Generate tentative correspondences} for each pair of the synthesized views of \\
\hskip1em $I_1$ and $I_2$ independently using the 2nd closest ratio. 
\State \textbf{3 Add correspondences to the general list}. \\
\hskip1em Reproject corresponding features to original images. 
\State \textbf{4 Filter duplicate, ``one-to-many'' and ``many-to-one'' matches}. 
\State \textbf{5 Geometrically verify tentative correspondences} using ORSA~\cite{Moisan2004}\\ 
\hskip1em while estimating F.
\end{algorithmic}
\end{algorithm}

Recently, Morel et.al.~\cite{Morel2009} proposed a new matching pipeline -- see Alg.~\ref{alg:ASIFT}. The authors showed that view synthesis extends the handled range of viewpoint differences.
The ASIFT algorithm starts by generating synthetic views (described in Section~\ref{mods:synthetic-view-generation}) for both images. Next, feature detection and description are performed using standard SIFT~\cite{Lowe2004} in each synthesized view. Tentative correspondences are formed for all pairs of views synthesized  from the first and second image. The matching stage thus entails $n^2$ independent matching problems, where $n$ is the number of synthesized views per image. The set of correspondences between the images is the union of results for all synthesized pairs.
The  \emph{duplicate filtering} stage of ASIFT prunes correspondences with small spatial distance (2~pixels) of local features in both images -- all such correspondences except one (random) are eliminated from the final correspondence set.  ``\emph{One-to-many}'' correspondences -- correspondences of features which are close to each other (are situated in radius of $\sqrt{2}$ pixels) in one image while spread in other synthetic views are also eliminated, despite the fact that some of the pairs can be correct.
Finally, geometric verification is performed by ORSA~\cite{Moisan2004}. ORSA is a RANSAC-based method, which exploits an a-contrario approach to detect incorrect epipolar geometries. Instead of having a constant error threshold, ORSA looks for matches that have the highest ``diameter'', i.e. matches which cover a large image area. ASIFT was shown to match images of a scene with viewpoint difference up to $80^\circ$ for planar objects \cite{Morel2009}. Computational costs are in the order of tens of seconds to a few minutes.

The latest extensions of wide-baseline matching pipeline are limited to modifications of the ASIFT algorithm. Liu et.al.~\cite{Liu2012} synthesized perspective warps rather than affine. Pang et.al ~\cite{Pang2012} replaced SIFT by SURF~\cite{Bay2006} in the ASIFT algorithm to reduce the computation time. Forssen and Lowe~\cite{Forssen2007} proposed to detect MSER on scale pyramid, which might be seen as scale synthesis.
\section{The MODS algorithm}
\label{sec:mods}
The main idea of the proposed iterative MODS algorithm (see Alg.~\ref{alg:MODS}) is to repeat a sequence of two-view matching procedures, until a required number of geometrically verified correspondences is found. In each iteration, a different and potentially complementary detector is used and a different set of views synthesized.
The algorithm starts with fast detectors with limited invariance proceeding progressively with more complex, robust, but computationally costly ones. MODS is thus capable of solving simple matching problem fast without loosing the ability to deal with
very difficult cases where a combination of detectors is employed to extend the state-of-the-art.

The adopted sequence of detectors and view synthesis  parameters is an outcome of extensive experimental search.  The objective was to solve the most challenging problems in  the development set, i.e. to correctly recover their two-view geometry,  while keeping the speed comparable to standard single-detector wide-baseline matchers for simple problems.  Details about the
selected configuration and the optimization process are given in Section~\ref{mods:parameters}. The rest of the section describes the steps involved in the iterations of the MODS algorithm, which is compared to the standard two view matching and ASIFT pipelines.

\begin{algorithm}[tb]
\small
\caption{{\bf MODS}}
\label{alg:MODS}
\begin{algorithmic}
\Require{$I_1$, $I_2$ -- two images; $\theta_m$ -- minimum required number of matches; \\$S_{\text{max}}$ -- maximum number of iterations. }
\Ensure{Fundamental or homography matrix F or H; list of corresponding points.}\\
\hskip-1em\textbf{Variables}: $N_{\text{matches}}$ -- detected correspondences, $\text{Iter}$ -- current iteration.
\algrule
\While {$(N_{\text{matches}} < \theta_{m})$ \textbf{and} $(\text{Iter} < S_{\text{max}})$} 
\For {$I_1$ and $I_2$ independently}
\State \textbf{1.1 Generate synthetic views} according to the \\
\hskip3em scale-tilt-rotation-detector-descriptor setup for the Iter (Tables~\ref{tab:MODS-configuration}, \ref{tab:view-synth-configuration}).
\State \textbf{1.2 Detect and describe local features}.
\State \textbf{1.3 Reproject local features} to original image. \\ 
\hskip3em Add described features to general list.
\EndFor
\State \textbf{2 Generate tentative correspondences} using the first geom. inconsistent rule. 
\State \textbf{3 Filter duplicate matches}. 
\State \textbf{4 Geometrically verify tentative correspondences} with DEGENSAC~\cite{Chum2005DEGENSAC}\\
\hskip2em while estimating F or H.
\State \textbf{5 Geometrically verify inliers} with local affine frame shape.
\EndWhile
\end{algorithmic}
\end{algorithm}
\subsection{Synthetic views generation} 
\label{mods:synthetic-view-generation}
\begin{figure}[tb]
\centering
\includegraphics[width=0.49\linewidth]{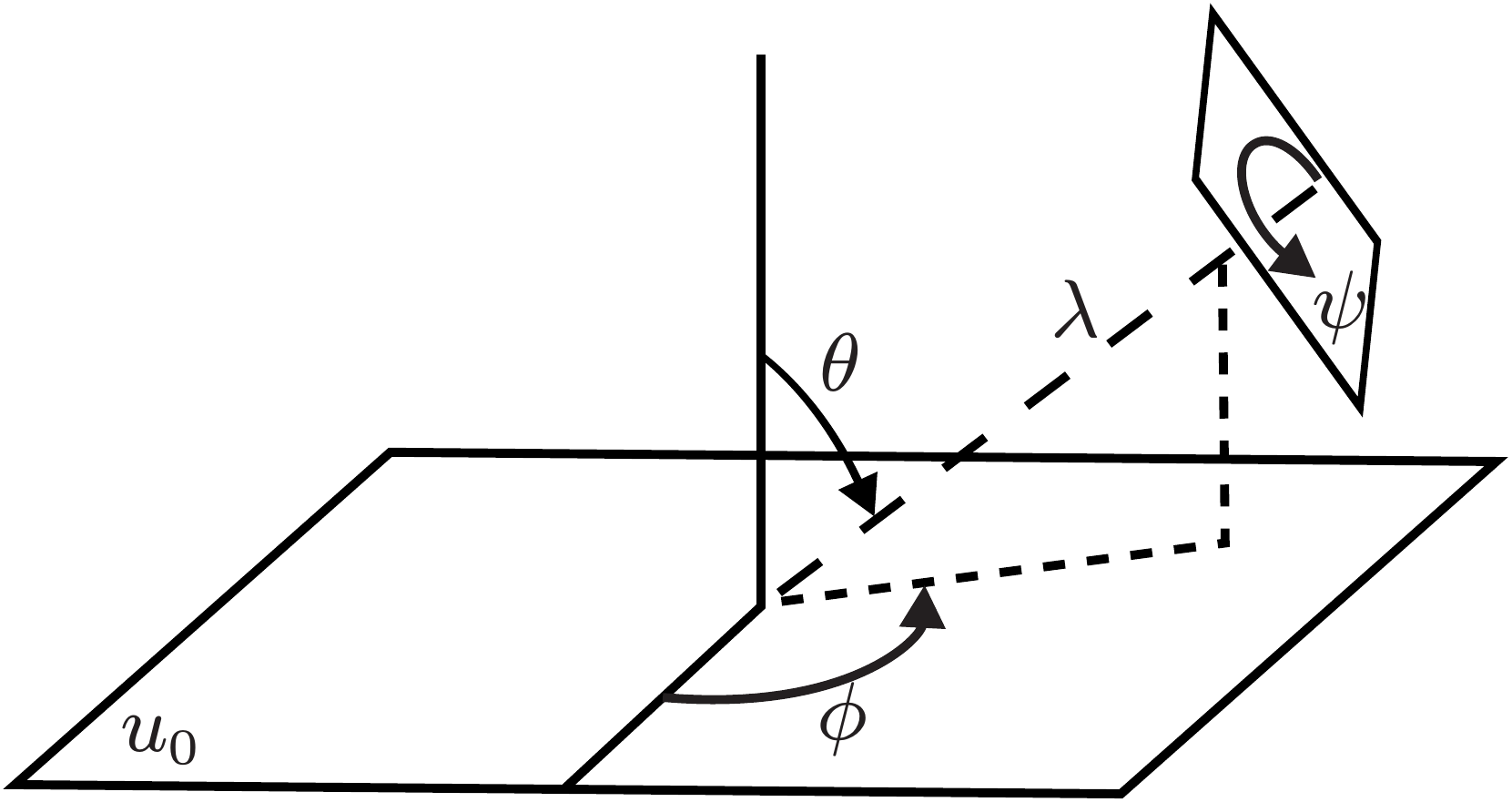}
\includegraphics[width=0.49\linewidth]{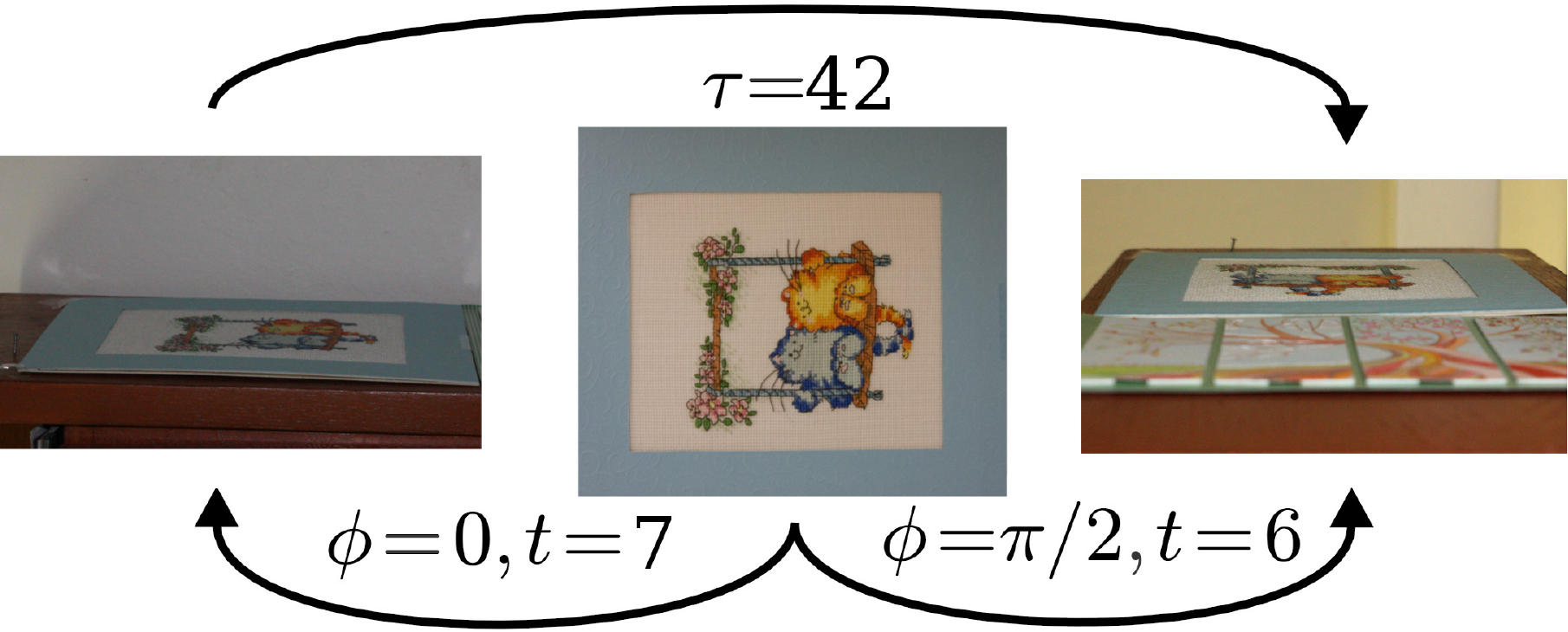}
\caption{(left) the affine camera model~(\ref{eq:affine-matrix-decomposition}). Latitude $\theta = \arccos{1/t}$ -- latitude, longitude  $\phi$, scale $\lambda$ scale. (right) Transitional tilt $\tau$ for absolute tilt $t$ and rotation $\phi$.}
\label{fig:affine-camera}
\end{figure}

MODS (Alg.~\ref{alg:MODS}) starts by synthetic view generation.  It is well known that a homography $H$ can be approximated by an affine transformation $A$ at a point using the first order Taylor expansion. The affine transformation can be uniquely decomposed by SVD into a rotation, skew, scale and rotation around the optical axis~\cite{Hartley2004}. In \cite{Morel2009}, the authors proposed to decompose the affine transformation $A$ as 
\begin{equation}
\begin{array}{rcl}
\label{eq:affine-matrix-decomposition}
\footnotesize
A&=&H_\lambda{}R_1(\psi)T_t R_2 (\phi) = \\[3pt]
&=&\lambda{}%
\left(
\footnotesize
\begin{array}{@{\hspace{0pt}}c@{\hspace{6pt}}c@{\hspace{0pt}}}
\cos{\psi} & -\sin{\psi} \\
\sin{\psi} & \cos{\psi}\\
\end{array}\right)\!\left(%
\footnotesize
\begin{array}{
@{\hspace{0pt}}c@{\hspace{6pt}}c@{\hspace{0pt}}}
t & 0  \\
0 & 1 \\        
\end{array}\right)\!\left(%
\footnotesize
\begin{array}{
@{\hspace{0pt}}c@{\hspace{6pt}}c@{\hspace{0pt}}}
\cos{\phi} & -\sin{\phi}\\
\sin{\phi} & \cos{\phi}\\
\end{array}\right)
\end{array}
\end{equation}
where $\lambda{} > 0$, $R_1$ and $R_2$ are rotations, and $T_t$ is a diagonal matrix with $t > 1$. Parameter $t$ is called the absolute tilt, $\phi \in \langle 0,\pi)$ is the optical axis longitude and $\psi \in \langle 0 ,2\pi)$ is the rotation of the camera around the optical axis (see Figure~\ref{fig:affine-camera}). Each synthesized view is thus parametrized by the tilt, longitude and optionally the scale and represents a sample of the view-sphere resp. view-volume around the original image.

The view synthesis proceeds in the following steps: at first, a scale synthesis is performed by building a Gaussian scale-space with Gaussian $\sigma={\sigma_{\text{base}}}\cdot{S}$ and downsampling factor $S$ ($S<1$). Then, each image in the scale-space is in-plane rotated by longitude $\phi$ with step $\Delta\phi={\Delta\phi_{\text{base}}}/{t}$. In the third step, all rotated images are convolved with a Gaussian filter with $\sigma = \sigma_{\text{base}}$ along the vertical  and  $\sigma = t\cdot{}\sigma_{\text{base}}$ along the horizontal direction to eliminate aliasing in a final tilting step. The final tilt is applied by shrinking the image along the horizontal direction by factor $t$. The synthesis parameters are: the set of scales \{S\}, $\Delta\phi_{\text{base}}$ -- the longitude sampling step at tilt $t=1$, the set of simulated tilts~\{t\}.
\subsection{Local feature detection and description} 
The second step of MODS is detection and description of local features.
%
%
It is known that different local feature detectors are suitable for different types of images~\cite{Mikolajczyk2005a} and that some detectors are complementary in the image structures they respond to~\cite{Aanaes2012}. Our experiments show (see Section~\ref{sec:main}) that combining detectors improves the overall robustness and speed of the matching procedure.

MODS combines a fast similarity covariant FAST (in ORB implementation) detector and affine covariant detectors MSER and Hessian-Affine. 
The normalized patches are described by the binary descriptor BRIEF~\cite{Rublee2011} (in ORB implementation) and a recent modification of SIFT~\cite{Lowe2004} -- the RootSIFT~\cite{Arandjelovic2012}. The local feature frames computed on the synthesized views are backprojected to the coordinate system of the original image by the known affine matrix $A$ and associated with the descriptor and the originating synthetic view. MODS steps configuration are specified in Table~\ref{tab:MODS-configuration}. 

For the MSER and Hessian-Affine detectors, the fast affine feature extraction process from~\cite{Lenc2014} was applied.

\begin{table}[tb]
\caption{MODS step configurations are defined by a detector, descriptor and the set of the synthesized views. RootSIFT is used for all detectors but ORB which is described by BRIEF.}
\label{tab:MODS-configuration}
\footnotesize 
\centering
\bgroup
\tabulinesep=0.4mm
\begin{tabu} spread 0pt{|X[-1m,c]|X[-1,m,l]|}
\hline
Iter. & \centering Setup\\                                                        \hline
1 & ORB,$\{S\} = \{1\}$, $\{t\} = \{1\}$, $\Delta\phi=360\degree{}/t$ \\          \hline
2 & ORB,$\{S\} = \{1\}$, $\{t\} = \{1;5;9\}$, $\Delta\phi=360\degree{}/t$ \\      \hline
3 & MSER,$\{S\} = \{1;0.25;0.125\}$, $\{t\} = \{1\}$, $\Delta\phi=360\degree{}/t$\\       \hline
4 & MSER,$\{S\} = \{1;0.25;0.125\}$, $\{t\} = \{1;3;6;9\}$, $\Delta\phi=360\degree{}/t$\\ \hline
5 & HessAff, $\{S\} = \{1\}$, $\{t\} = \{1;2;4;6;8\}$, $\Delta\phi=360\degree{}/t$\\      \hline
6 & HessAff, $\{S\} = \{1\}$, $\{t\} = \{1;2;4;6;8\}$, $\Delta\phi=120\degree{}/t$\\      \hline
7& HessAff , $\{S\} = \{1\}$, $\{t\} = \{1;2;4;6;8;10\}$, $\Delta\phi=60\degree{}/t$ \\
\hline
\end{tabu}
\egroup
\end{table}
\begin{table}[tb]
\caption{MODS variants tested. The corresponding steps are the same as in Table~\ref{tab:MODS-configuration}. The tarred MODS:$3^{*}$-$6^{*}$ configuration slightly differs from Table~\ref{tab:MODS-configuration} and corresponds to~\cite{Mishkin2013}.}
\label{tab:MODS-variants}
\footnotesize 
\centering
\bgroup
\tabulinesep=0.4mm
\begin{tabu}{|l|c|c|c|c|c|c|c|}
\hline
Name & \multicolumn{7}{c|}{Steps}\\
\hline
MODS == MODS:1-7 & 1 & 2 & 3 & 4 & 5 &6 &7\\
MODS:2-7 & - & 2 & 3 & 4 & 5 &6 &7\\
MODS:$3^{*}$-6& - & - & $3^{*}$ & $4^{*}$ & 5 &6 &-\\
MODS:3-7& - & - & 3 & 4 & 5 &6 &7\\
MODS:3,2-7& 3 & - & 2 & 4 & 5 &6 &7\\
MODS:2,4,7& - & 2 & - & 4 & - &- &7\\
MODS:5,1-7& 5 & 1 & 2 & 3 & 4 &6 &7\\
\hline
\end{tabu}
\egroup
\end{table}
\subsection{Tentative correspondence generation}
\begin{figure}[tb]
\centering
\includegraphics[width=1\linewidth]{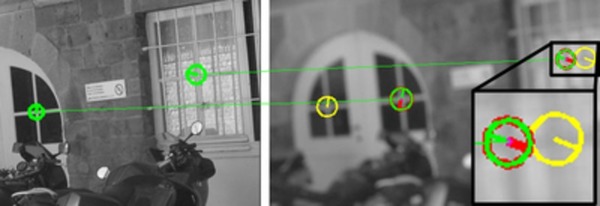}\\
\vspace{1 mm}
\includegraphics[width=1\linewidth]{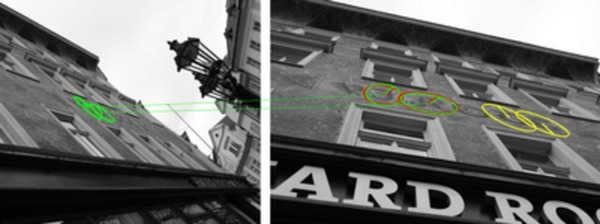}
\caption{Green regions -- correct correspondences rejected by the standard {first to second closest ratio} test (second closest region is in red), but recovered  by the {first to first geometrically inconsistent ratio} (first geometrically inconsistent region is in yellow) matching strategy. DoG regions (top), MSERs (bottom).}
\label{fig:1st-inconsistent-illustration}
\end{figure}

The next step of the MODS algorithm is the generation of tentative correspondences.
Different strategies for the computation of tentative correspondences in wide-baseline matching were proposed. The standard method for matching SIFT(-like) descriptors is based on ratio of the distances to the closest and the second closest descriptors in the other image~\cite{Lowe2004}.
While the performance of this test is in general very good, it degrades when multiple observations of the same feature are present. 
In this case, the presence of similar descriptors will lead to the first to second SIFT ratio to be close to 1 and the correspondences will "annihilate" each other, despite the fact they represent the same geometric constraints and are therefore not mutually contradictory (see Figure~\ref{fig:1st-inconsistent-illustration}). 
The problem of multiple detections is amplified in matching by view synthesis since covariantly detected local features are often repeatedly discovered in multiple synthetic views. 

To address this problem, we propose a modified matching strategy denoted \emph{first to first geometrically inconsistent -- FGINN} . Instead of comparing the first to the second closest descriptor distance, the distance of the first descriptor and the closest descriptor that is geometrically inconsistent with the first one is used. We call descriptors in one image {geometrically inconsistent} if the Euclidean distance between centers of the regions is $\geq n$ pixels (default: $n=10$). 
The difference of the first-to-second closest ratio strategy and the FGINN strategy is illustrated in~Figure~\ref{fig:1st-inconsistent-illustration}.

\subsection{Geometric verification}
The last step of the MODS is the geometric verification. It consists of three substeps. 

\subsubsection{Duplicate filtering} 
The redetection of covariant features in synthetic views results in duplicates in tentative correspondences. The \emph{duplicate filtering} prunes correspondences with close spatial distance ($\approx$ 5~pixels) of local features in both images -- all these correspondences except one -- with smallest descriptor distance ratio -- are eliminated from the final correspondences list.
The number of pruned correspondences can be used later for evaluating the quality (probability of being correct) in PROSAC-like~\cite{Chum2005PROSAC} geometric verification.

\subsubsection{RANSAC}
The LO-RANSAC~\cite{Chum2003} algorithm searches for the maximal
set of geometrically consistent tentative correspondences. The model of the
transformation is set either to homography or epipolar geometry, or
automatically determined by a DegenSAC~\cite{Chum2005DEGENSAC} procedure.

\subsubsection{Local affine frame check}
\begin{figure}[hptb]]
\centering
\includegraphics[width=0.99\linewidth]{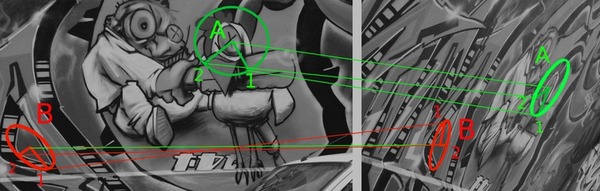}
\caption{The LAF-check. While centers of both regions A and B are consistent with found homography, farthest (1) and closest (2) points of the ellipse pass the check only for region A.}
\label{fig:laf-check}
\end{figure}

Since the epipolar geometry constraint is much less restrictive than a homography, wrong correspondences consistent with some (random) fundamental matrix appear. The local affine frame consistency check (LAF-check) eliminates virtually all incorrect correspondences. The procedure uses coordinates of the closest and furthest ellipse points from the ellipse center of both matched local affine frames to check whether the whole local feature is consistent with estimated geometry model (see Figure~\ref{fig:laf-check}). 
The check is performed for the geometric model obtained by  RANSAC. Regions which do not pass the check are discarded from the list of inliers. If the number of correspondences after the LAF-check is fewer than the user defined minimum, matcher continues with the next step of view synthesis. 

\section{Implementation and parameter setup}             \label{mods:parameters}
In this subsection, we discuss the tilt-rotation-detector setups of the MODS algorithm, and threshold selection for the \emph{first to first geometrically inconsistent -- FGINN} matching strategy validation.

\subsection{View synthesis for different detectors and descriptors}
The two main parameters of the view synthesis, tilt \{t\} sampling and the latitude step $\Delta\phi_{\text{base}}$, were explored in the following synthetic experiment. 

\begin{figure*}[htbp]
\centering
\includegraphics[height=1.65cm]{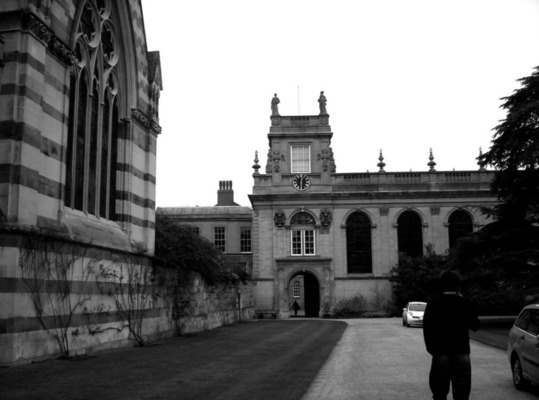}
\includegraphics[height=1.65cm]{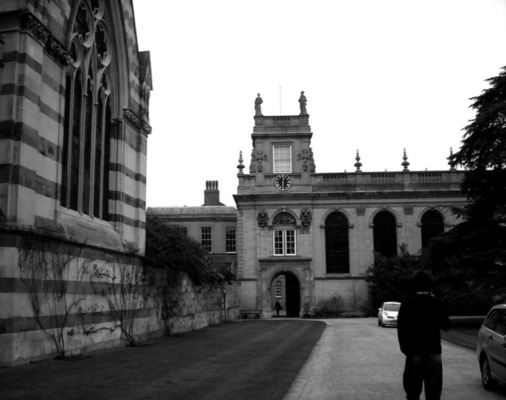}
\includegraphics[height=1.65cm]{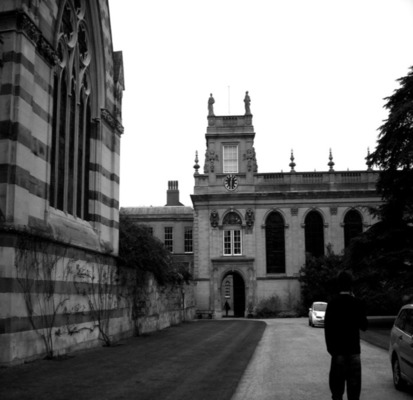}
\includegraphics[height=1.65cm]{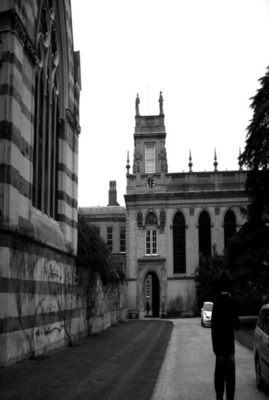}
\includegraphics[height=1.65cm]{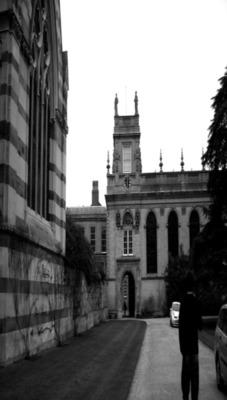}
\includegraphics[height=1.65cm]{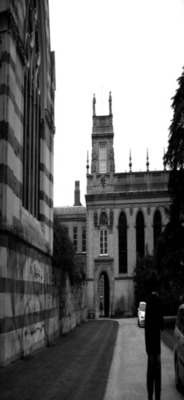}
\includegraphics[height=1.65cm]{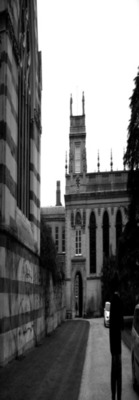}
\includegraphics[height=1.65cm]{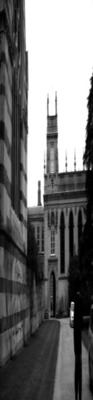}
\includegraphics[height=1.65cm]{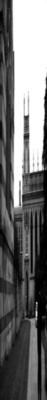}\\
\includegraphics[height=1.49cm]{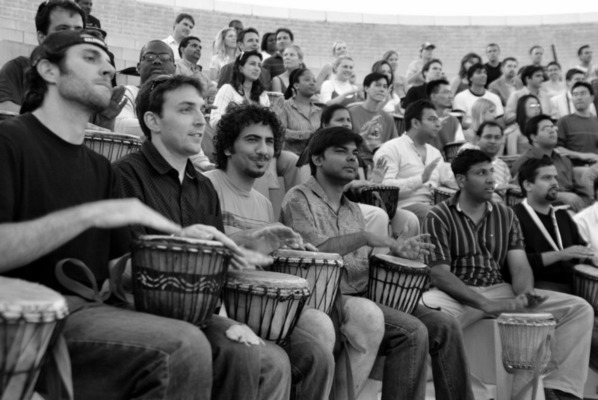}
\includegraphics[height=1.49cm]{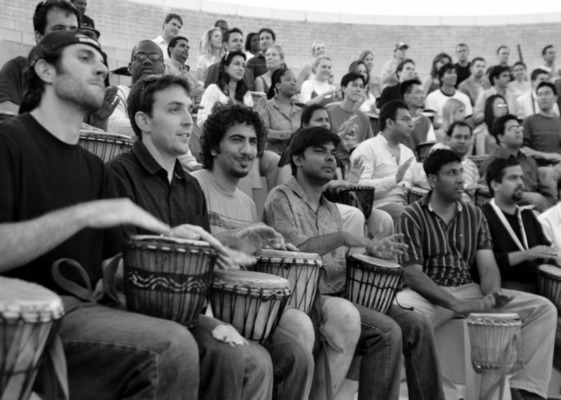}
\includegraphics[height=1.49cm]{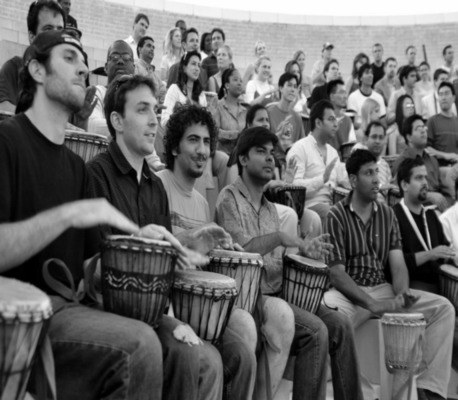}
\includegraphics[height=1.49cm]{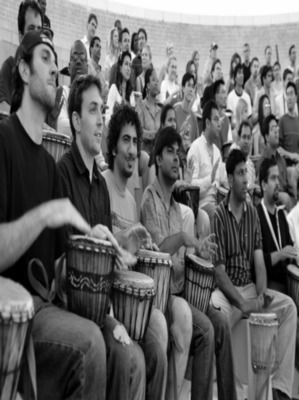}
\includegraphics[height=1.49cm]{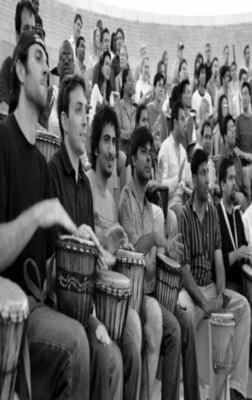}
\includegraphics[height=1.49cm]{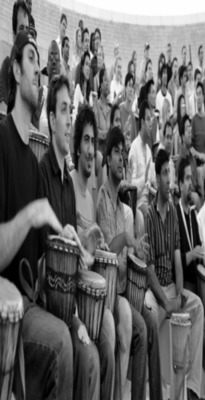}
\includegraphics[height=1.49cm]{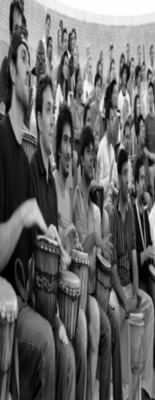}
\includegraphics[height=1.49cm]{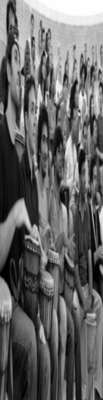}
\includegraphics[height=1.49cm]{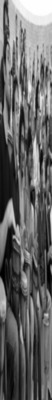}\\
\caption{Two examples of image sets from the synthetic dataset. Original unwarped images are from the Oxford buildings dataset~\cite{Philbin2007}.}
\label{fig:synthetic-dataset}
\end{figure*}

\begin{figure*}[htbp]
\includegraphics[width=0.49\linewidth]{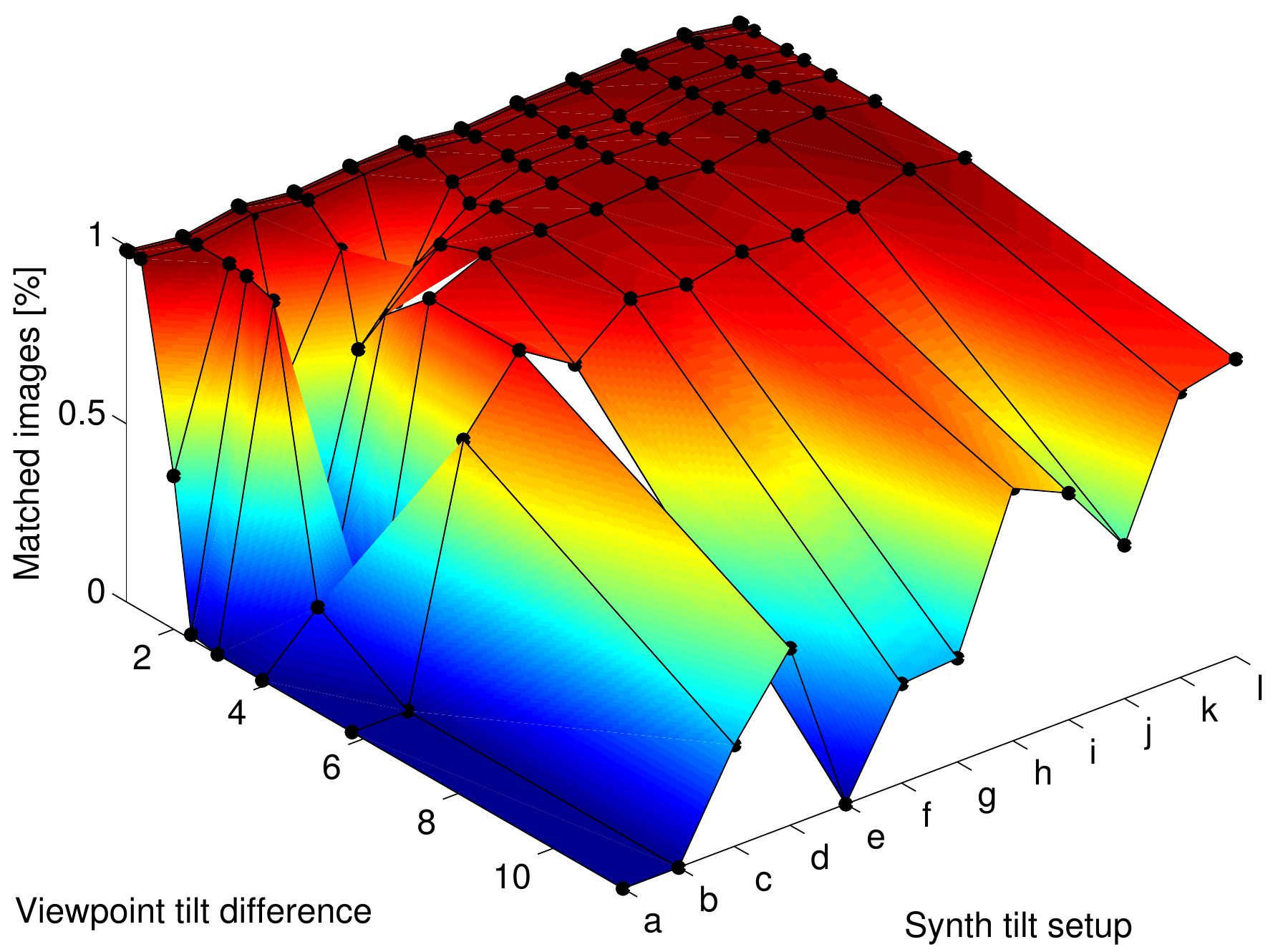}
\includegraphics[width=0.49\linewidth]{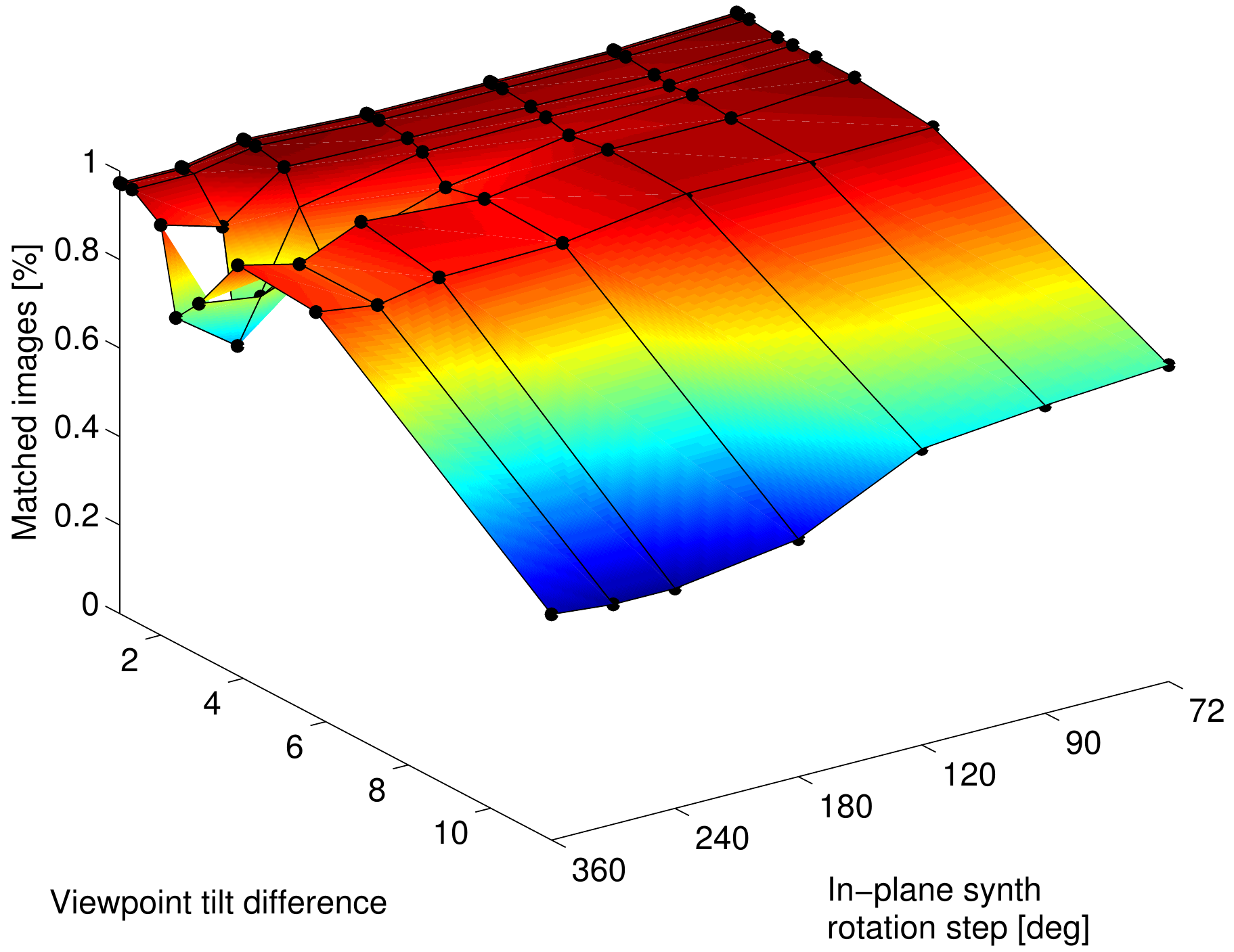}\\
\includegraphics[width=0.49\linewidth]{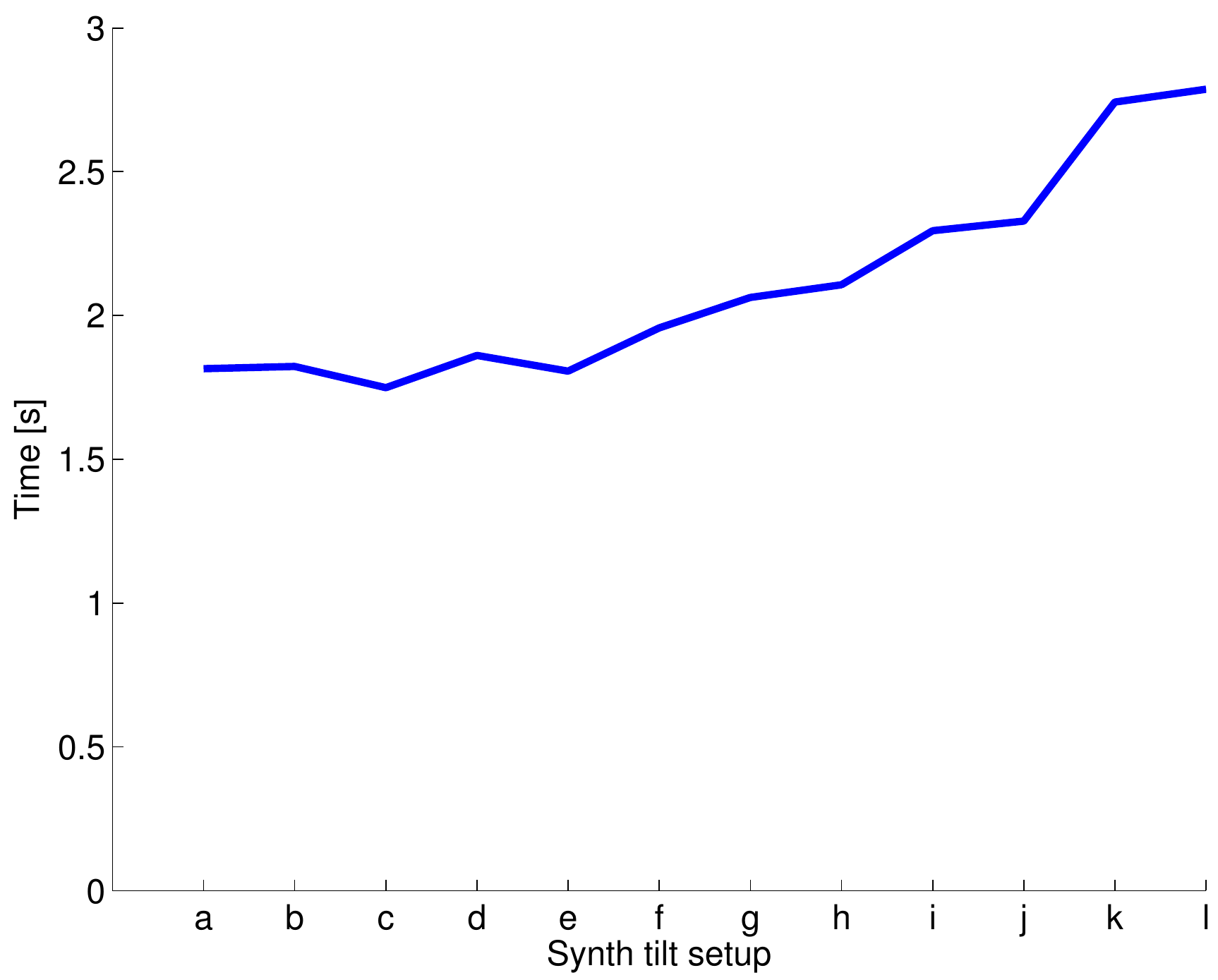}
\includegraphics[width=0.49\linewidth]{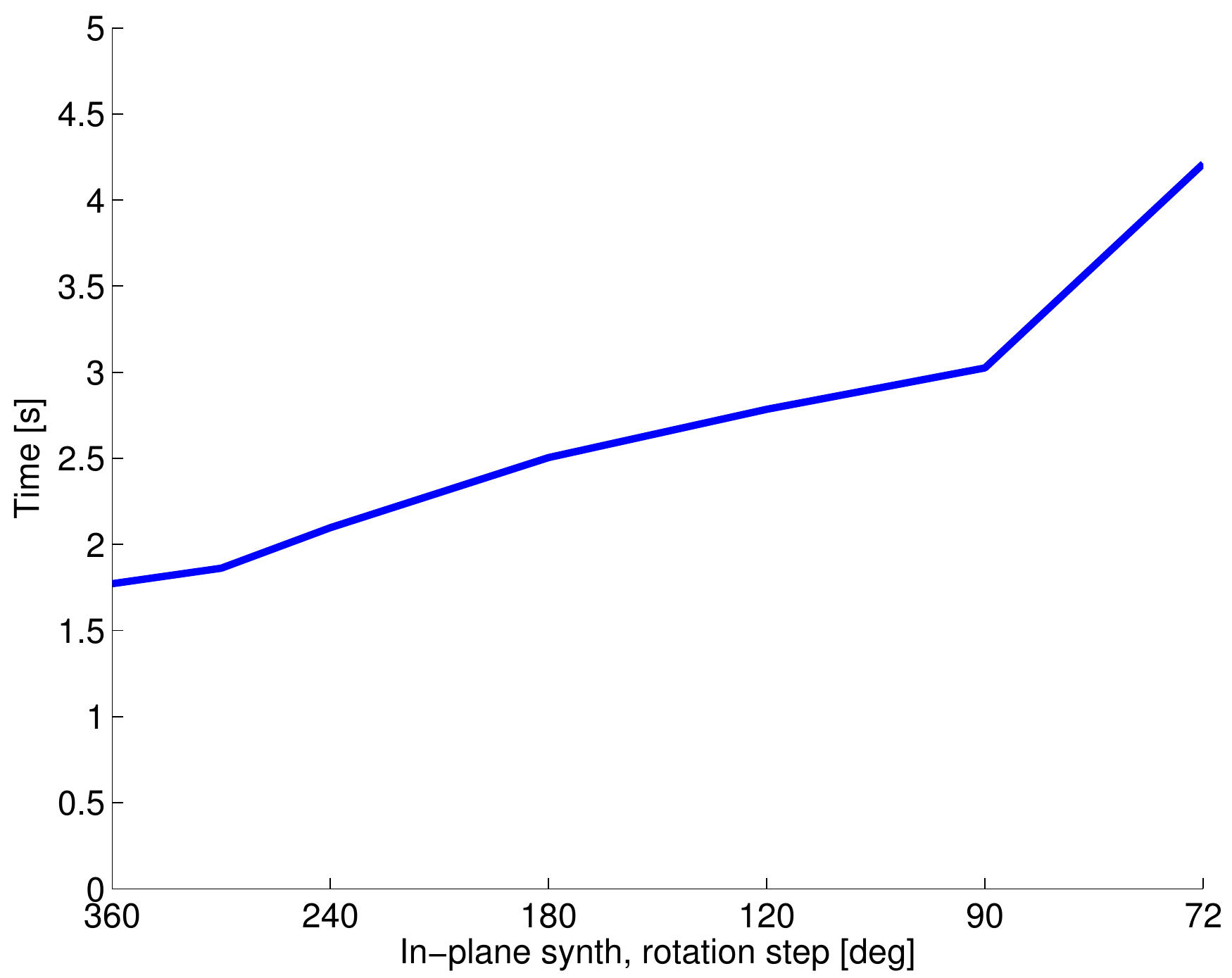}\\
\caption{Top: the percentage of images matched  depending the synthetic viewpoint difference for the DoG detector with tilt configurations given in Table~\ref{tab:view-synth-configs}.
Left: different tilt synthesis configurations (for $\Delta\phi = 240\degree$), right: different rotation synthesis configurations (for tilt set (d) = \{1,5,9\}).   
Bottom: running time per image pair for the respective configuration.}
\label{fig:dog-tilt-rot-setup}
\end{figure*}

A set of simulated views with latitudes angles $\theta=($0, 20, 40, 60, 65, 70, 75, 80, 85$\degree$), corresponding to tilt series $t=($1.00, 1.06, 1.30, 2.00, 2.36, 2.92, 3.86, 5.75, 11.47$)$\footnote{assuming that the original image is the fronto-parallel view} was generated for each of 150 random images from the  Oxford Building Dataset\footnote{available at http://www.robots.ox.ac.uk/$\sim$vgg/data/oxbuildings/}~\cite{Philbin2007}. Example images are shown in Figure~\ref{fig:synthetic-dataset}. The ground truth affine matrix $A$ was computed for each simulated view using equation~(\ref{eq:affine-matrix-decomposition}) and used in the final verification step. 
The original image was matched against its warped version, and  the running time and number of inliers for each combination of the detector, tilt and rotation (see Table~\ref{tab:view-synth-configs}) were computed. In all, 84 setups for each of the 8 detectors on the 150 image pairs were evaluated. 
As an example, we show the relation between the density of the view-sphere sampling and the number of images matched for the DoG detector in Figure~\ref{fig:dog-tilt-rot-setup}.  

Since our goal is to find a variety of detector-tilt-rotation configurations operating with different matching ability -- run-time trade-offs, we defined ``easy", ``medium" and ``hard" problems on the synthetic dataset. Successful two-view matching was defined as recovering $n \geq 50$ ground truth correspondences on a synthetically warped image. The threshold is set high -- synthetic warping of an image is underestimating the reduction of the number of matchable features induced by the effects of a corresponding viewpoint change e.g. due to non-planarity of the scene or illumination changes.
The matcher is considered to solve an ``easy'' problem if percentage of the matched images $f\geq 50\%$ of total images, ``medium'' if $f \geq 90\%$ of images matched and solved ``hard'' if  $f \geq 99\%$ of the images are matched.

%
%
%

The experiment with the synthetically warped dataset gives a hint  about the limits of configurations. Three configurations that solved the maximum tilt difference for each case fastest for a given detector were selected for evaluation. The configurations are specified in Table~\ref{tab:view-synth-configuration}. 

\begin{figure*}[htbp]
\includegraphics[width=0.34\linewidth]{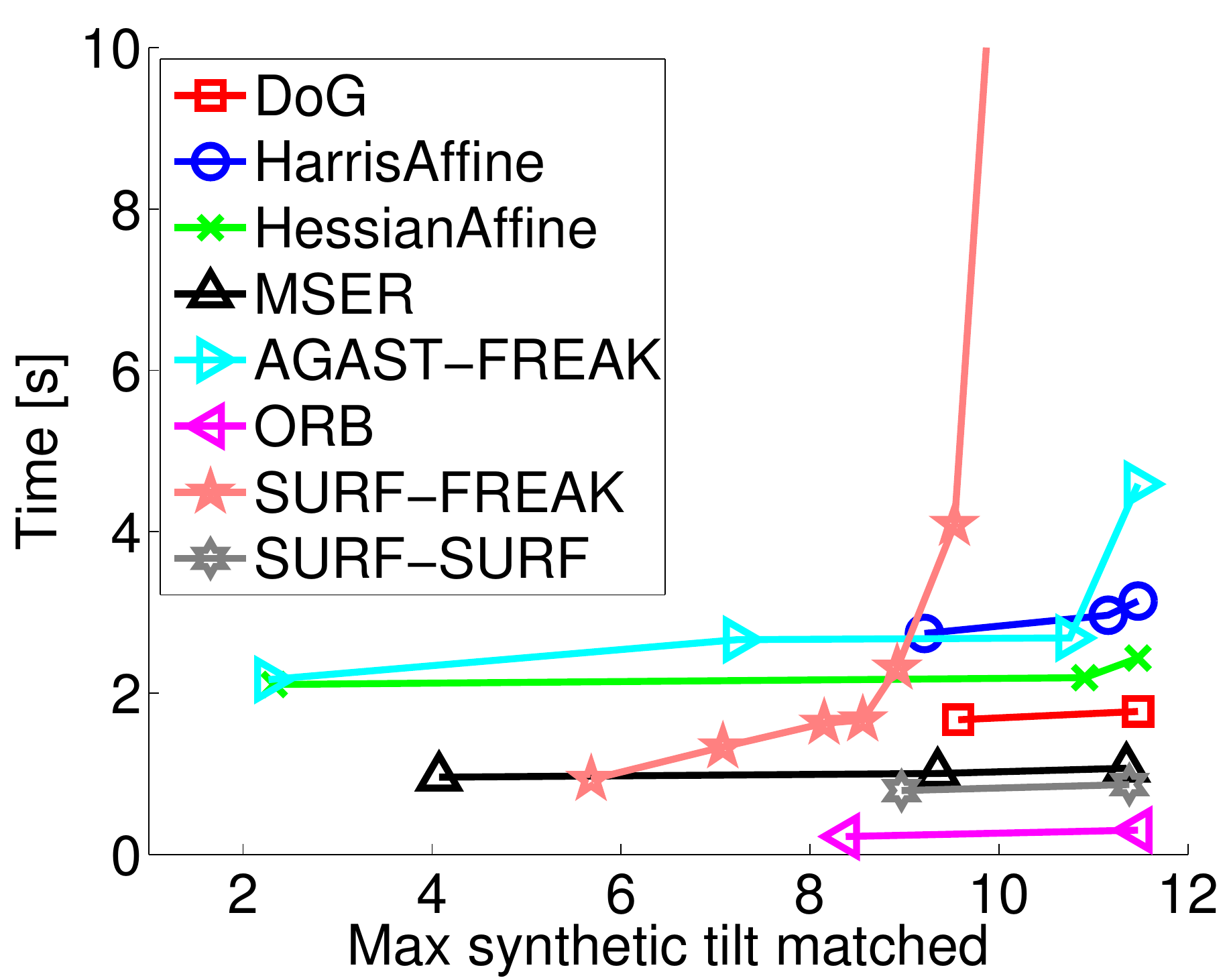}
\includegraphics[width=0.31\linewidth]{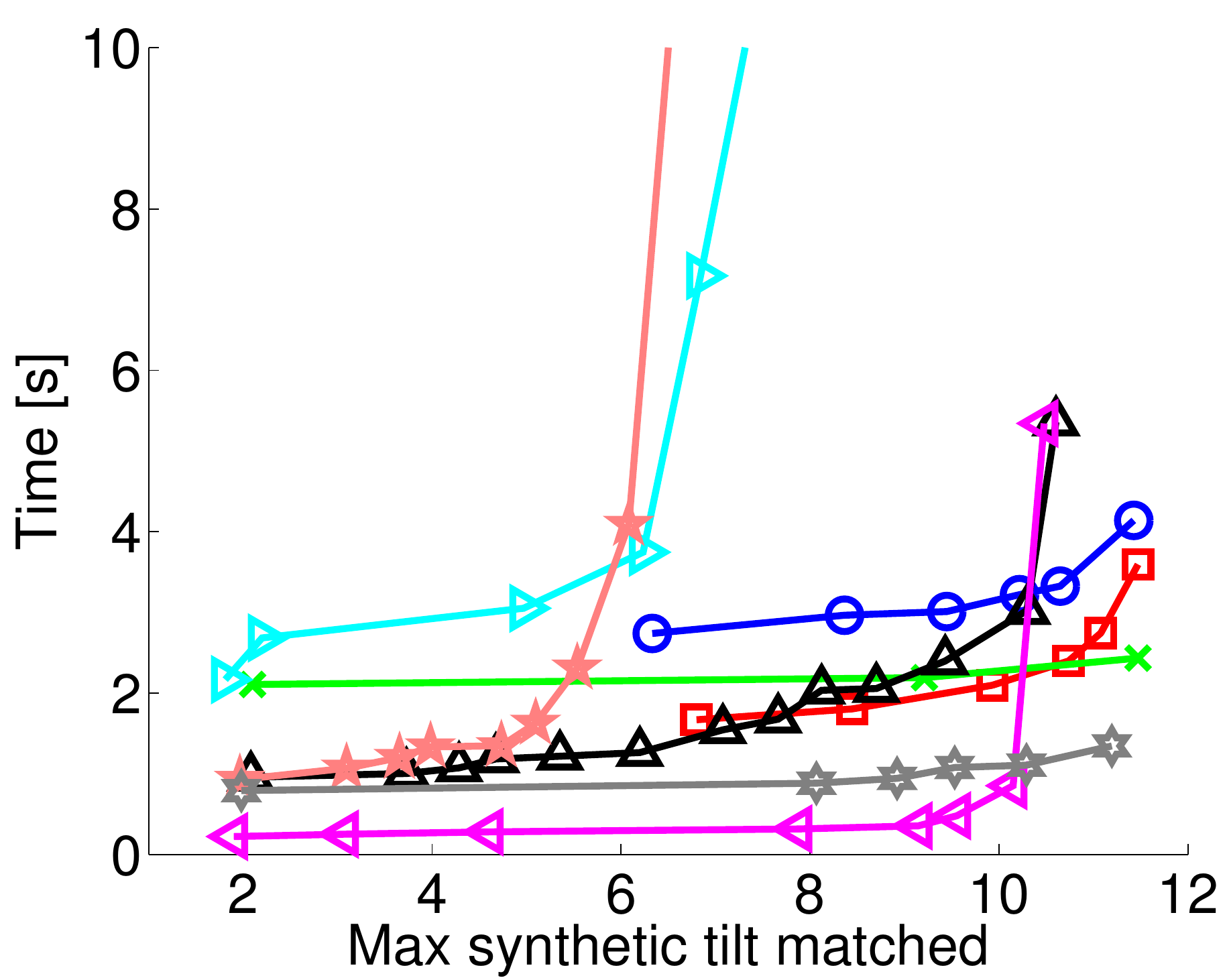}
\includegraphics[width=0.31\linewidth]{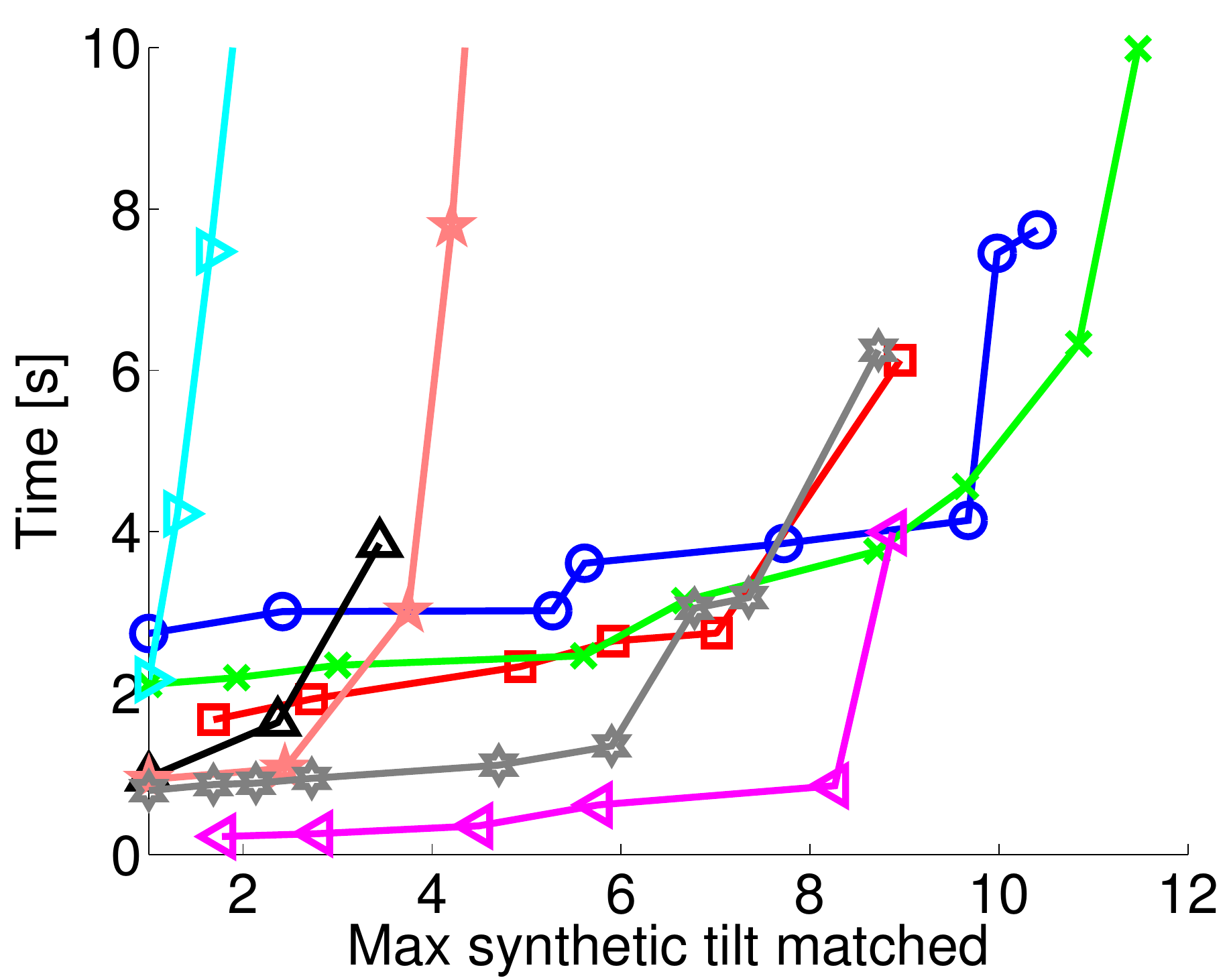}\\
\caption{Performance of view synthesis configurations  on the synthetic dataset. Average time needed  to match 
(left)``easy'', (center) ``medium'' and (right)``hard'' problems. The time for the fastest detector configuration that solved corresponding problems for a given tilt is shown.}
%
\label{fig:view-synth-configs}
\end{figure*}

The average time necessary to match a given synthetic tilt difference for different detectors with the optimal configuration is shown in Figure~\ref{fig:view-synth-configs}. The computations were performed on the Intel i7 3.9GHz (8 cores) desktop with 8Gb RAM with parallel processing.

Note that view synthesis significantly increases the matching performance of all detectors, but not uniformly. The left plot of Figure~\ref{fig:view-synth-configs} shows that a very sparse viewsphere sampling greatly improves matching at almost no computational cost for all detectors. However, after reaching a certain density,  additional views do not add correspondences in the hardest cases -- see the right graph of Figure~\ref{fig:view-synth-configs}. The ORB detector-descriptor clearly outperforms other detectors in terms of speed, but fails to match all images with the maximum tilt difference. The Hessian-Affine shown the best performance and it matched all pairs. 

\begin{table*}[tb]
\caption{View synthesis configuration evaluation. Configuration is a triplet - the detector, descriptor and the set of the synthesized views.}
\label{tab:view-synth-configs}
\footnotesize
\centering
\begin{tabu}spread 0pt{*{2}{|X[-1m,l]}|}
\hline
Detector-descriptor combination& DoG-SIFT, Hessian-Affine-SIFT, Harris-Affine-SIFT, MSER-SIFT, SURF-SURF, SURF-FREAK, AGAST-FREAK, ORB\\
\hline
Tilt set&  a = \{1\},  b = \{1,2\}, c = \{1,8\}, d = \{1,5,9\}, e = \{1,4\},f = \{1,4,8\}, g = \{1,2,4,8\}, h = \{1,3,6,9\}, i = \{1,4,8,10\}, j = \{1,2,4,6,8\}, 
 k = \{1,$\sqrt{2}$, 2, 2$\sqrt{2}$, 4, 4$\sqrt{2}$, 8\}, l = \{1,2,3,4,5,6,7,8,9\}.\\
\hline
$\phi_{base} [\degree]$ & 360, 240, 180, 120, 90, 72, 60\\
\hline
\end{tabu}
\end{table*}

\begin{table}[tb] \caption{View synthesis configurations with best synthetic dataset performance.}
\label{tab:view-synth-configuration} 
\footnotesize 
\centering 
\bgroup
\tabulinesep=0.1mm
\begin{tabu}spread 0pt{*{4}{|X[-1m,l]}|}
\hline
 & \multicolumn{3}{c|}{Configurations}\\
\hline
Detector&\centering {\sc Easy} &\centering {\sc Medium} & \centering{\sc Hard}\\
\hline
DoG-SIFT & $\{t\} = \{1;5;9\}$, $\Delta\phi=360\degree{}/t$  & $\{t\} = \{1;2;3;4;5;6;7;8;9\}$, $\Delta\phi=180\degree{}/t$ & $\{t\} = \{1;2;4;6;8\}$, $\Delta\phi=60\degree{}/t$  \\
\hline
HarrAff-SIFT &  $\{t\} = \{1;3;6;9\}$, $\Delta\phi=360\degree{}/t$ & $\{t\} = \{1;2;3;4;5;6;7;8;9\}$, $\Delta\phi=360\degree{}/t$ & $\{t\} = \{1;2;3;4;5;6;7;8;9\}$, $\Delta\phi=120\degree{}/t$ \\
\hline
HessAff-SIFT &  $\{t\} = \{1;5;9\}$, $\Delta\phi=360\degree{}/t$ & $\{t\} = \{1;5;9\}$, $\Delta\phi=360\degree{}/t$ & $\{t\} = \{1;2;4;6;8\}$, $\Delta\phi=60\degree{}/t$  \\
\hline
MSER-SIFT  &  $\{t\} = \{1;8\}$, $\Delta\phi=360\degree{}/t$ &  $\{t\} =  \{1;\sqrt{2}; 2; 2\sqrt{2}; 4;$ $4\sqrt{2}; 8\}$, $\Delta\phi=60\degree{}/t$ & $\{t\} = \{1;3;6;9\}$, $\Delta\phi=60\degree{}/t$  \\
\hline
AGAST-FREAK  &  $\{t\} = \{1;4;8;10\}$, $\Delta\phi=360\degree{}/t$ & $\{t\} = \{1;3;6;9\}$, $\Delta\phi=60\degree{}/t$ & $\{t\} = \{1;3;6;9\}$, $\Delta\phi=72\degree{}/t$  \\
\hline
ORB &  $\{t\} = \{1;5;9\}$, $\Delta\phi=360\degree{}/t$& $\{t\} = \{1;2;3;4;5;6;7;8;9\}$, $\Delta\phi=90\degree{}/t$ & $\{t\} =  \{1;\sqrt{2}; 2; 2\sqrt{2}; 4;$ $4\sqrt{2}; 8\}$, $\Delta\phi=72\degree{}/t$ \\
\hline
SURF-FREAK &  $\{t\} =  \{1;\sqrt{2}; 2; 2\sqrt{2}; 4;$ $4\sqrt{2}; 8\}$, $\Delta\phi=60\degree{}/t$ & $\{t\} =  \{1;\sqrt{2}; 2; 2\sqrt{2}; 4;$ $4\sqrt{2}; 8\}$, $\Delta\phi=90\degree{}/t$& $\{t\} =  \{1;\sqrt{2}; 2; 2\sqrt{2}; 4;$ $4\sqrt{2}; 8\}$, $\Delta\phi=72\degree{}/t$ \\
\hline
SURF-SURF &  $\{t\} = \{1;5;9\}$, $\Delta\phi=360\degree{}/t$ & $\{t\} = \{1;2;3;4;5;6;7;8;9\}$, $\Delta\phi=360\degree{}/t$& $\{t\} = \{1;2;3;4;5;6;7;8;9\}$, $\Delta\phi=72\degree{}/t$ \\
\hline
\end{tabu}
\egroup
\end{table}

\subsection{First geometrically inconsistent nearest neighbor ratio correspondence selection strategy}
\begin{figure}[tb]
\includegraphics[width=0.49\linewidth]{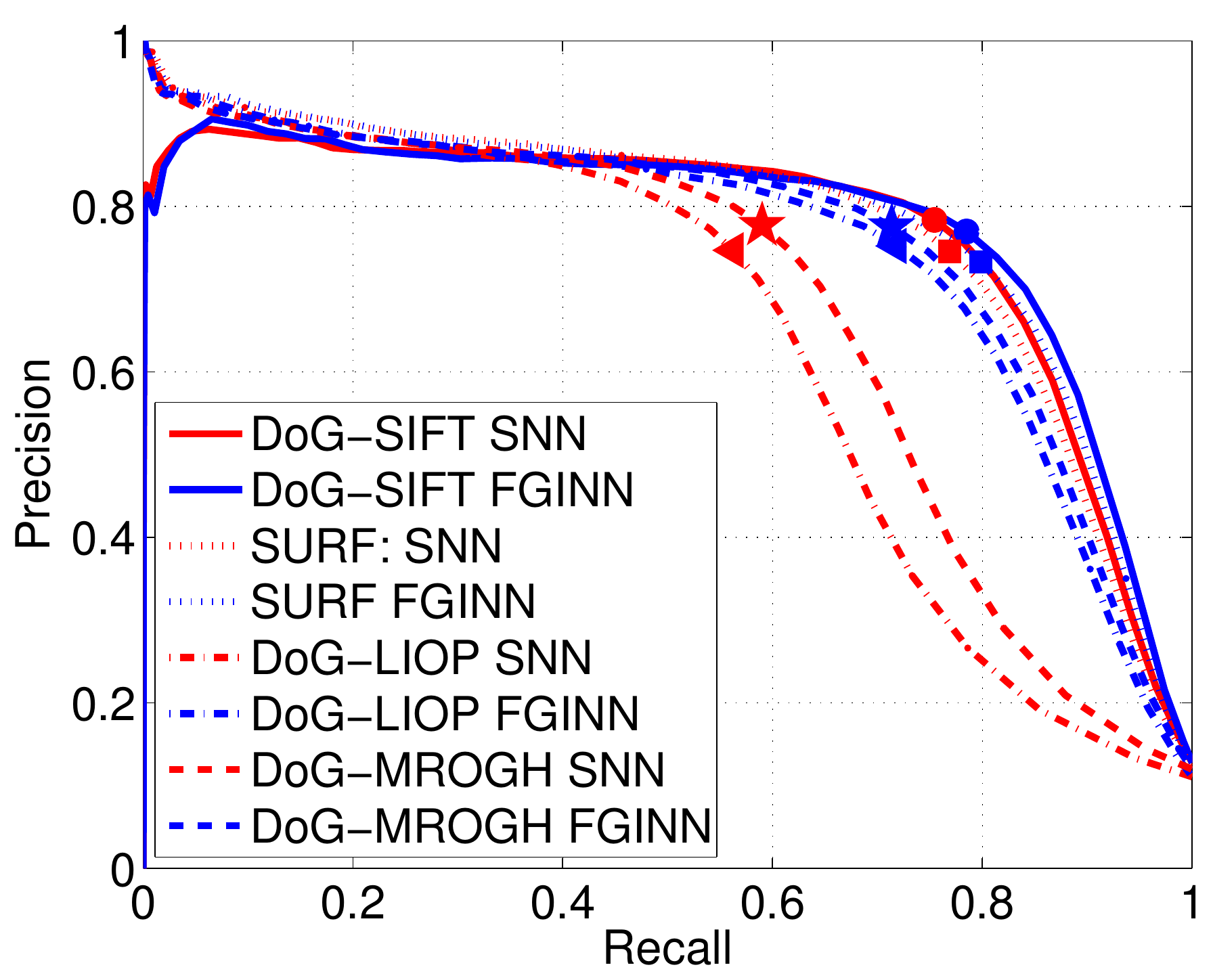}
\includegraphics[width=0.49\linewidth]{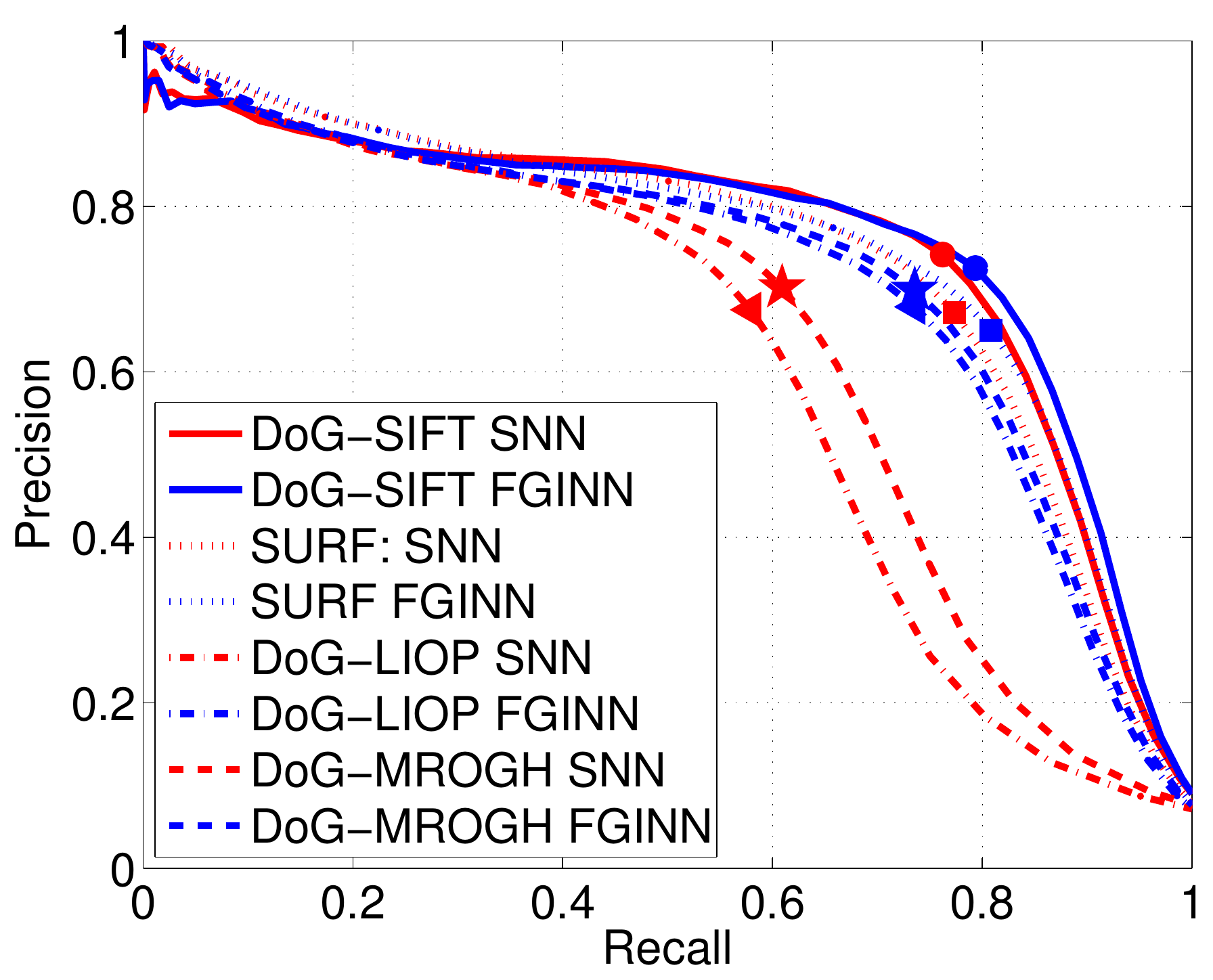}\\
 \caption{Comparison of the FGINN and SNN matching strategies for SIFT, SURF, LIOP and MROGH on images from the Oxford~\cite{Mikolajczyk2005a} and Cordes et.al.~\cite{Cordes2011} datasets. Markers show operationg points for the common distance ratio threshold = 0.8.  Matching without (left) view synthesis an using view synthesis (right) with parameters  $\{t\} = \{1;5;9\}$, $\Delta\phi=360\degree{}/t$. The recall and precision of the correspondence filtering step, therefore the maximum recall is 1 when all correspondences are kept.}
 \label{fig:1st-to-2nd-synth}
 \end{figure} 
The following protocol was used to find the thresholds and to evaluate the performance of the proposed First Geometrically Inconsistent Nearest Neighbor \emph{FGINN} strategy. First, similarity covariant regions were detected using the DoG detector (we also tried Hessian-Affine, MSER and SURF, with very similar results) and described using four popular descriptors -- RootSIFT, SURF, LIOP~\cite{Gupta2010} and MROGH~\cite{Fan2012} which are typically matched with the second-nearest region SNN strategy.  Then for each keypoint descriptor, the  first, second and first geometrically inconsistent descriptors in the other image were found. The matching keypoints were then labeled as correct if their Sampson error was within 1 pixel of the ground truth location given by homography for the image pair, and incorrect otherwise. 

The experiment  was performed on 26 image pairs of the publicly available  datasets~\cite{Mikolajczyk2005a},\cite{Cordes2011} (image pairs 1-3, precise homography provided) and ~\cite{Lebeda2012} (the homography was estimated using provided precise ground truth correspondences).
The recall-precision curves for correspondences from all images were plot with a varying  ratio threshold from 0 to 1 in Figure~\ref{fig:1st-to-2nd-synth}.
The FGINN curves for SIFT and SURF slightly outperform standard SNNs, while for LIOP and MROGH the difference is much more significant. The significantly higher benefit of  the FGINN rule for LIOP and MROGH can be explained by their lower sensitivity to keypoint shift which in turn means that undesirable suppression of keypoints happens in a larger neighborhood. 
The lower sensitivity to shifts was experimentally verified.
\section{Experiments}        \label{sec:main}     
We have tested MODS and, as a baseline, ASIFT\footnote{Reference code from
http://demo.ipol.im/demo/my\_affine\_sift} and single detector configurations specified in Table~\ref{tab:view-synth-configuration} on seven public. datasets~\cite{Moreels2007},\cite{Mishkin2013},\cite{Altwaijry2013}, \cite{Hauagge2012}, \cite{Kelman2007}, \cite{Aguilera2012}, \cite{Mikulik2014}.
\paragraph{Implementation details of the MODS algorithm and parameter setting}
The kd-tree algorithm from FLANN library~\cite{Muja2009} was used to efficiently find the N-closest descriptors.  The distance ratio thresholds of the FGINN matching strategy were experimentally selected based on the CDFs of matching and non-matching descriptors.

The MODS algorithm allows to set the minimum desired number of inliers which have a very low probability to be a random result as a stopping criterion. The recommended value -- 15 inliers to the homography -- did not produce a false positive results in experiments. Computations were performed on Intel i3 CPU @ 2.6GHz with 4Gb RAM with 4 cores.
\subsection{MODS variants testing on Extreme Viewpoint and Oxford Dataset}
\label{main:real-world-testing}
To evaluate the performance of matching algorithms, we introduce a two-view matching evaluation dataset\footnote{Available at http://cmp.felk.cvut.cz/wbs/index.html} with extreme viewpoint changes, see Table~\ref{tab:image-dataset}. The dataset includes image pairs from publicly available datasets: {\sc adam} and {\sc mag}~\cite{Morel2009}, {\sc graf}~\cite{Mikolajczyk2005a} and {\sc there}~\cite{Cordes2011}. The ground truth homography matrices were estimated by LO-RANSAC using correspondences from all detectors in view synthesis configuration $\{t\} = \{1;\sqrt{2};2;2\sqrt{2};4;4\sqrt{2};8\}$, $\Delta\phi=72\degree{}/t$. The number of inliers for each image pair was $\geq$
50 and the homographies were manually inspected. For the image pairs {\sc graf} and {\sc there} precise homographies are provided by
Cordes~et.al.~\cite{Cordes2011}. Transition tilts $\tau$ were computed using equation~(\ref{eq:affine-matrix-decomposition}) with SVD decomposition of the linearized homography at center of the first image of the pair (see Table ~\ref{tab:image-dataset}). Oxford~\cite{Mikolajczyk2005a} dataset with 42 image pairs (1-2, ..., 1-6) was used for easier wide baseline problems.

\paragraph{Experimental protocol} The evaluated algorithms matched image pairs and the output keypoints correspondences were checked with ground truth homographies. The image pair is considered as solved, when at least 10 output correspondences are correct.

\begin{table}[htbp]
\caption{The Extreme View Dataset -- EVD. Image sources: C -- \cite{Cordes2011}, Ox -- ~\cite{Mikolajczyk2005a}, M -- \cite{Morel2009}. }
\label{tab:image-dataset}
\centering
\scriptsize
\bgroup
\def\arraystretch{1.1}
\setlength{\tabcolsep}{.54em}
\tabulinesep=0.5mm
\begin{tabu}spread 0pt{*{16}{|X[-1m,c]}|}
\hline
\# & 1 & 2& 3&4&5&6&7&8&9&10&11&12&13&14&15\\
\hline
Name & {\sc there} &{\sc graf}&{\sc adam}&{\sc mag}&{\sc grand}&{\sc pkk}&{\sc face}&{\sc girl}&{\sc shop}&{\sc dum}&{\sc index}&{\sc cafe}&{\sc fox}&{\sc cat}&{\sc vin}\\
Ref. & C & Ox &M & M& EVD &EVD & EVD& EVD& EVD & EVD &EVD&EVD &EVD & EVD&EVD \\
$\tau$ -- tilt &6.3 &3.6 &4.8& 20& 2.9 & 7.1&6.9 &8.0&9.1 & 6.9&8.5&11.9&22.5&47&49.8\\
\hline
\shortstack{Size \\$[$pxl$]$ }&\shortstack{1536\\x\\1024} &\shortstack{800\\x\\640} & \shortstack{600\\x\\450}& \shortstack{600\\x\\450}& \shortstack{1000\\x\\667} & \shortstack{1000\\x\\750}&\shortstack{1000\\x\\750} &\shortstack{1000\\x\\750}&\shortstack{1000\\x\\562}&\shortstack{1000\\x\\729} & \shortstack{1000\\x\\750}&\shortstack{800\\x\\533}&\shortstack{1000\\x\\563}&\shortstack{1000\\x\\598}&\shortstack{1000\\x\\715}\\
\hline
\end{tabu}
\egroup
\\
\bgroup
\def\arraystretch{1.2}%
\setlength{\tabcolsep}{.16667em}
\tabulinesep=0.2mm
\begin{tabu*}spread 0pt{*{9}{|X[-1m,c]}|}
\hline
\# & Image 1 & Image 2 &\# & Image 1 & Image 2&\# & Image 1 & Image 2 \\
\hline
1
& \includegraphics[height=1.1cm]{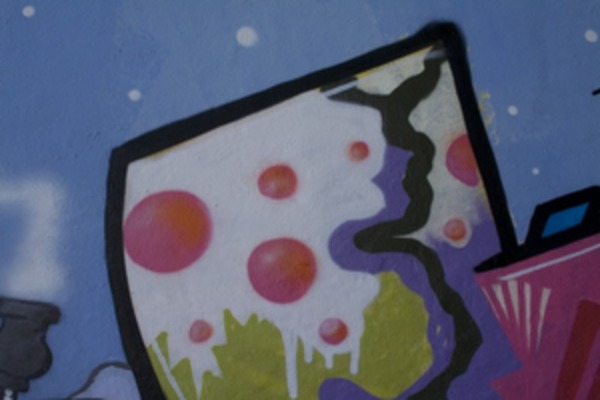} 
& \includegraphics[height=1.1cm]{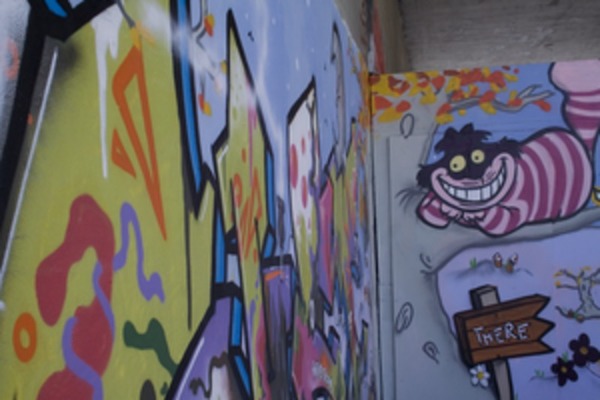}
& 6 
& \includegraphics[height=1.1cm]{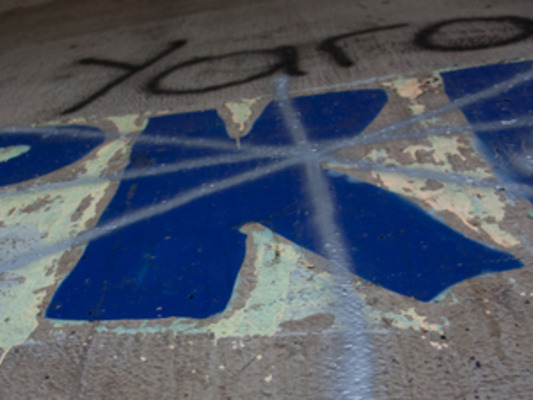} 
& \includegraphics[height=1.1cm]{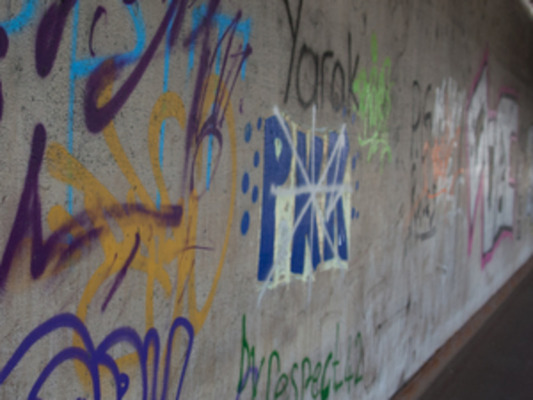} 
&11
& \includegraphics[height=1.1cm]{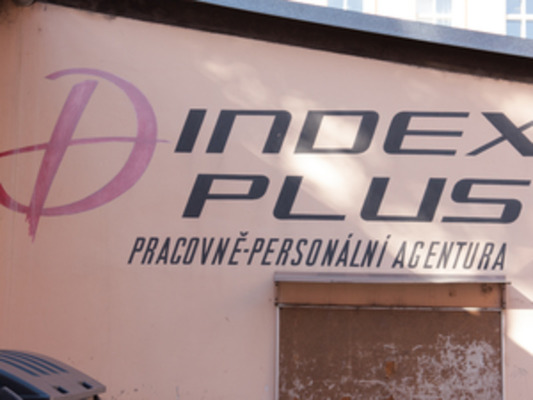} 
& \includegraphics[height=1.1cm]{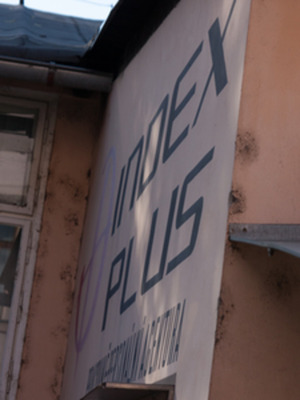} \\ 
\hline
2
& \includegraphics[height=1.1cm]{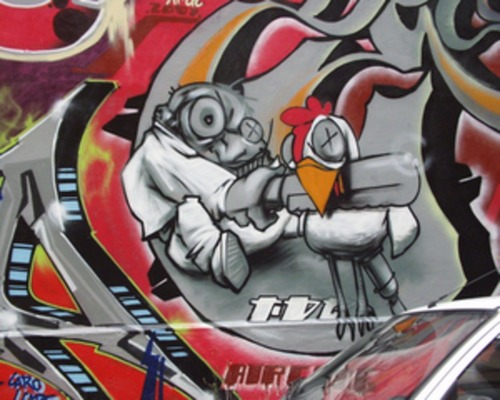} 
& \includegraphics[height=1.1cm]{graf2.jpg}
&7
& \includegraphics[height=1.1cm]{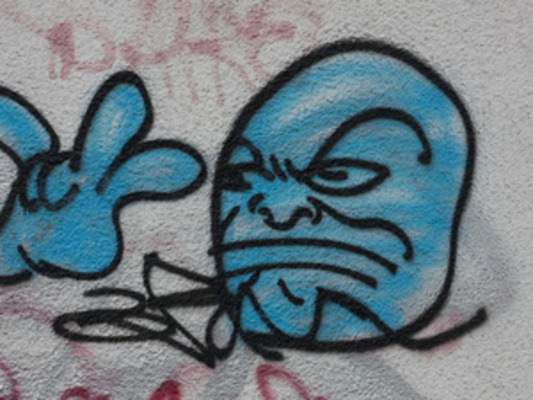} 
& \includegraphics[height=1.1cm]{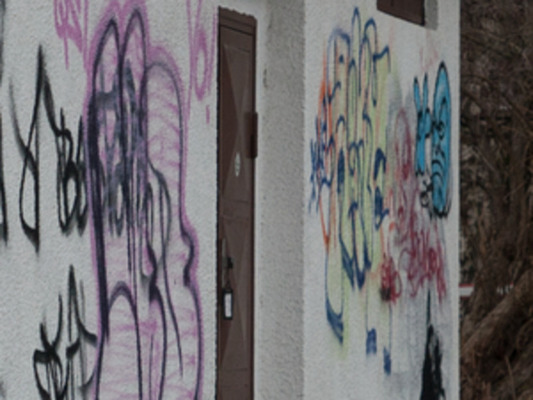}
&
12& \includegraphics[height=1.1cm]{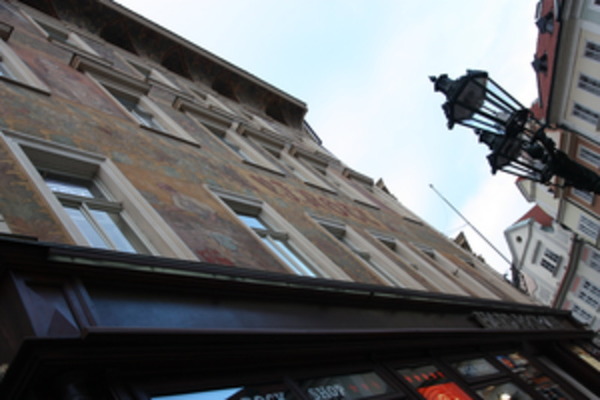} 
& \includegraphics[height=1.1cm]{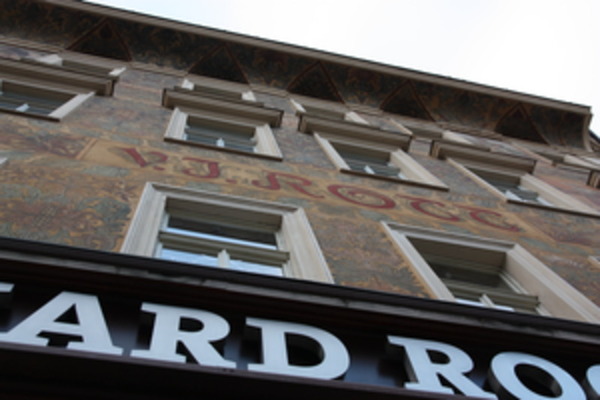} \\ 
\hline
3 
& \includegraphics[height=1.1cm]{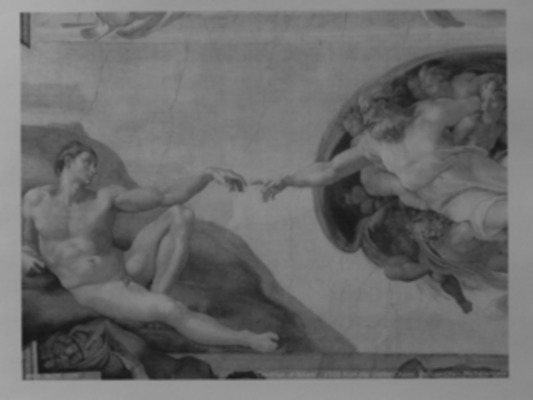} 
& \includegraphics[height=1.1cm]{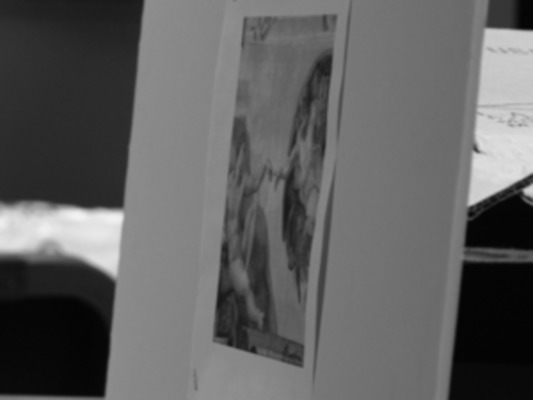} 
&8
& \includegraphics[height=1.1cm]{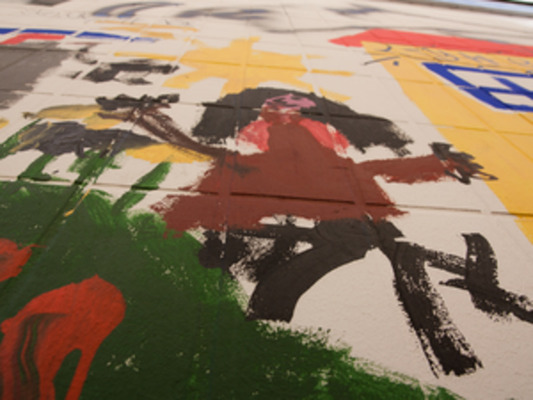} 
& \includegraphics[height=1.1cm]{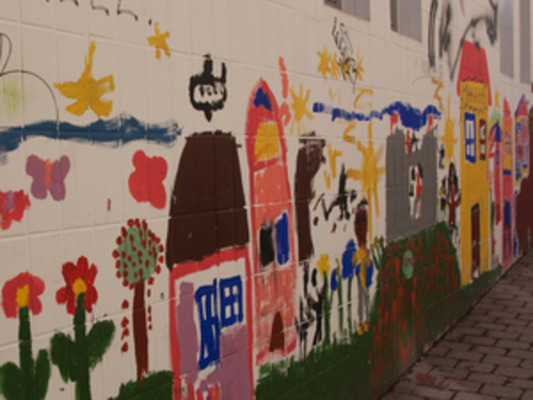}
& 13
& \includegraphics[height=1.1cm]{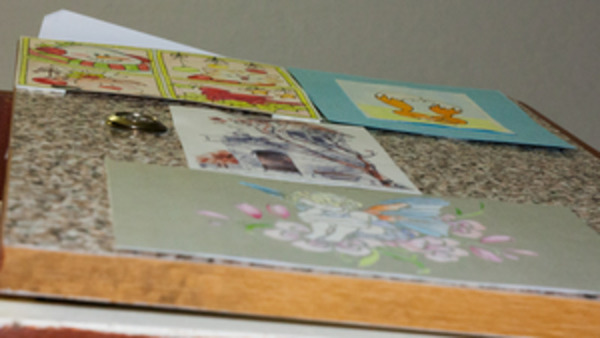} 
& \includegraphics[height=1.1cm]{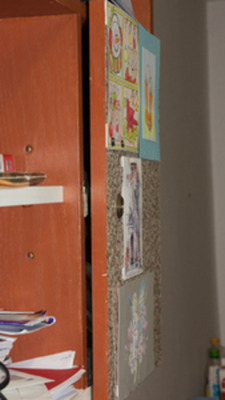}\\ 
\hline
4
& \includegraphics[height=1.1cm]{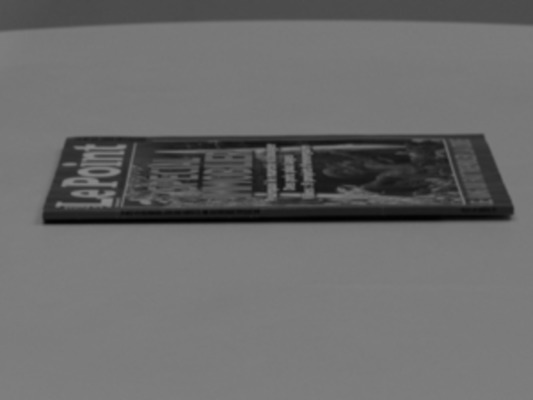} 
& \includegraphics[height=1.1cm]{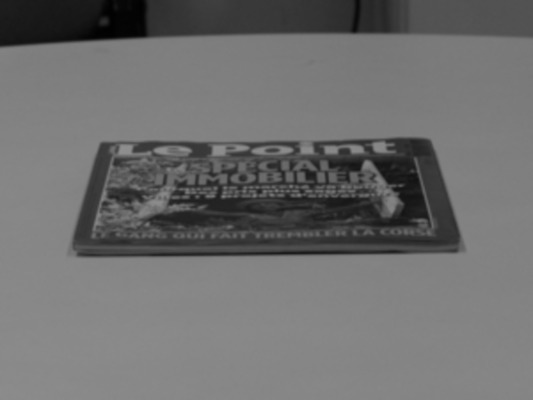}
&9 
& \includegraphics[height=1.1cm]{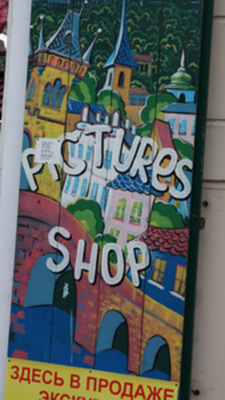} 
& \includegraphics[height=1.1cm]{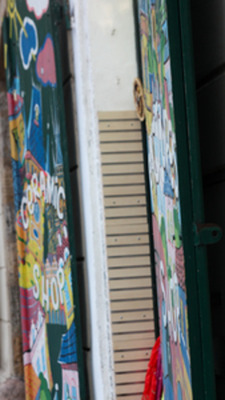}
&14 
&\includegraphics[height=1.1cm]{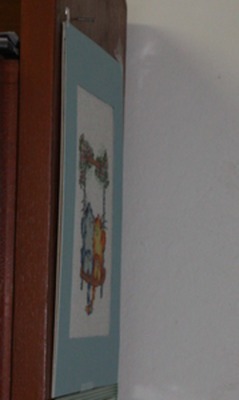} 
& \includegraphics[height=1.1cm]{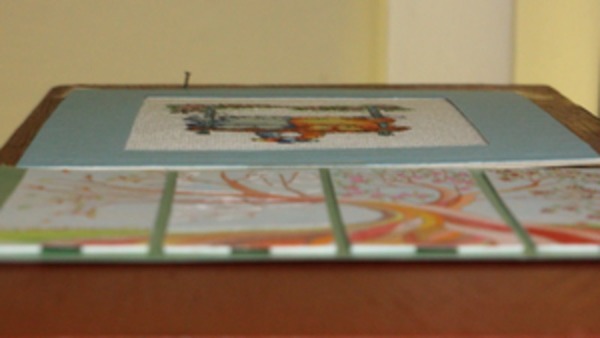} \\
\hline
5
&  \includegraphics[height=1.1cm]{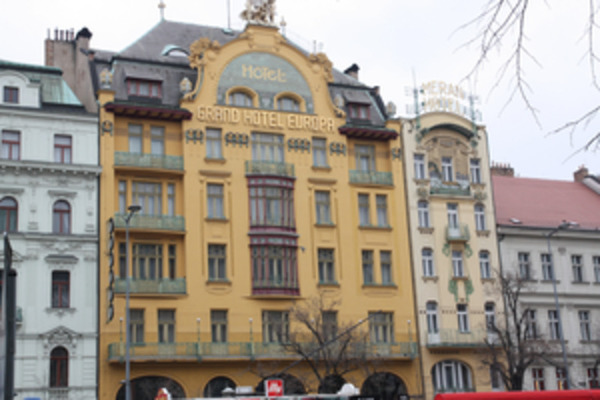} 
& \includegraphics[height=1.1cm]{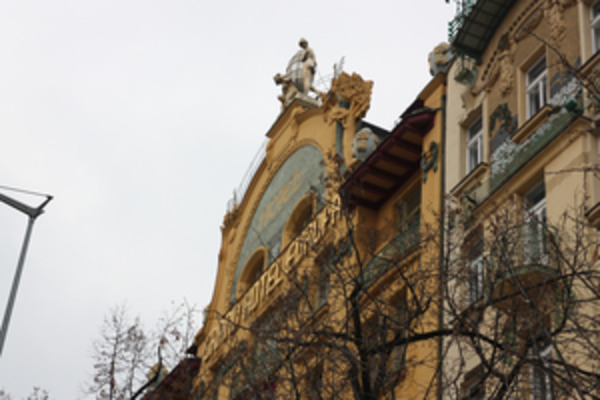}
&10
& \includegraphics[height=1.1cm]{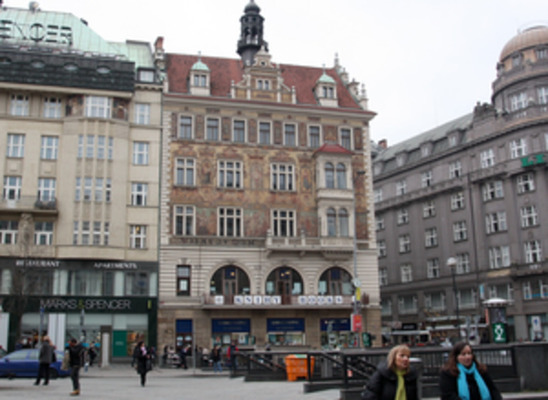} 
& \includegraphics[height=1.1cm]{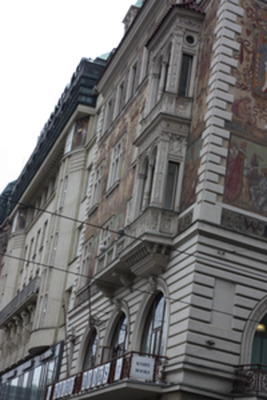}
&15
& \includegraphics[height=1.1cm]{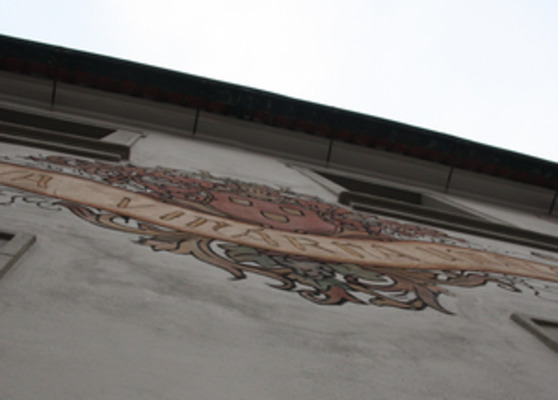}
& \includegraphics[height=1.1cm]{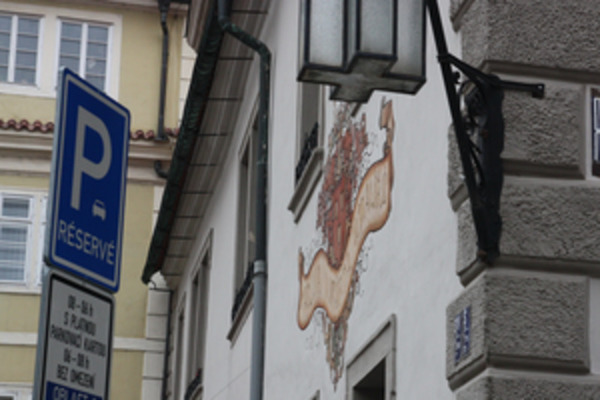} \\ 
\hline
\end{tabu*}
\egroup
\end{table}
%
\begin{table}[tb]
\caption{ASIFT view synthesis configuration}
\label{tab:ASIFT-configuration}
\footnotesize 
\centering
\bgroup
\tabulinesep=0.4mm
\begin{tabu} spread 0pt{|X[-1m,c]|X[-1,m,l]|}
\hline
Detector& \centering View synthesis setup\\
\hline
DoG & $\{S\} = \{1\}$, $\{t\} = \{1;\sqrt{2};2;2\sqrt{2};4;4\sqrt{2};8\}$, $\Delta\phi=72\degree{}/t$\\
\hline
\end{tabu}
\egroup
\end{table}

\begin{figure*}[htbp]
\centering
\includegraphics[width=0.34\linewidth]{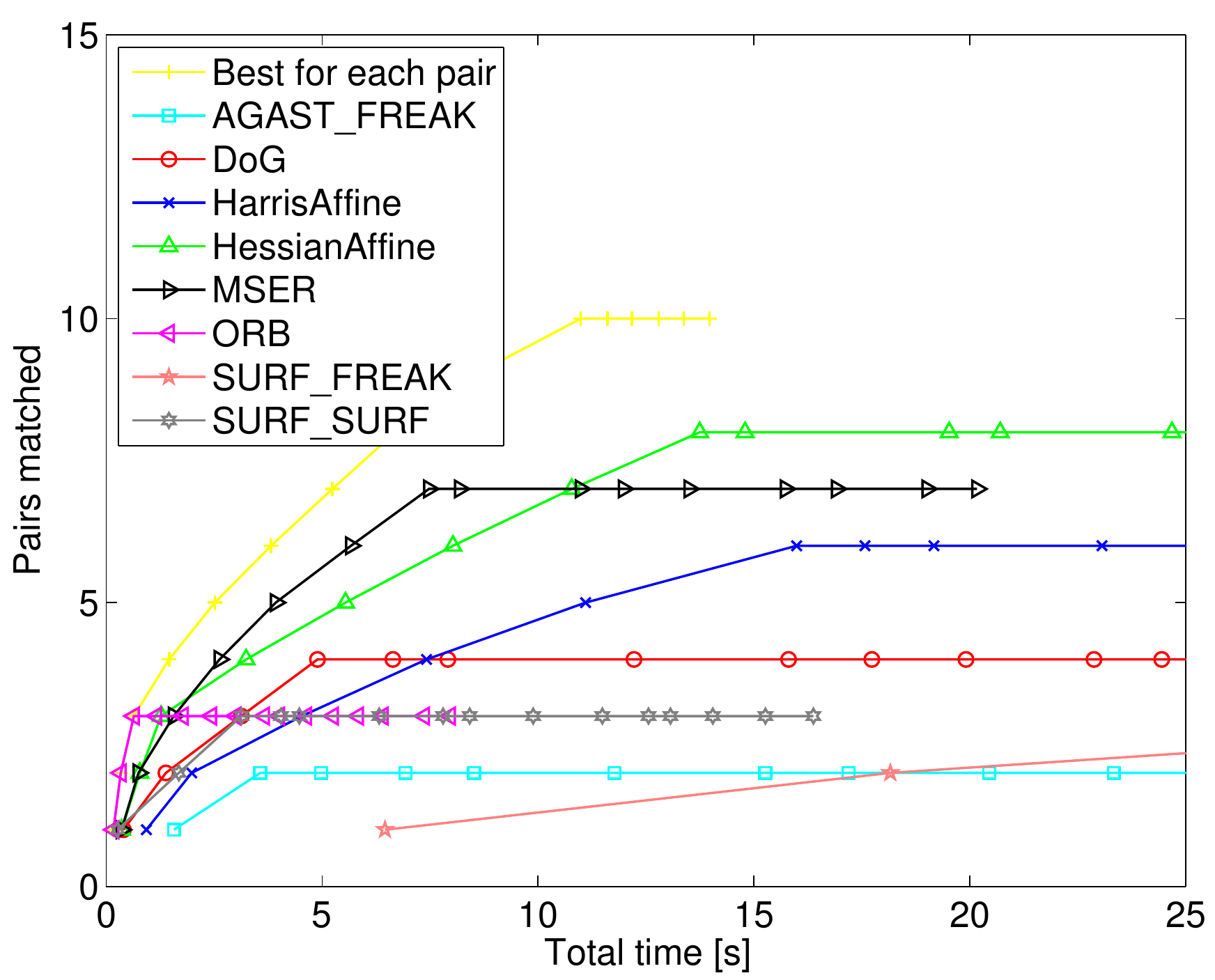}
\includegraphics[width=0.32\linewidth]{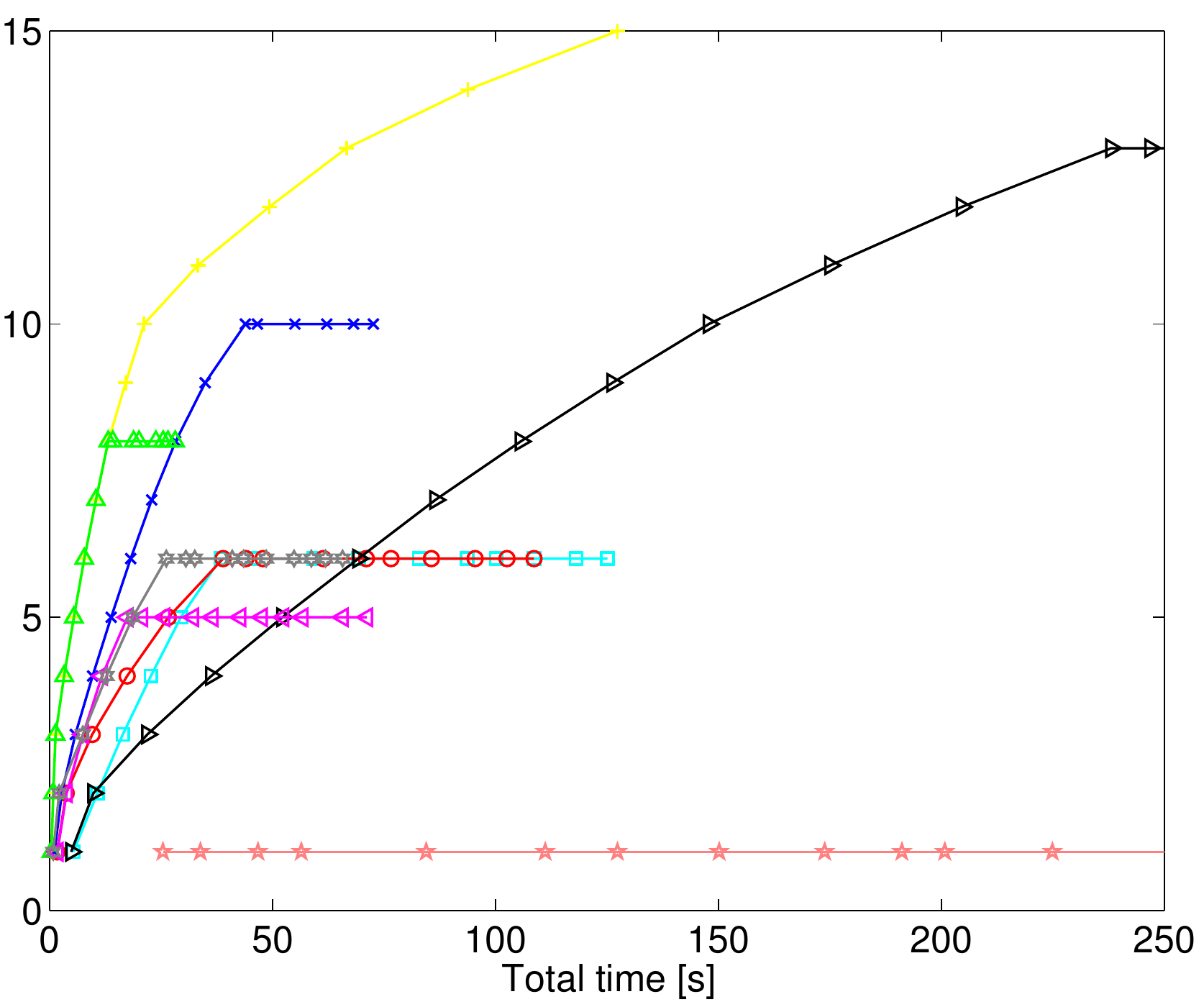}
\includegraphics[width=0.32\linewidth]{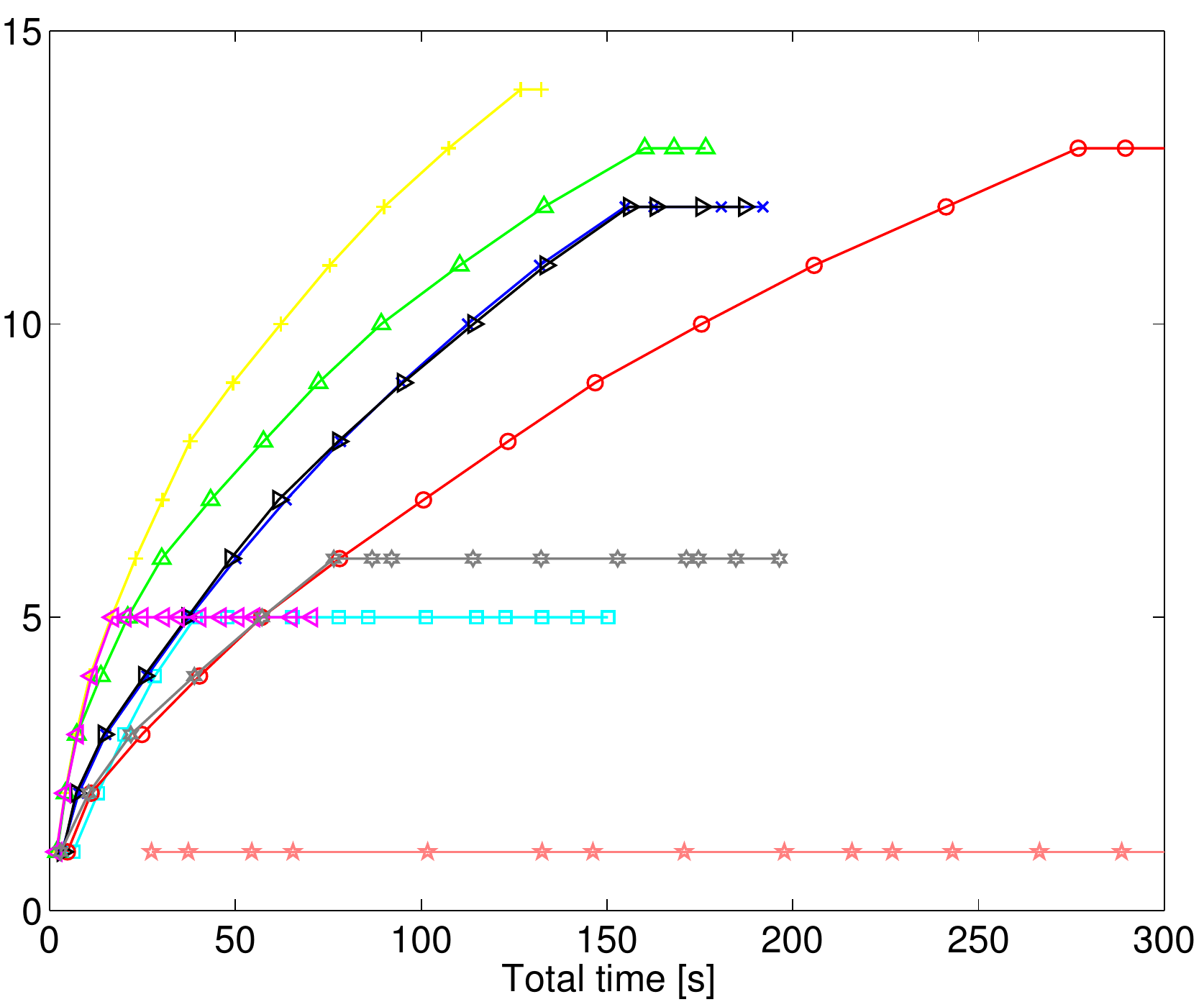}\\
\caption{Performance  of different configurations on the Extreme View Dataset. Cumulative percentage of image pairs successfully matched in a given time for configurations defined in
Table~\ref{tab:view-synth-configuration}. Each mark represents one image pair. The fastest detector configuration which was able to match each pair was selected. Left -- 'easy', middle -- 'medium', left -- 'hard' configurations. An images is considered matched if 10 correct inliers were found.}
\label{fig:evd-configuration-performance}
\end{figure*}

Figure~\ref{fig:evd-configuration-performance} compares the different view synthesis configurations. Note that no single detector solved all image pairs. The Hessian-Affine, MSER, Harris-Affine and DoG successfully solved resp. 13, 13, 12 and 13 out of the 15 image pairs however, at the expense of the high computational cost. We also noticed that if one would know the suitable detector and configuration for each image, it is possible to match all image pairs. 

\begin{figure*}[htbp]
\centering
\includegraphics[width=0.51\linewidth]{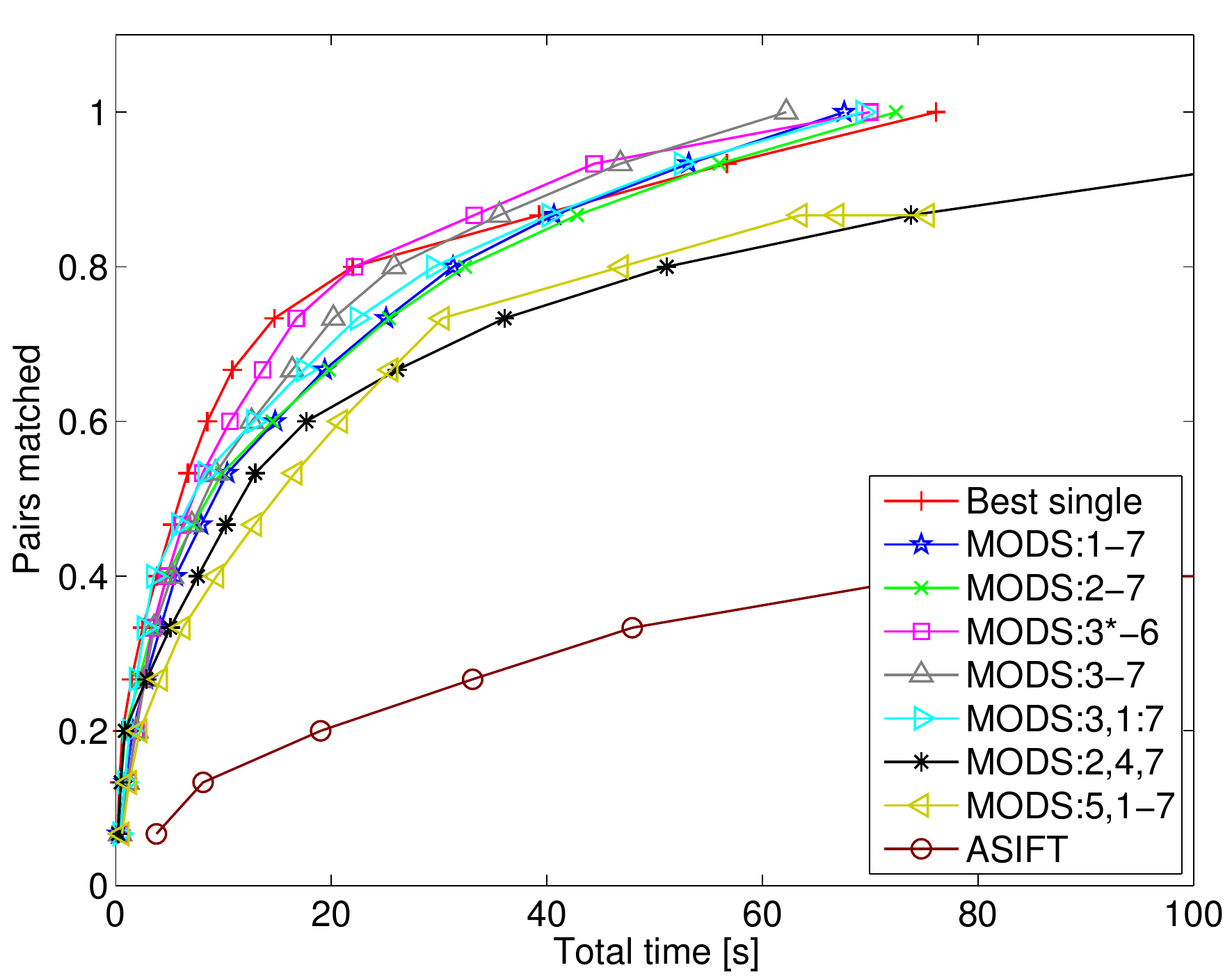}
\includegraphics[width=0.48\linewidth]{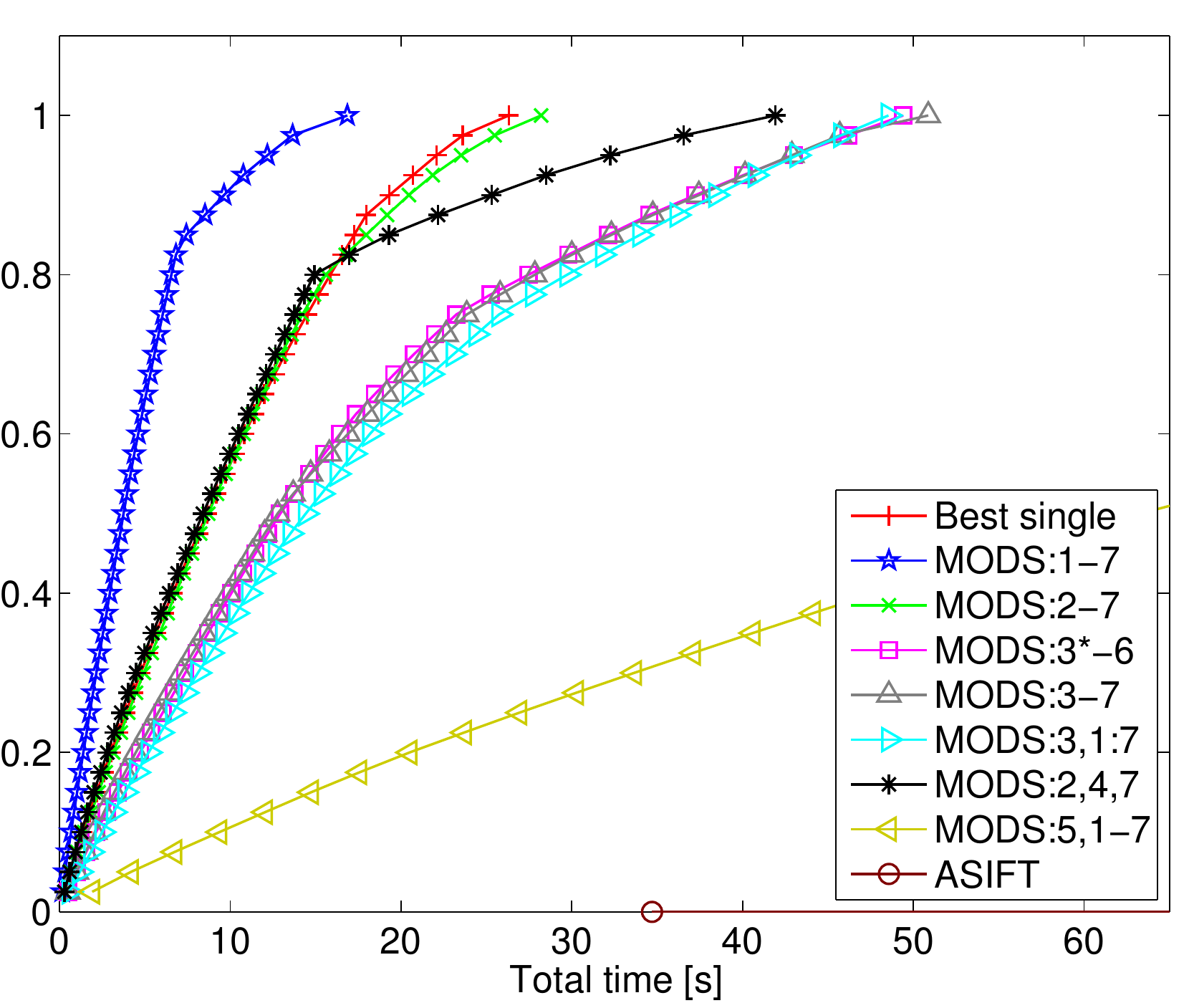}\\
\caption{Performance of MODS configurations specified in Table~\ref{tab:MODS-variants}. Cumulative percentage of image pairs successfully matched in time. Left - EVD dataset, right - Oxford dataset. The graphs are cropped. 
Each mark represents one image pair. The fastest single detector configuration which is able to match each pair was selected and plotted separately. Image considered matched if 10 correct inliers was found.}
\label{fig:MODS-variants-performance}
\end{figure*}

The MODS algorithm with more time-consuming configurations solves all image pairs and does it faster than a suitable configuration for each image pair -- see Figure~\ref{fig:MODS-variants-performance}. We have tested several variants of the MODS configurations, stated in Table~\ref{tab:MODS-variants}.
Experiments shows that the proposed MODS configuration is very fast on the easy WBS problems as in Oxford dataset (see Figure~\ref{fig:MODS-variants-performance}, right graph) and has very little overhead on the harder EVD dataset -- it is the second best after configuration without ORB steps. The results of the MODS medium configuration -- without first sparse synthesis step -- shows fruitfulness of the progressive view synthesis. 

ASIFT is able to match only 6 image pairs from the dataset. The ASIFT algorithm generates a lower number of correct inliers and works slower than our identical DoG configuration (which has the same tilt-rotation set). The main causes are the elimination of "one-to-many", including correct, correspondences, the inferiority of the standard second closest ratio matching strategy and a simple brute-force algorithm of matching used in ASIFT.

\begin{figure}[htb]
\centering
\includegraphics[width=.99\linewidth]{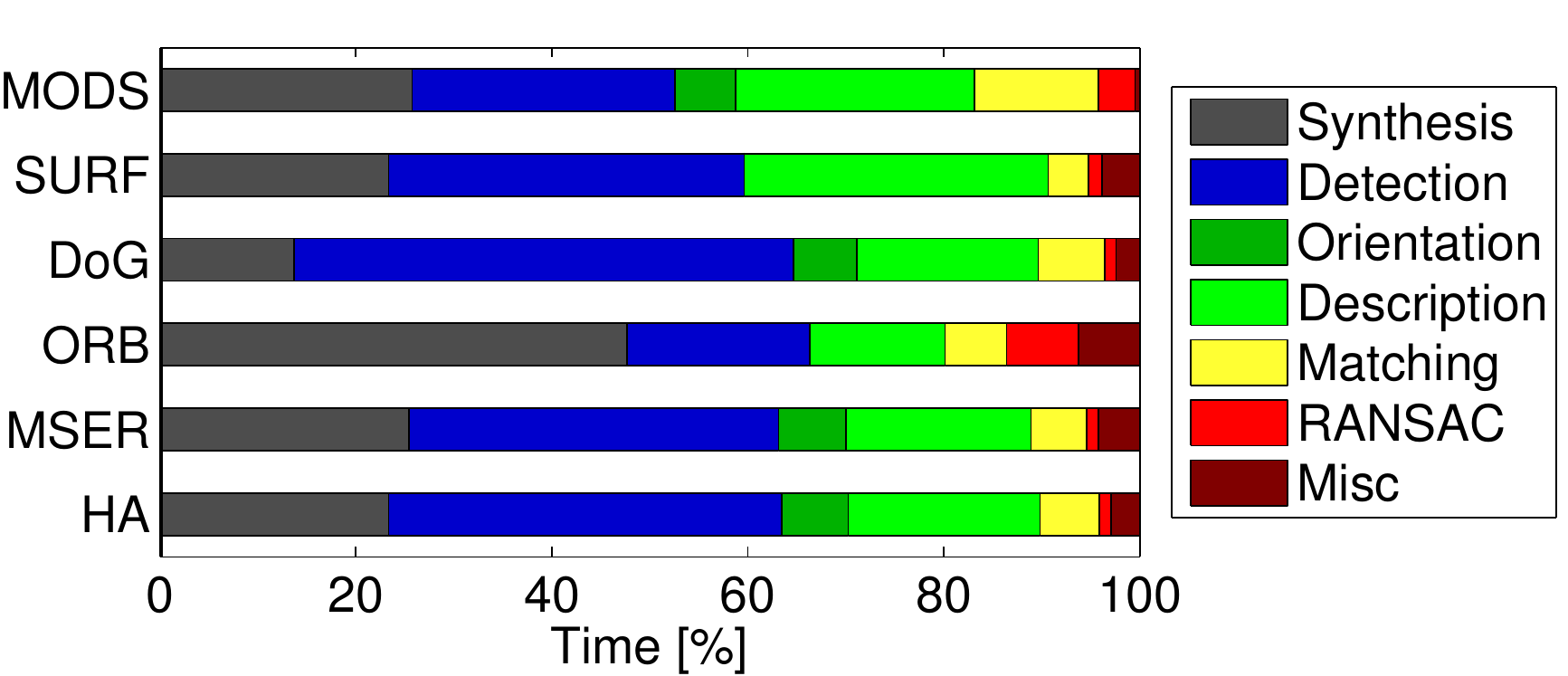} 
\caption{Percentages of time spent in the main stages of the matching with view synthesis process on a single core, {\sc easy} configuration. Detection and SIFT description, i.e. the dominant gradient estimation and the descriptor computation are the most time-consuming parts.}
 \label{fig:time-consist}
\end{figure}

Fig.~\ref{fig:time-consist} shows the breakdown of the computational time. The most time consuming parts -- detection and description (including the dominant orientation estimation)  -- take 40\%  and 35\%  resp. of the all time. Without applying the fast SIFT computation from ~\cite{Lenc2014}, the SIFT description takes more than 50\% of the time. The ORB is an exception - the synthesis is not so profitable, since it takes more time than detection and description itself. Note that the whole process is almost linear in the area of the synthesized views. The only super-linear part, matching, takes only 10\% of the time. 

\subsection{MODS testing on a non-planar dataset}
\label{main:3d-experiments}
\subsubsection{The dataset and evaluation protocol}
\label{experiments:protocol}
\begin{table}[tbp]
\caption{Reference views of the image sequences used in the evaluation (from~\cite{Moreels2007}).}
\label{tab:dataset}
\centering
\bgroup
\def\arraystretch{1.0}%
\setlength{\tabcolsep}{.5em}
\begin{tabu}spread 0pt{|X[-1m,c]|X[-1m,c]|X[-1m,c]|X[-1m,c]|}
\hline
Abroller& Bannanas & Camera2 &Car \\
\includegraphics[width=1\linewidth]{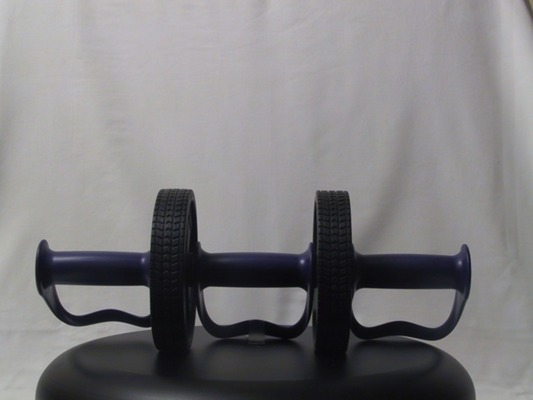} 
&\includegraphics[width=1\linewidth]{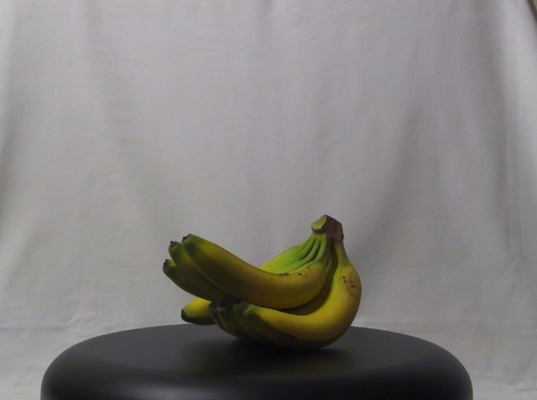} 
&\includegraphics[width=1\linewidth]{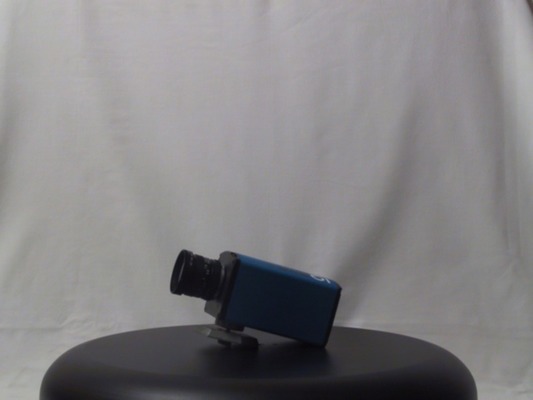} 
&\includegraphics[width=1\linewidth]{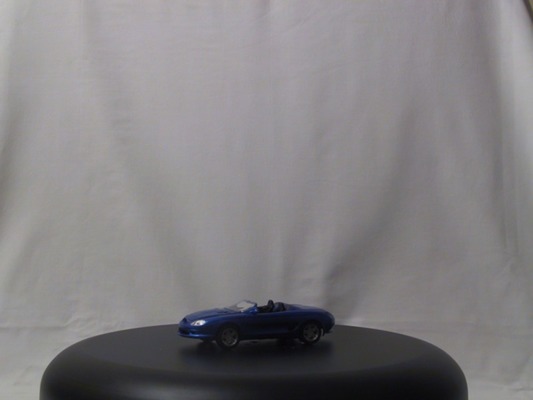}\\ 
\hline
Car2& CementBase & Cloth &Conch \\
\includegraphics[width=1\linewidth]{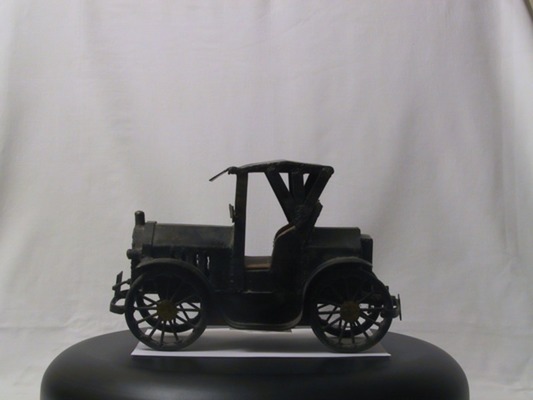} 
&\includegraphics[width=1\linewidth]{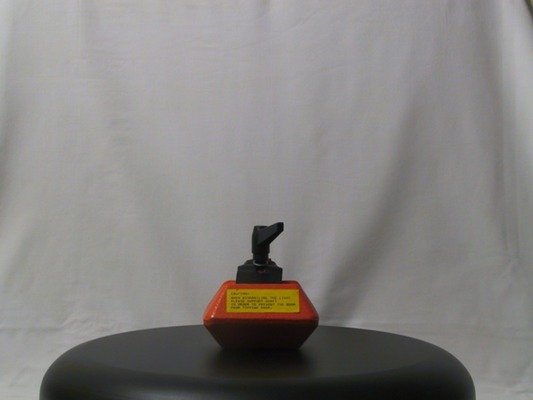} 
&\includegraphics[width=1\linewidth]{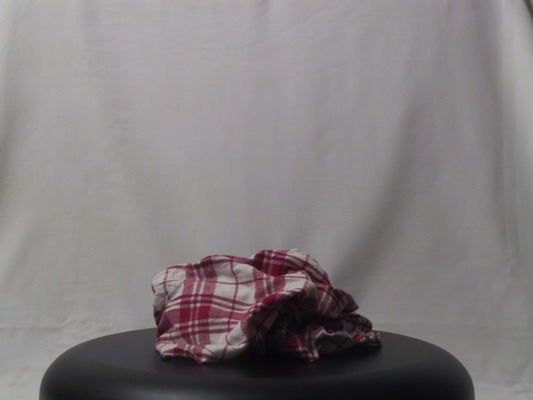} 
&\includegraphics[width=1\linewidth]{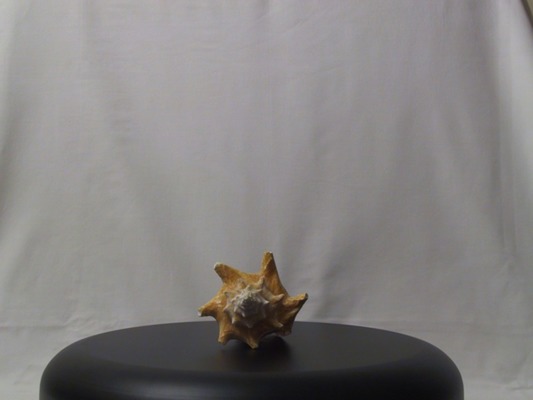}\\ 
\hline
Desk& Dinosaur & Dog &DVD \\
\includegraphics[width=1\linewidth]{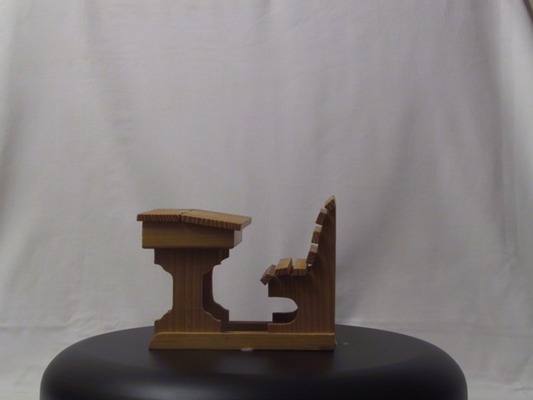} 
&\includegraphics[width=1\linewidth]{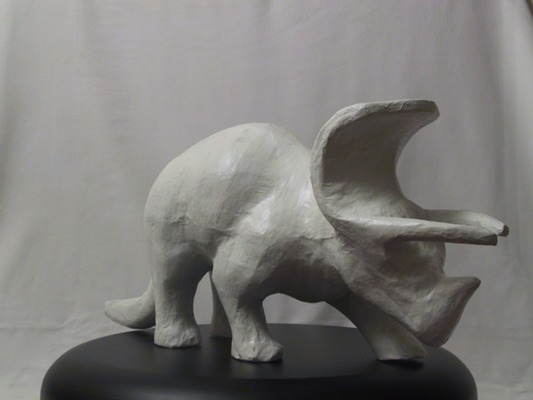} 
&\includegraphics[width=1\linewidth]{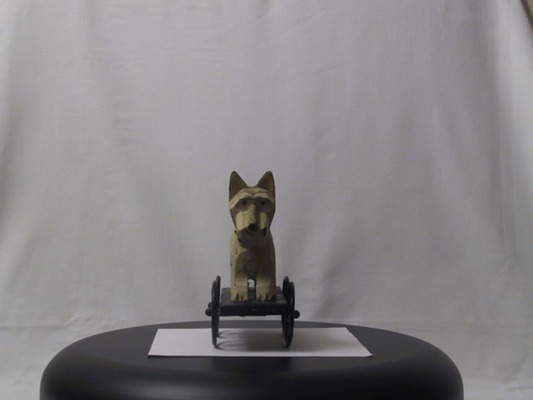} 
&\includegraphics[width=1\linewidth]{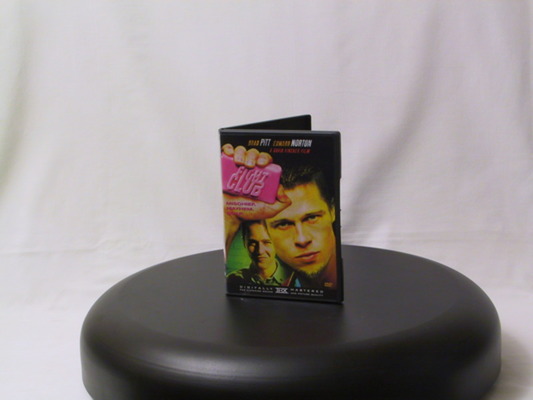}\\ 
\hline
FloppyBox& FlowerLamp & Gelsole &GrandfatherClock \\
\includegraphics[width=1\linewidth]{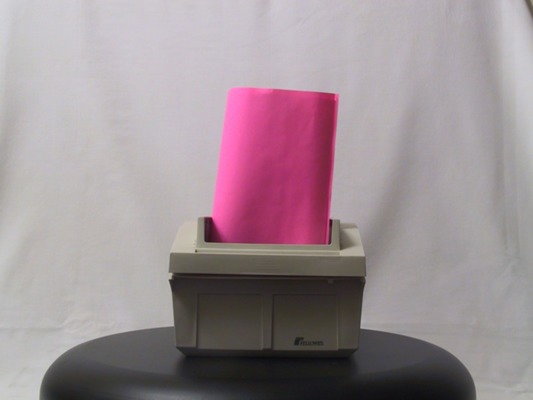} 
&\includegraphics[width=1\linewidth]{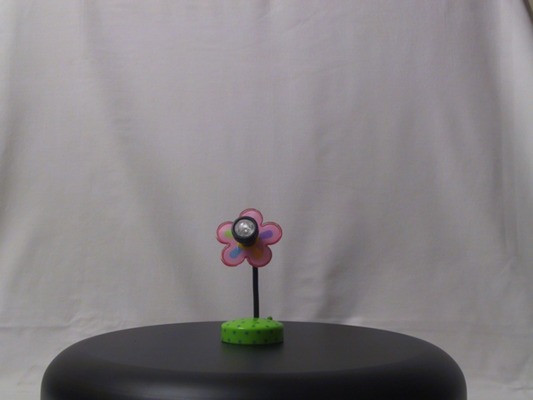} 
&\includegraphics[width=1\linewidth]{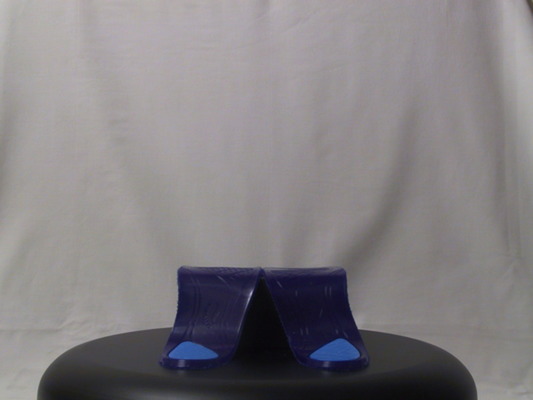} 
&\includegraphics[width=1\linewidth]{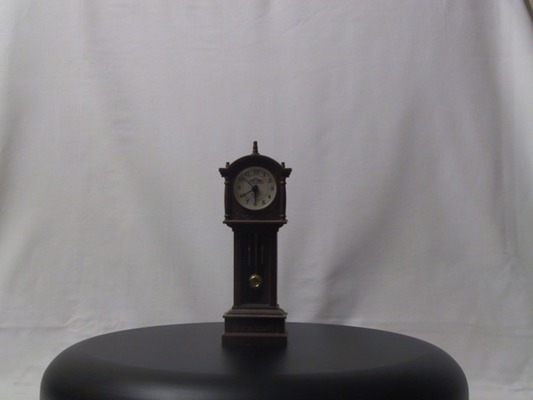}\\ 
\hline
Horse& Keyboard & Motorcycle &MouthGuard \\
\includegraphics[width=1\linewidth]{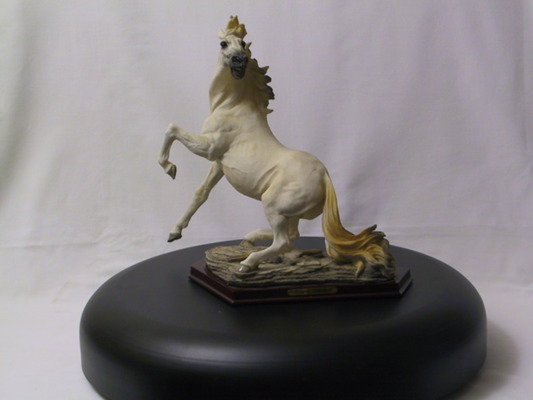} 
&\includegraphics[width=1\linewidth]{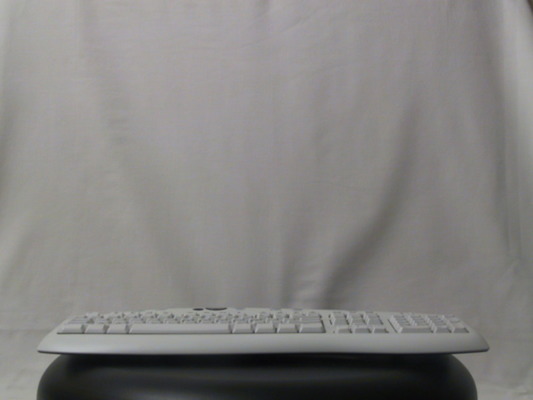} 
&\includegraphics[width=1\linewidth]{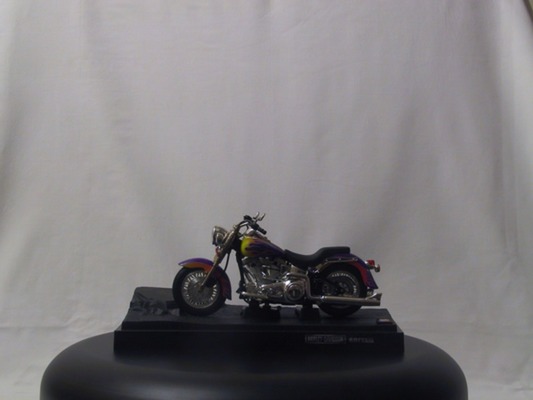} 
&\includegraphics[width=1\linewidth]{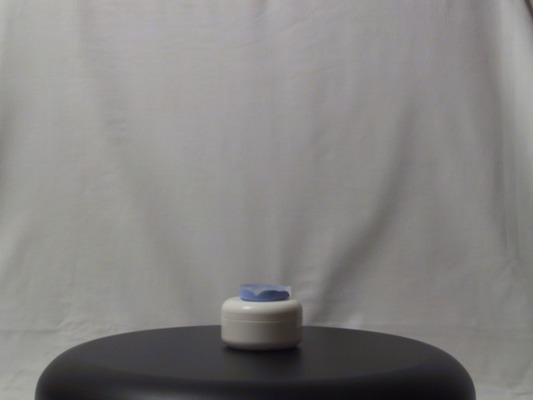}\\ 
\hline
PaperBin& PS2 & Razor &RiceCooker \\
\includegraphics[width=1\linewidth]{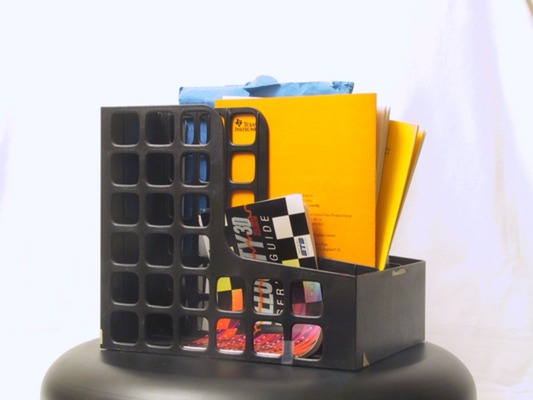} 
&\includegraphics[width=1\linewidth]{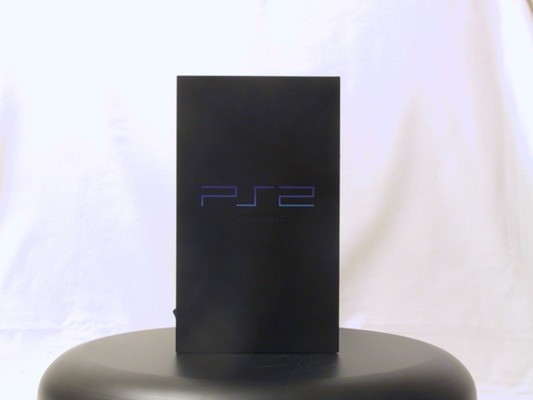} 
&\includegraphics[width=1\linewidth]{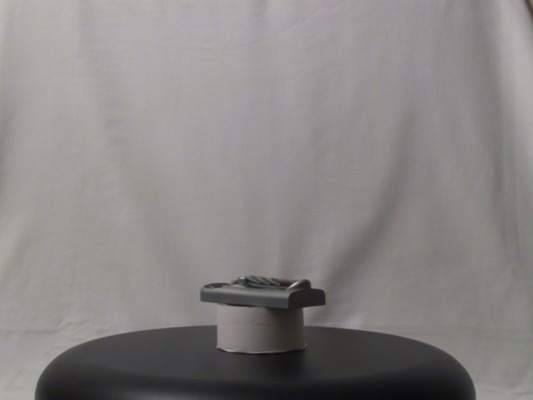} 
&\includegraphics[width=1\linewidth]{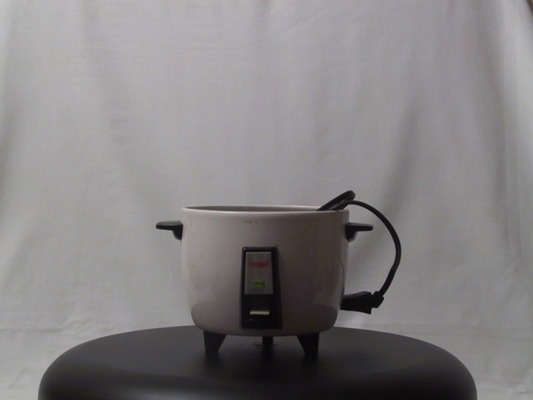}\\ 
\hline
Rock& RollerBlade & Spoons &TeddyBear \\
\includegraphics[width=1\linewidth]{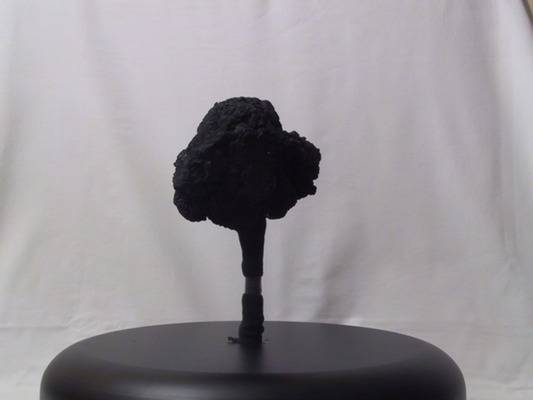} 
&\includegraphics[width=1\linewidth]{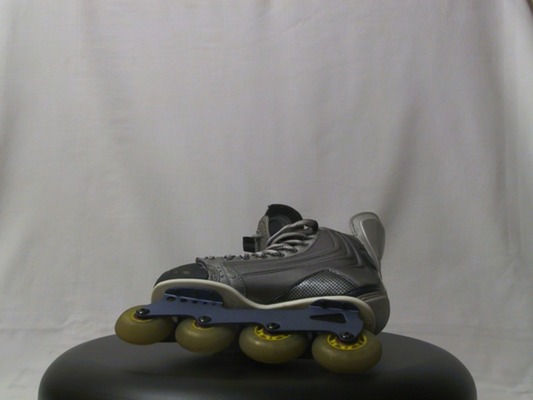} 
&\includegraphics[width=1\linewidth]{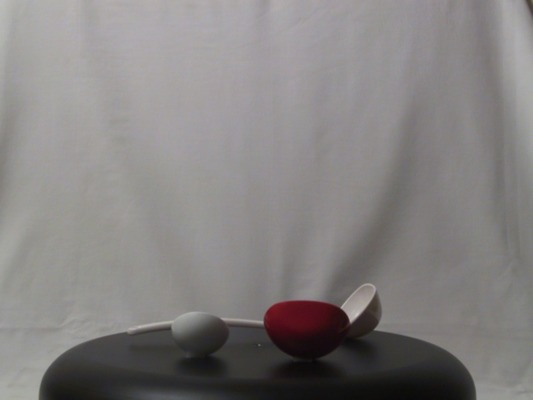} 
&\includegraphics[width=1\linewidth]{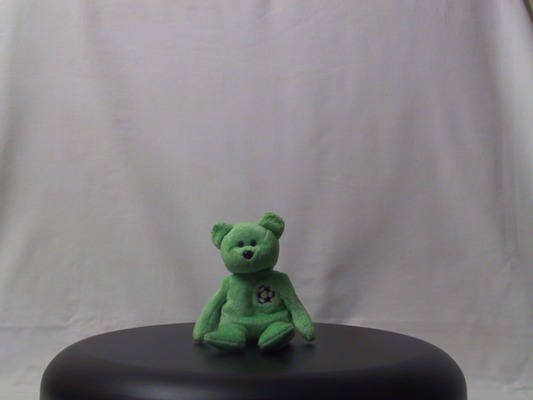}\\ 
\hline
Toothpaste& Tricycle & Tripod &VolleyBall \\
\includegraphics[width=1\linewidth]{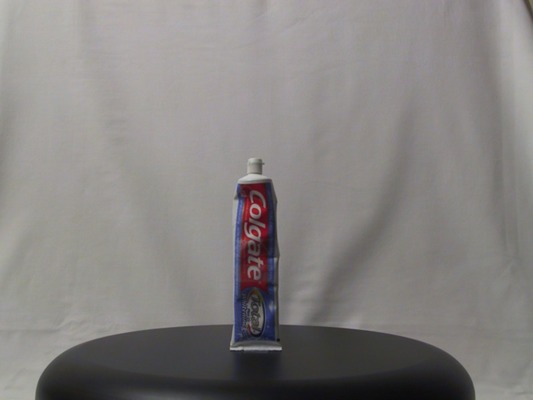} 
&\includegraphics[width=1\linewidth]{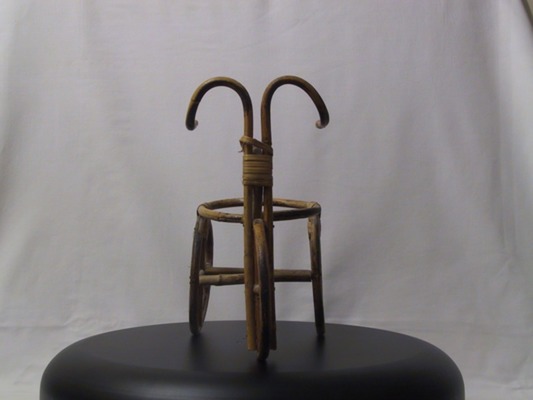} 
&\includegraphics[width=1\linewidth]{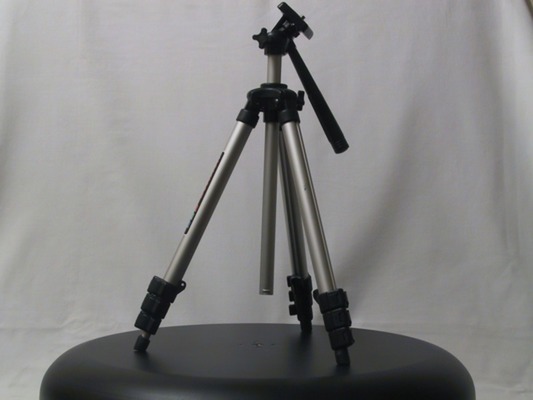} 
&\includegraphics[width=1\linewidth]{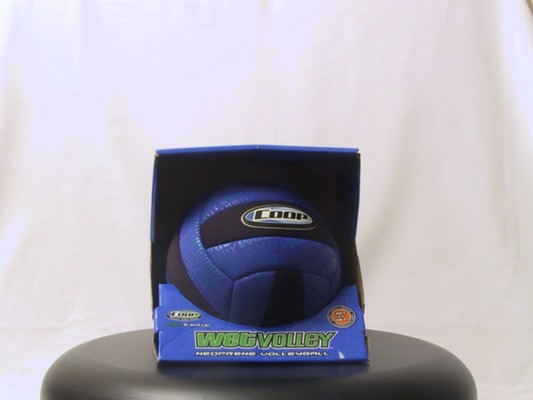}\\ 
\hline
\end{tabu}
\egroup
\end{table}
The evaluation dataset consists of 35 image sequences taken from the Turntable dataset~\cite{Moreels2007} (``Bottom'' camera) shown in Table~\ref{tab:dataset}. Eight image sets contain objects with relatively large planar surfaces and the remaining ones are low-textured, ``general  3D'' objects.

The view marked as ``$0^{\circ}$'' in the Turntable dataset was used as a reference view and $0-90^\circ$ and 270-355~$\degree$ views  with a $5^\circ$ step were matched against it using the procedure described in Sec.~\ref{sec:main}, forming a $[-90^{\circ},90^{\circ}]$ sequence. Note that the reference view is not usually the ``frontal'' or ``side'' view, but rather some intermediate view which caused asymmetry in results (see Fig.~\ref{fig:jet-angle}, Table~\ref{tab:results}).

The output of the matchers is a set of the correspondences and the estimated geometrical transformation. The accuracy of the matched correspondences was chosen as the performance criterion, similarly to the protocol in~\cite{Dahl2011}.
For all output correspondences, the symmetrical epipolar error~\cite{Hartley2004} $e_{\text{SymEG}}$ was computed according to the following expression:
\begin{equation}
\label{eq:symmetric-error-F}
e_{\text{SymEG}}\left(\mathbf{F},\mathbf{u},\mathbf{v}\right)=\left(\mathbf{v}^{\top}\mathbf{F}\mathbf{u}\right)^2 \times \left(\frac{1}{(\mathbf{F}\mathbf{u})^2_1 + (\mathbf{F}\mathbf{u})^2_2} +
\frac{1}{(\mathbf{F}^{\top}\mathbf{v})^2_1 + (\mathbf{F}^{\top}\mathbf{v})^2_2}\right),
\end{equation}
where $\mathbf{F}$ -- fundamental matrix, $\mathbf{u}$, $\mathbf{v}$ -- corresponding points, $(\mathbf{F}\mathbf{u})^2_j$ -- the square of the $j$-th entry of the vector $\mathbf{F}\mathbf{u}$.

The ground truth fundamental matrix was obtained from the difference in camera positions~\cite{Hartley2004}, assuming that turntable is fixed and the camera moved around the object, according to the following equation: 
\begin{multline}
\label{eq:fundamental-matrix}
F=K^{-\top}RK^{\top}[KR^{\top}t]_\times,\\
R=\left( \begin{array}{ccc}
\cos{\phi} & 0 &-\sin{\phi} \\
0&1&0\\
\sin{\phi} &0& \cos{\phi}
\end{array} \right), 
K=\left( \begin{array}{ccc}
\frac{mf}{\text{FR}_X} & 0 & \frac{m}{2} \\
0& \frac{nf}{\text{FR}_Y}  & \frac{n}{2} \\
0 &0& 1
\end{array} \right),
t=r\left( \begin{array}{ccc}
\sin{\phi} \\
0\\
1-\cos{\phi}
\end{array} \right) ,
\end{multline}
where $R$ is the orientation matrix of the second camera, $K$ -- the camera projection matrix, $t$ -- the virtual translation of the second camera, $r$ -- the distance from camera to the object, $\phi$ -- the viewpoint angle difference, $\text{FR}_X,\text{FR}_Y$ -- the focal plane resolution, $f$ -- the focal length, $m$, $n$ -- the sensor matrix width and height in pixels. The last five parameters were obtained from EXIF data. 

One of the evaluation problems is that background regions, i.e.~regions that are not on the object placed on the turntable, are often detected and matched influencing the geometry transformation estimation. The matches are correct, but consistent with an identity transform of the (background of) the test images, not the fundamental matrix associated with the movement of the object on the turntable. In order to solve this problem, the median value of the correspondence errors was chosen as the measure of precision because of its tolerance to the low number of outliers (e.g.~the above-mentioned background correspondences), and its sensitivity to the incorrect geometric model estimated by RANSAC. 

An image pair is considered as \emph{correctly} matched if the median symmetrical epipolar error on the correspondences using ground truth fundamental matrix is $\leq$~6~pixels.
\subsubsection{Results}
\label{experiments:results}
\begin{figure}[htbp]
\centering
\includegraphics[width=0.32\linewidth]{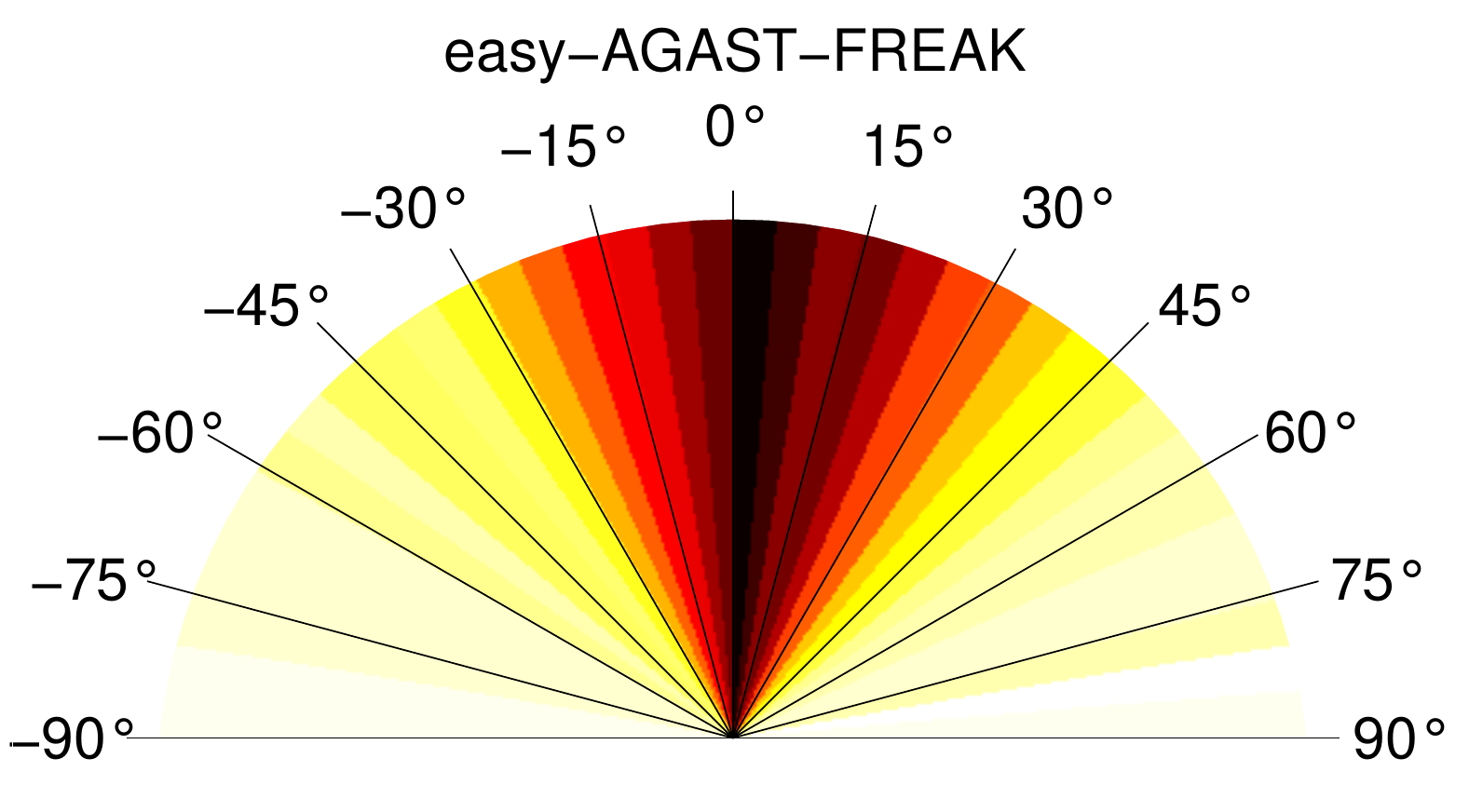}
\includegraphics[width=0.32\linewidth]{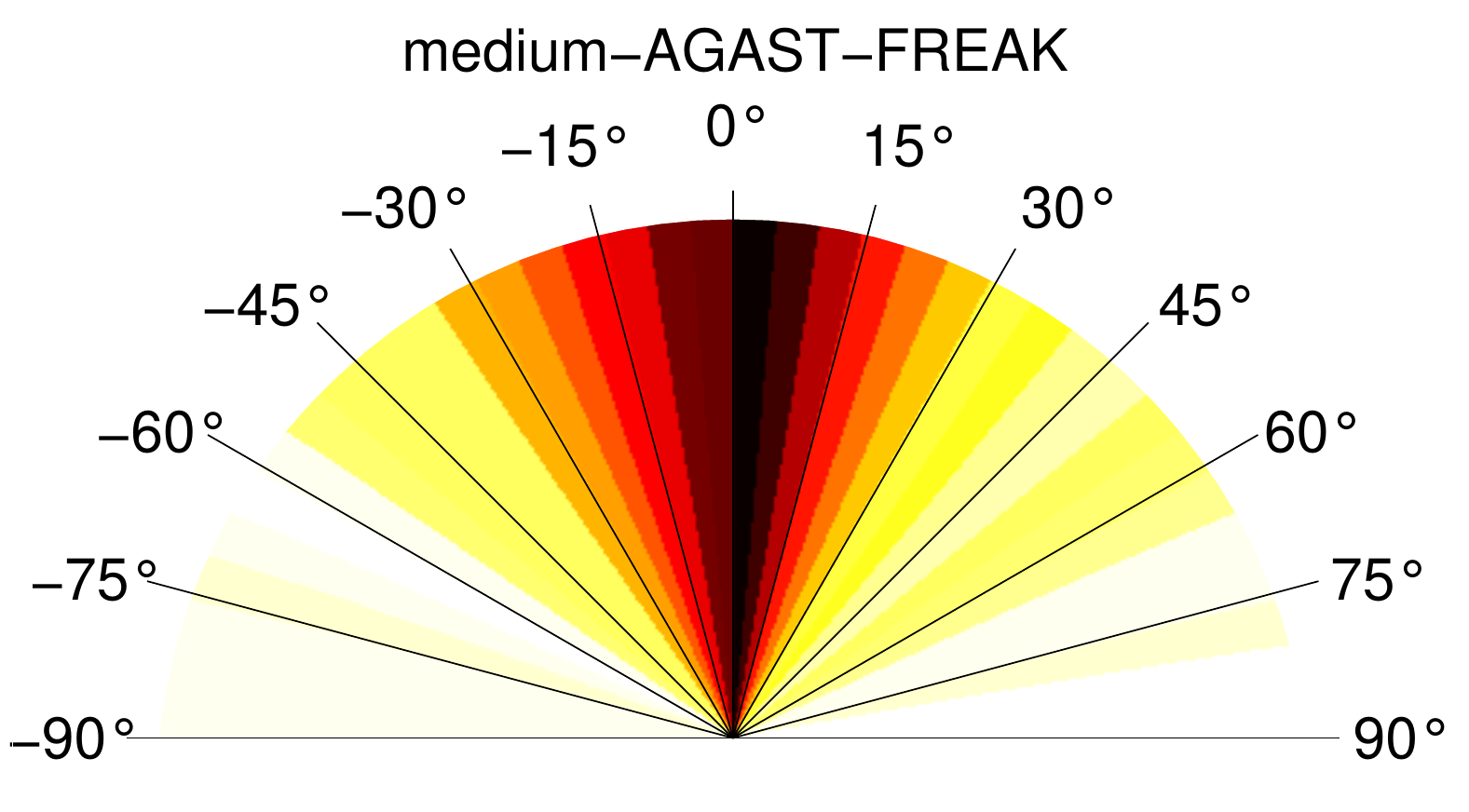}
\includegraphics[width=0.34\linewidth]{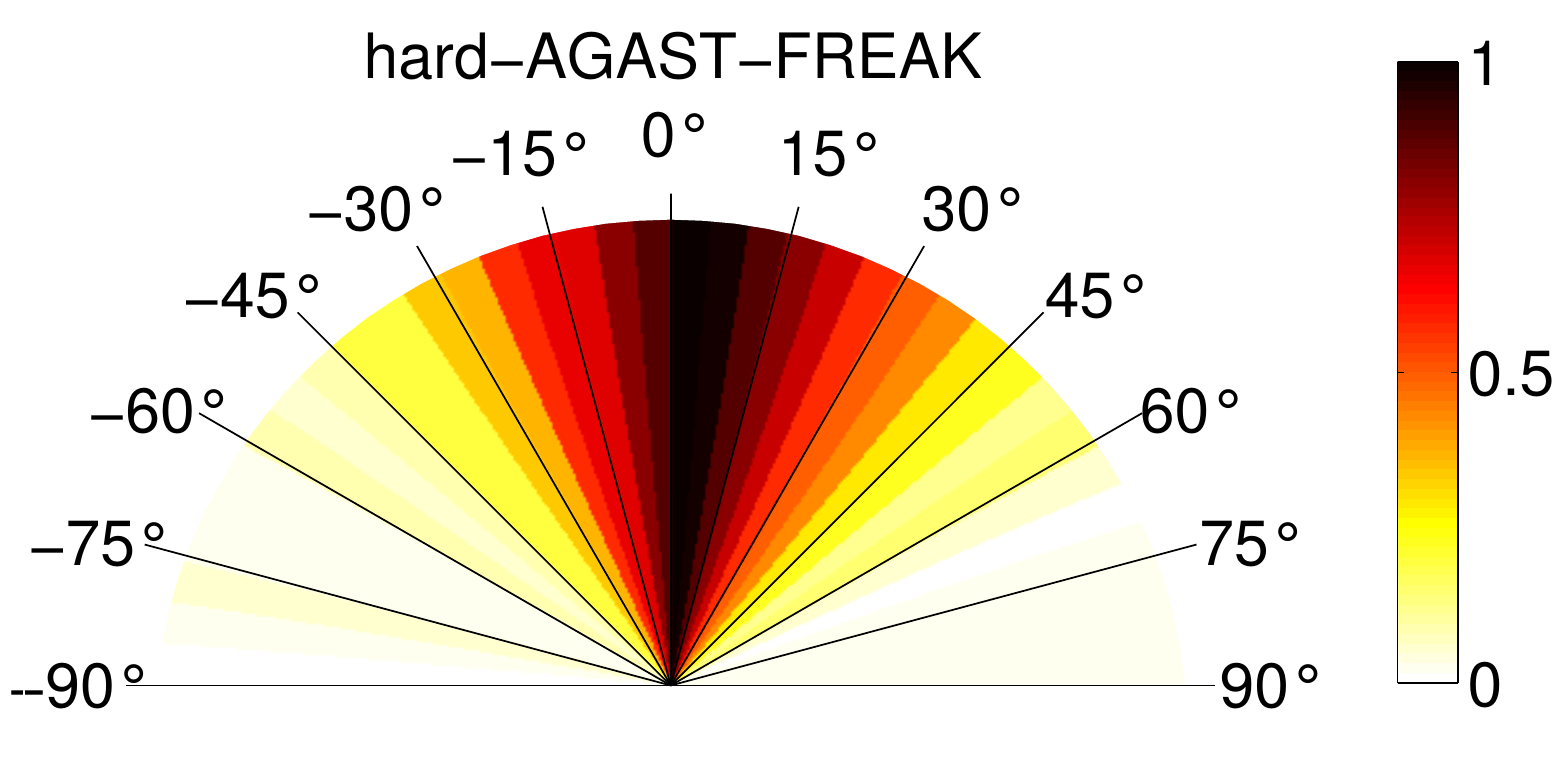}\\
\includegraphics[width=0.32\linewidth]{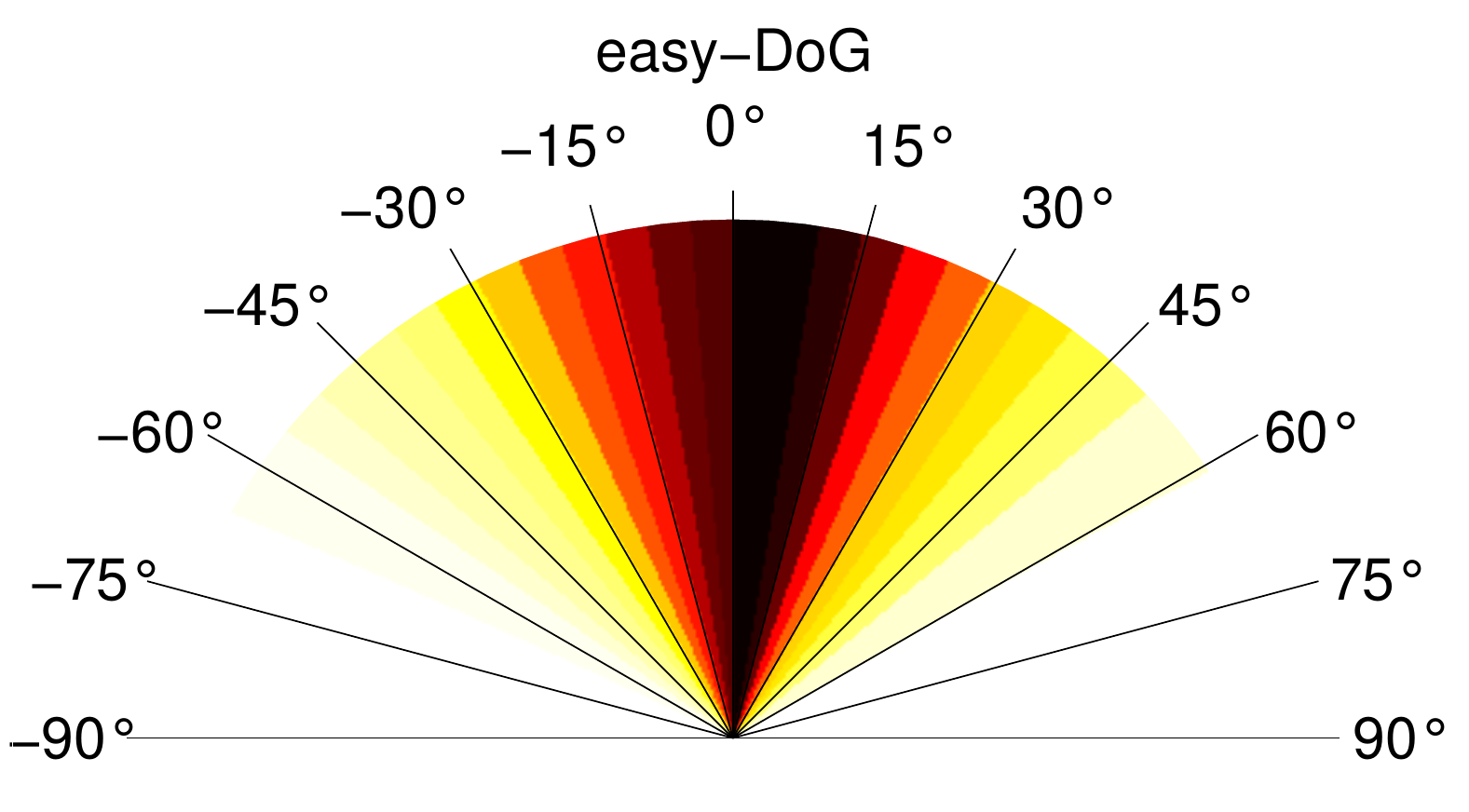}
\includegraphics[width=0.32\linewidth]{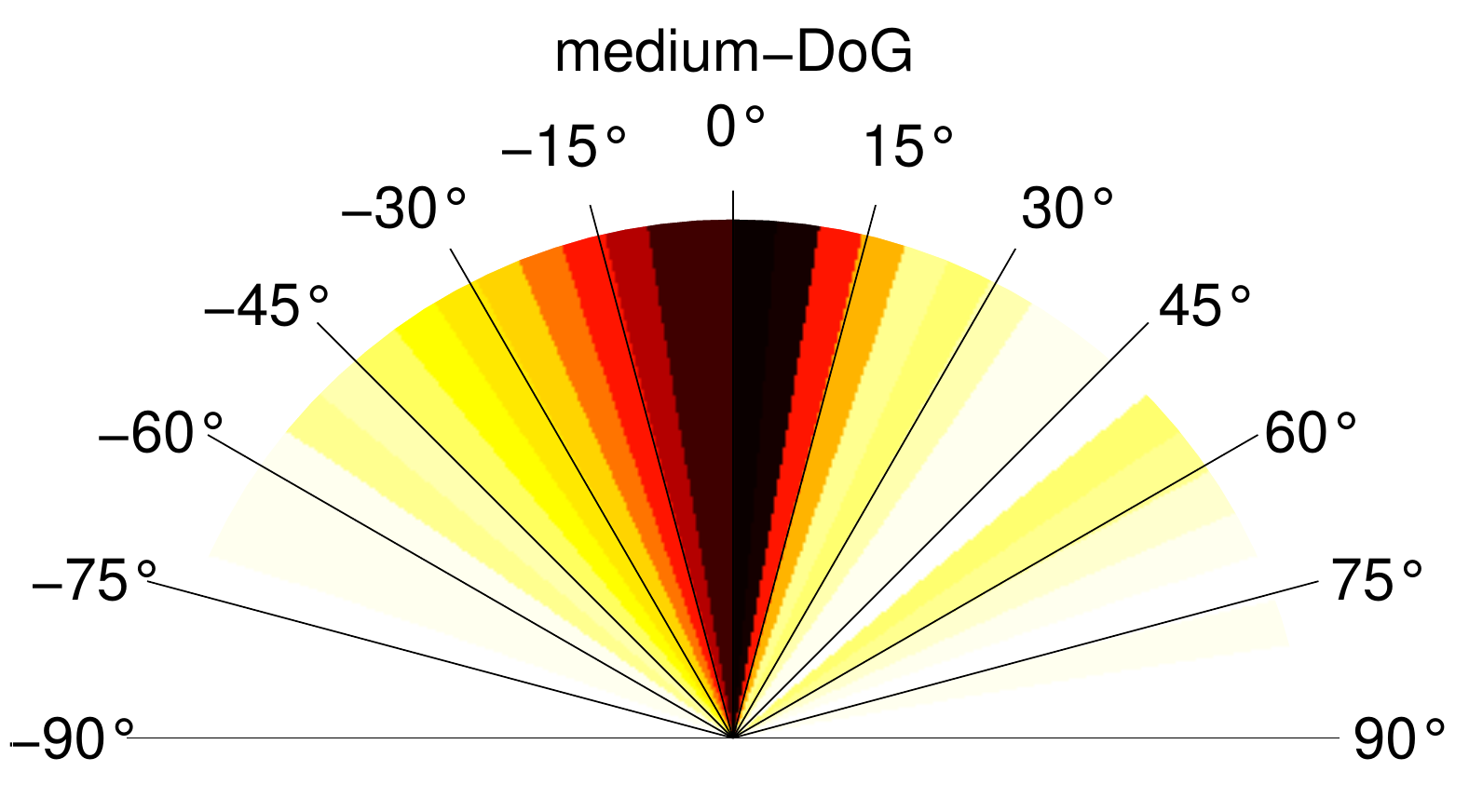}
\includegraphics[width=0.34\linewidth]{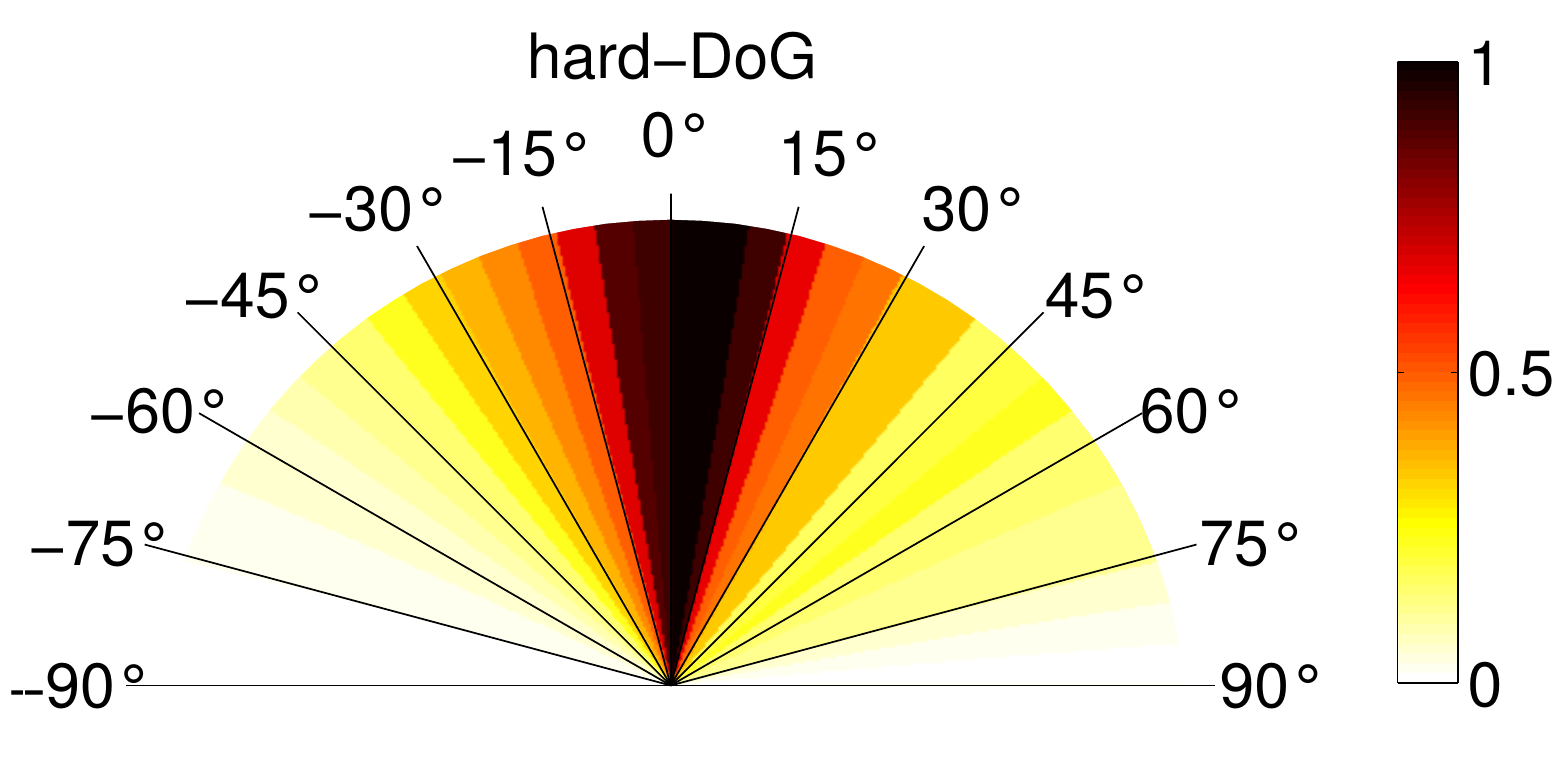}\\
\includegraphics[width=0.32\linewidth]{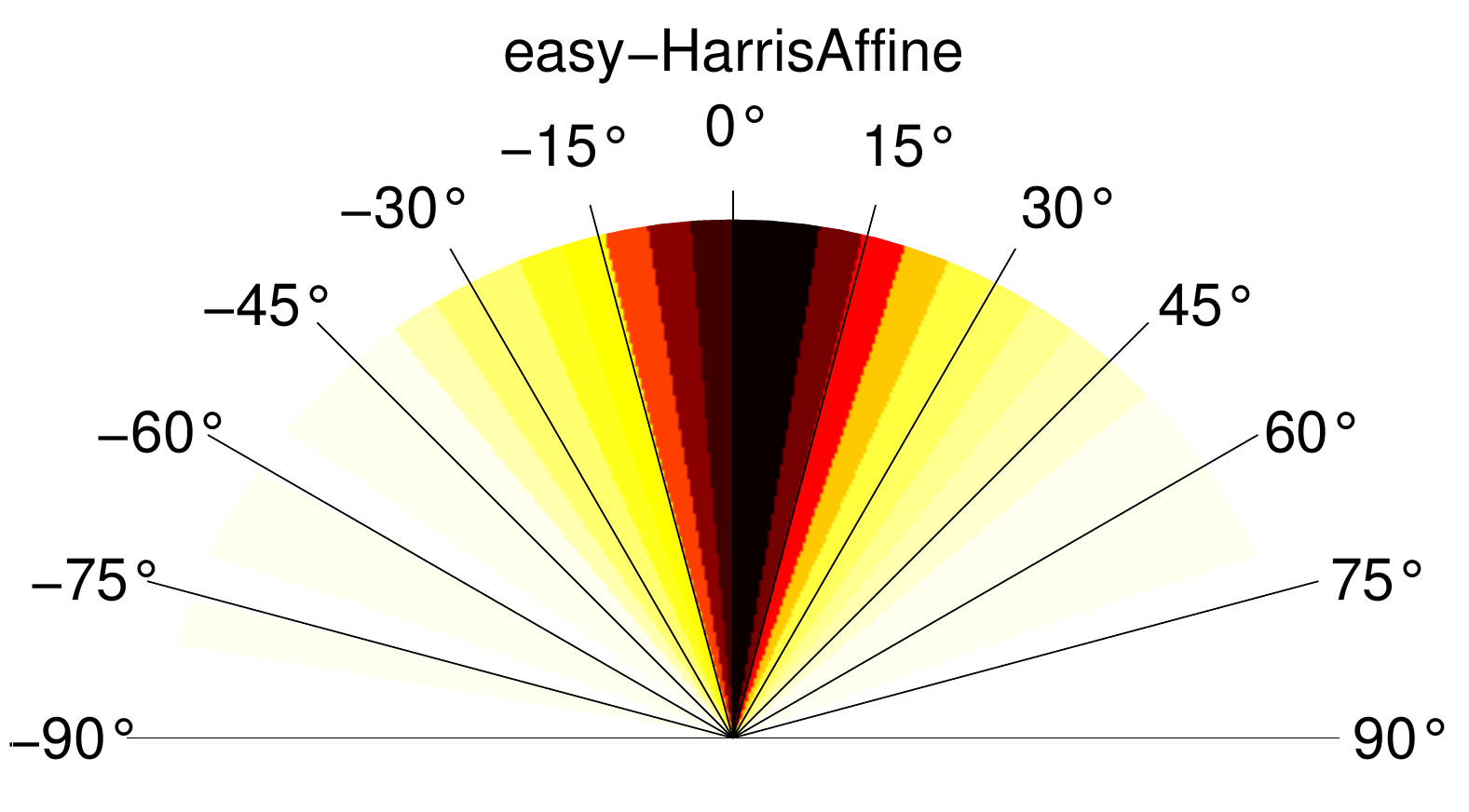}
\includegraphics[width=0.32\linewidth]{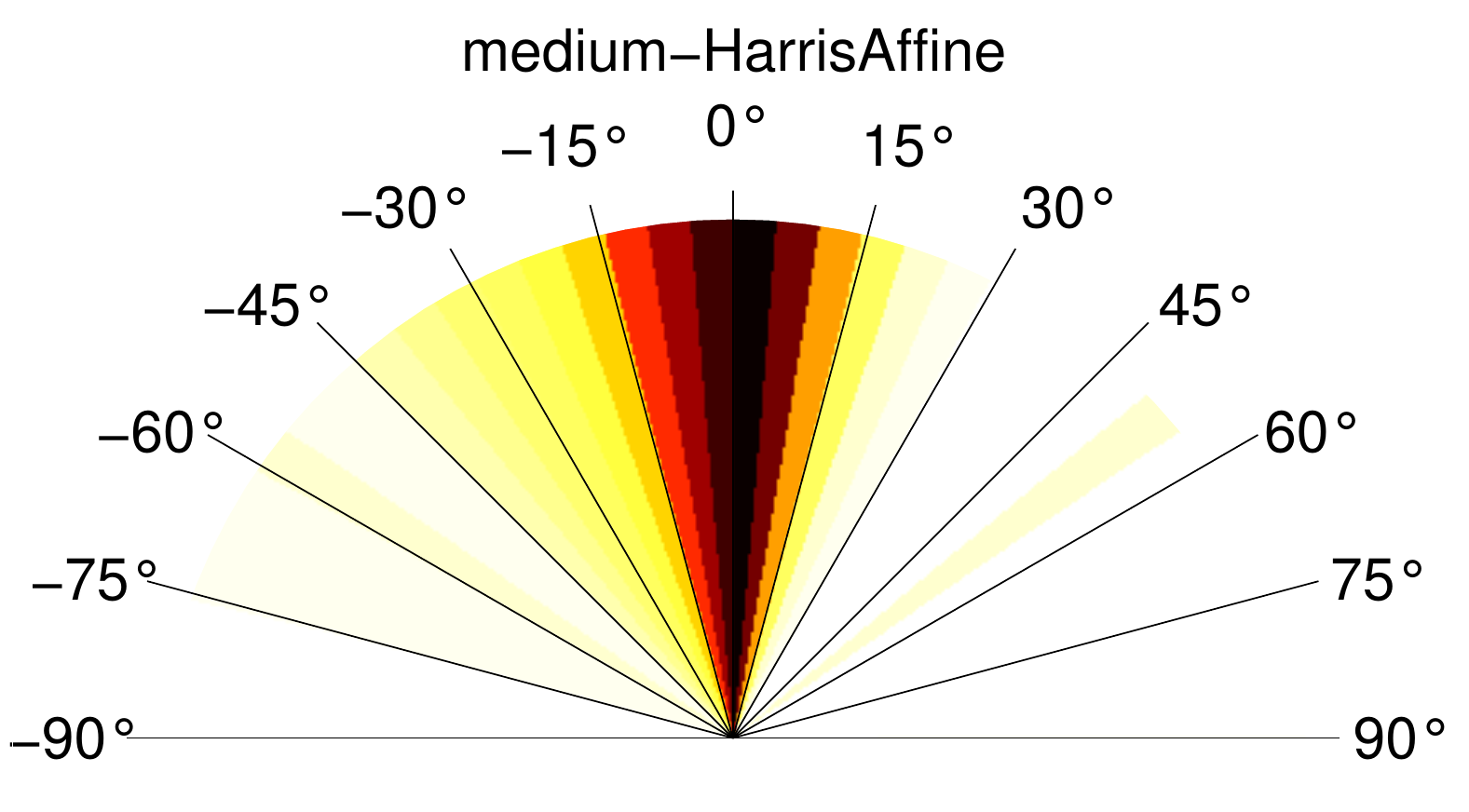}
\includegraphics[width=0.34\linewidth]{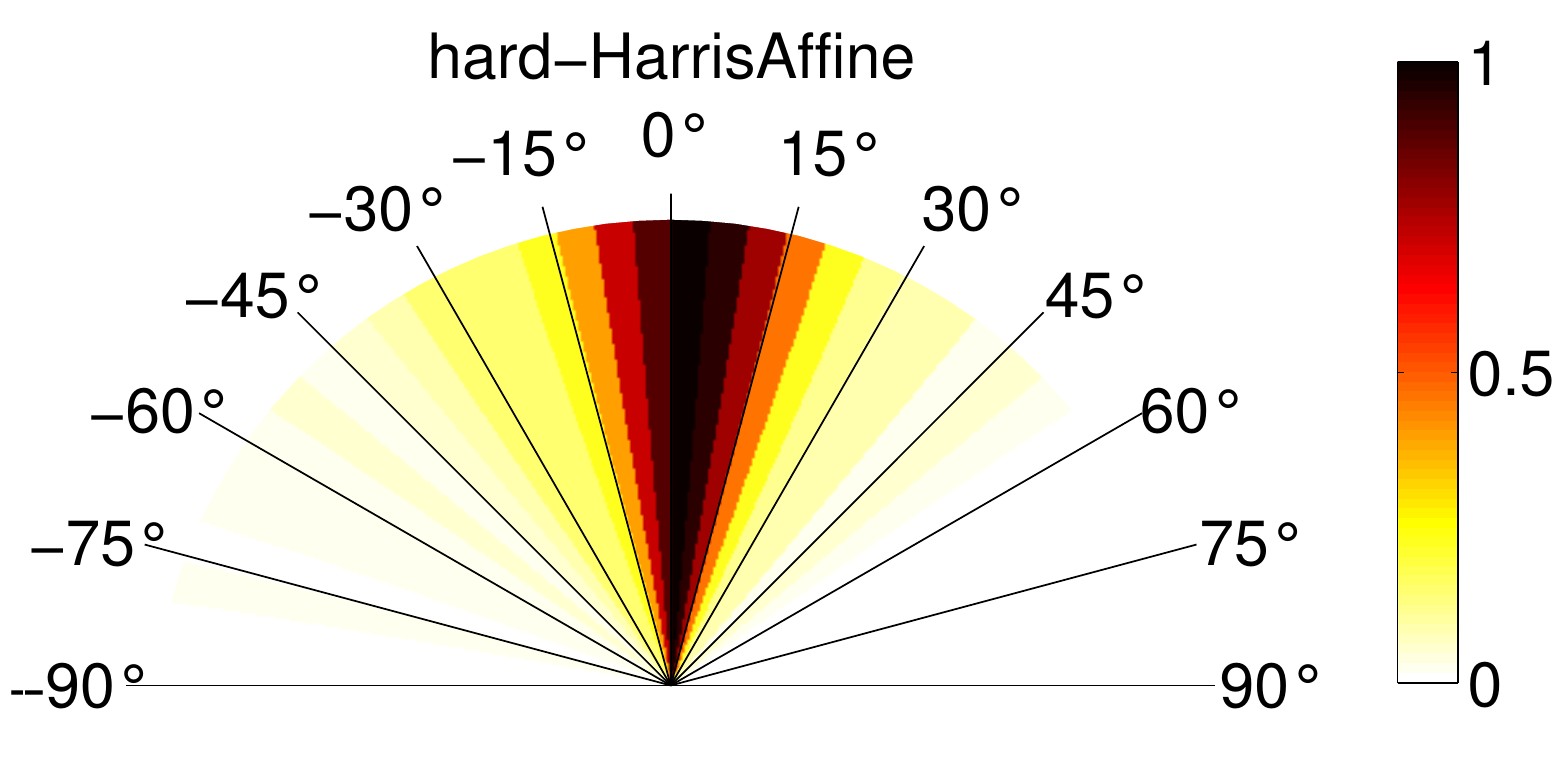}\\
\includegraphics[width=0.32\linewidth]{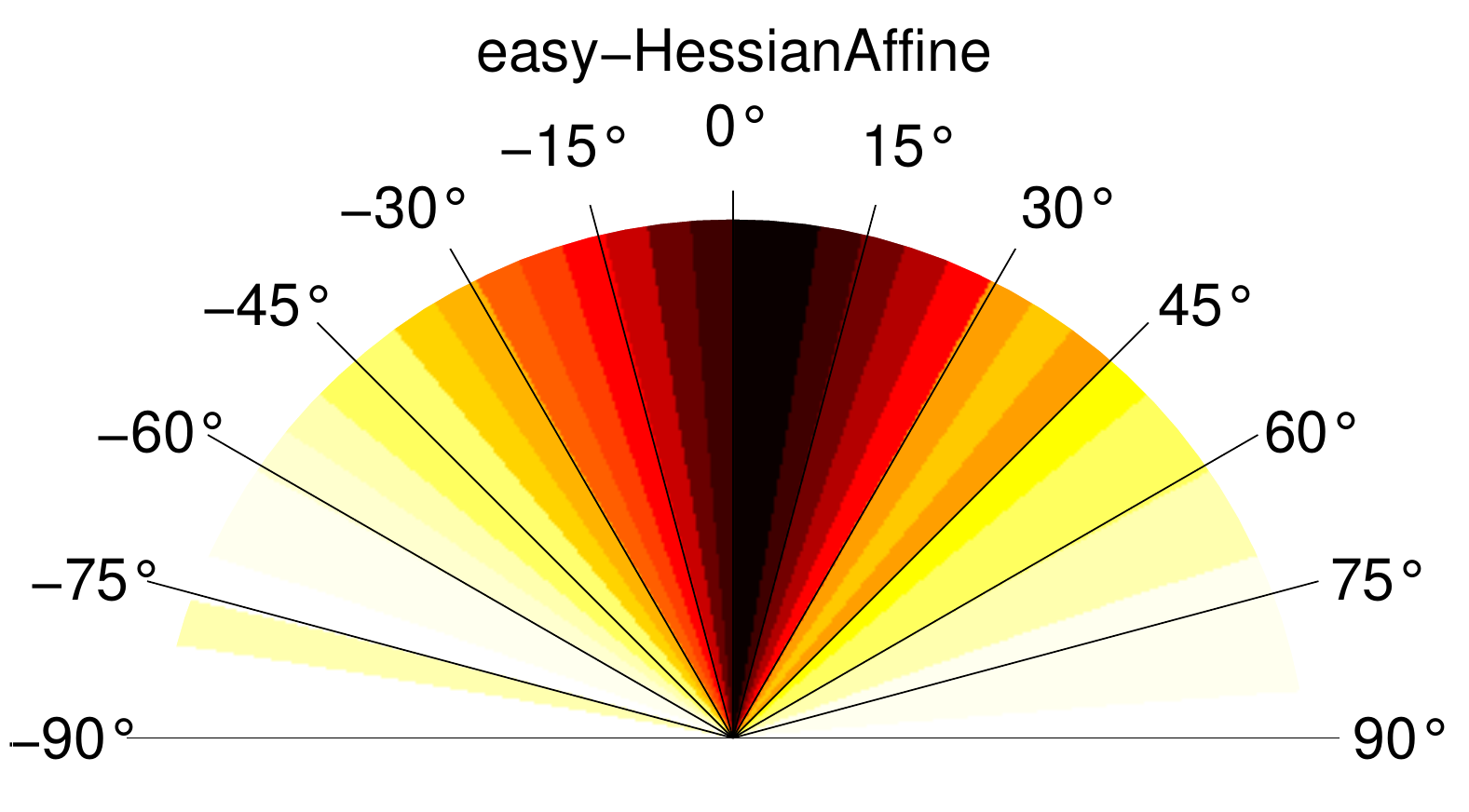}
\includegraphics[width=0.32\linewidth]{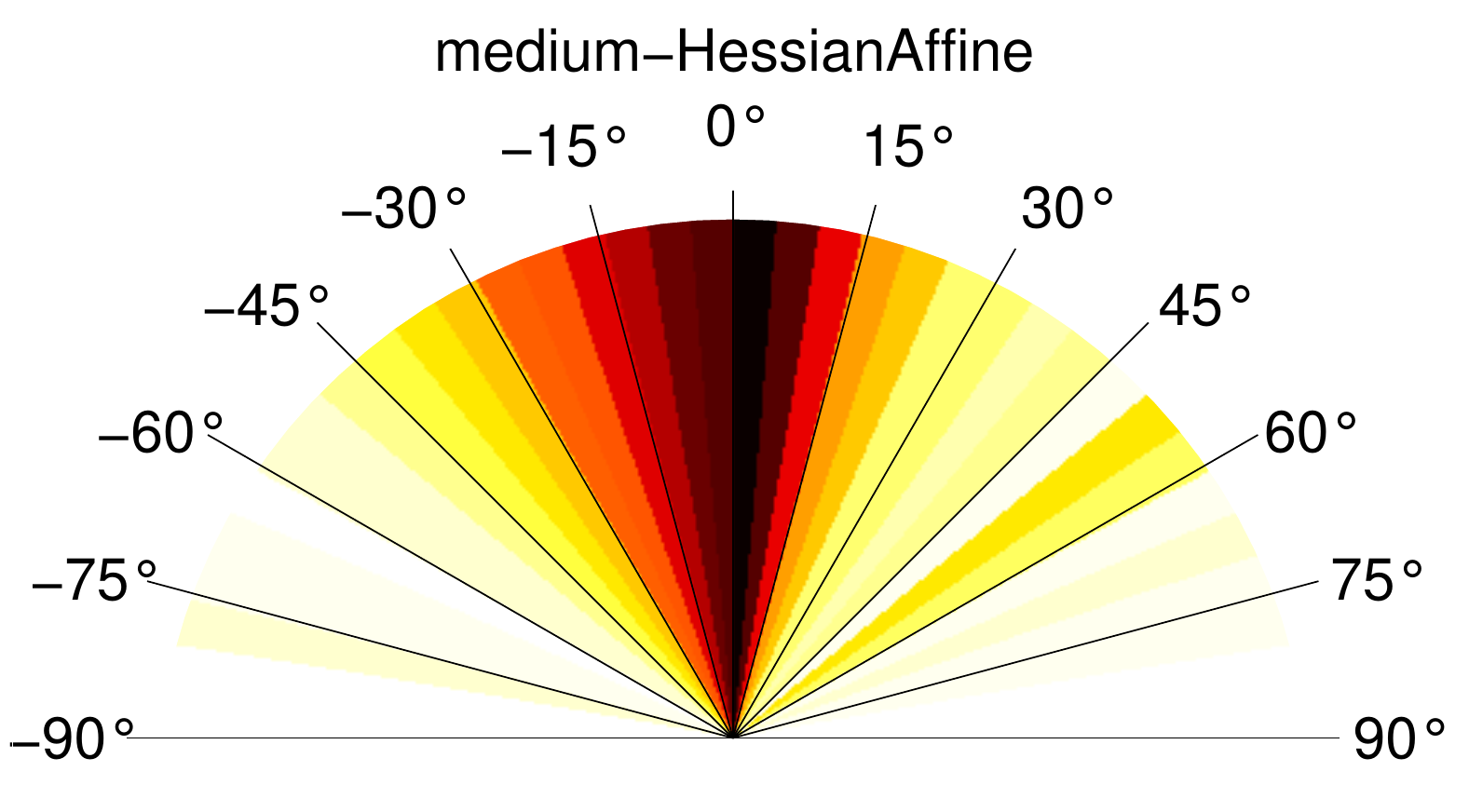}
\includegraphics[width=0.34\linewidth]{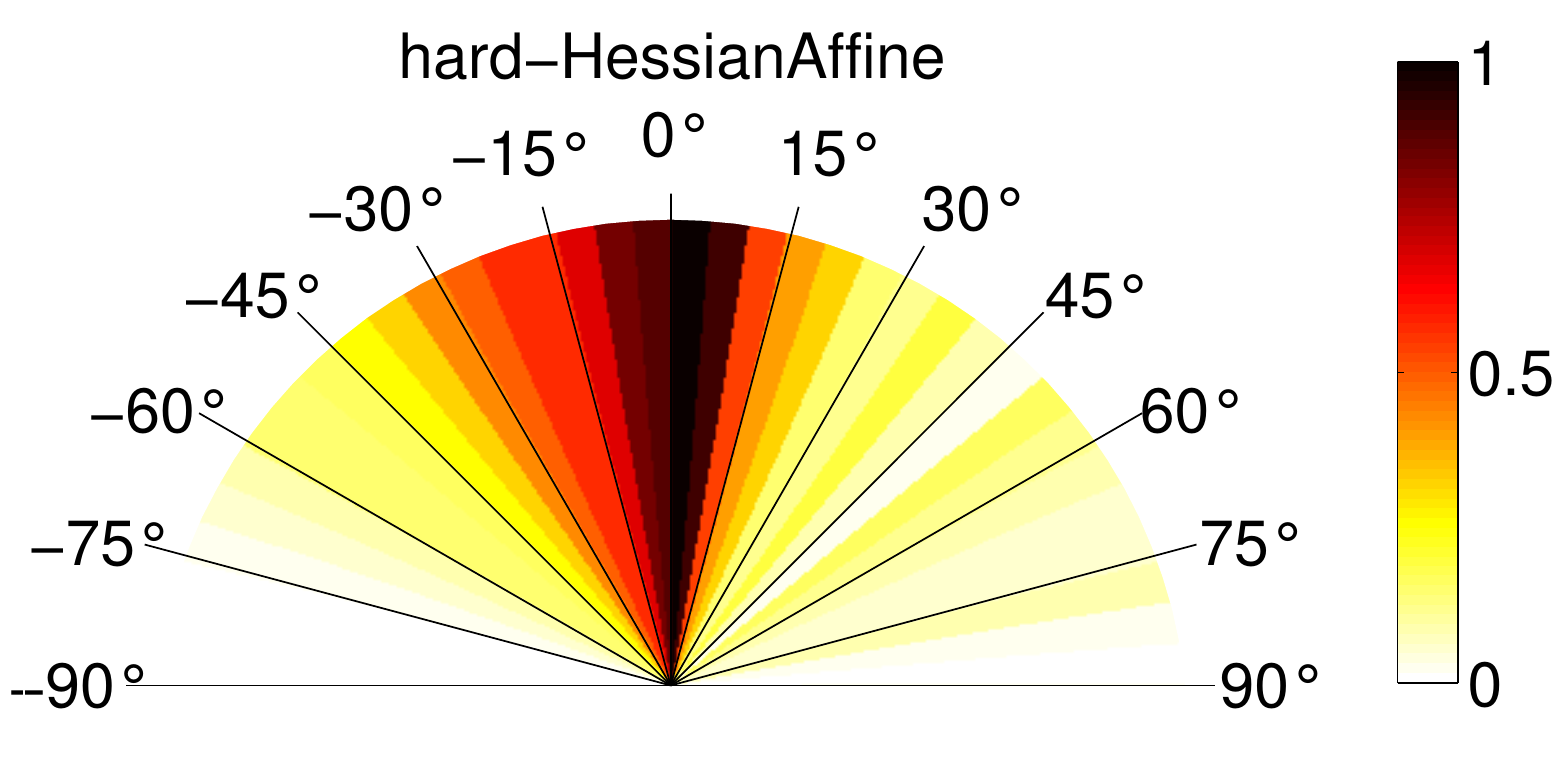}\\
\includegraphics[width=0.32\linewidth]{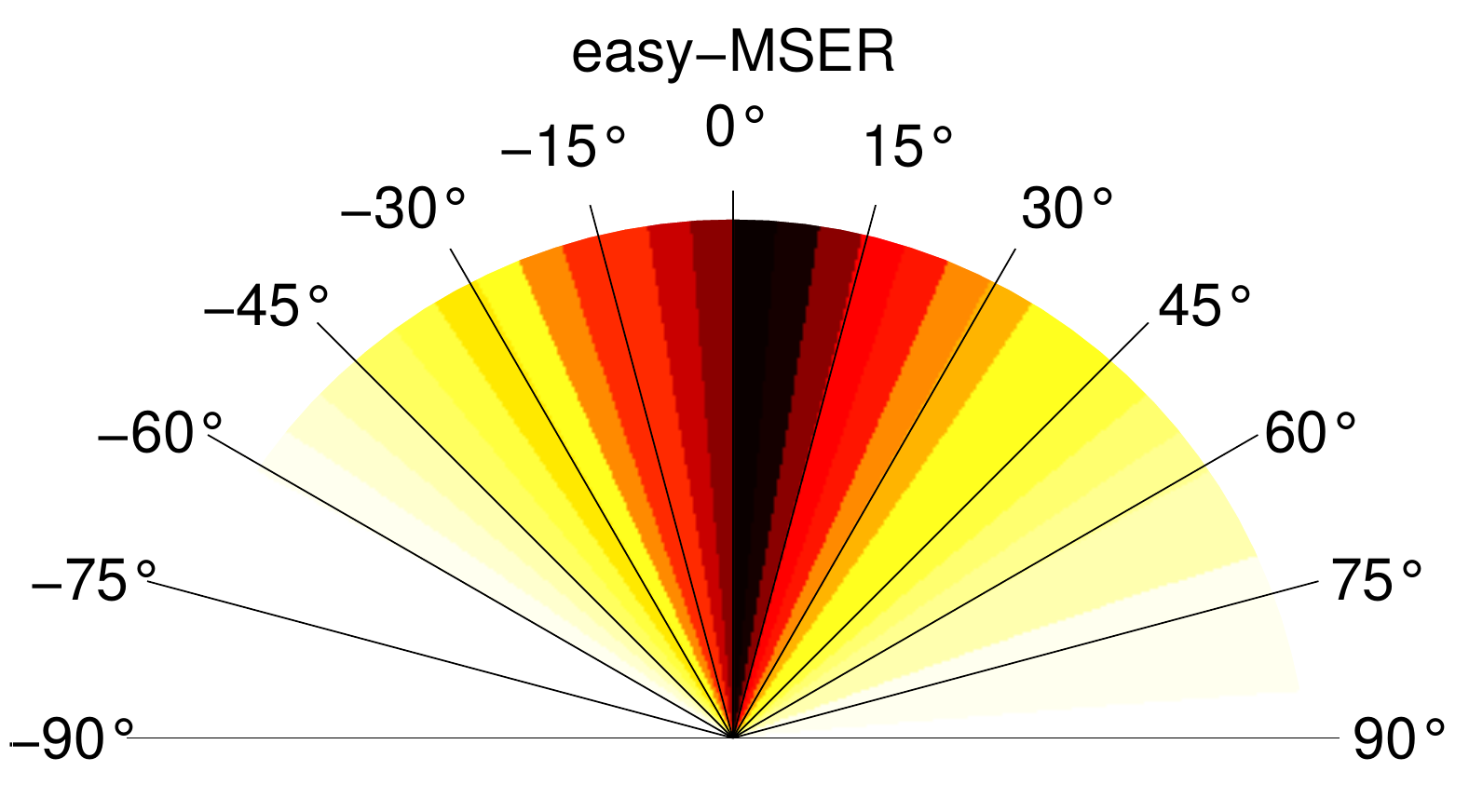}
\includegraphics[width=0.32\linewidth]{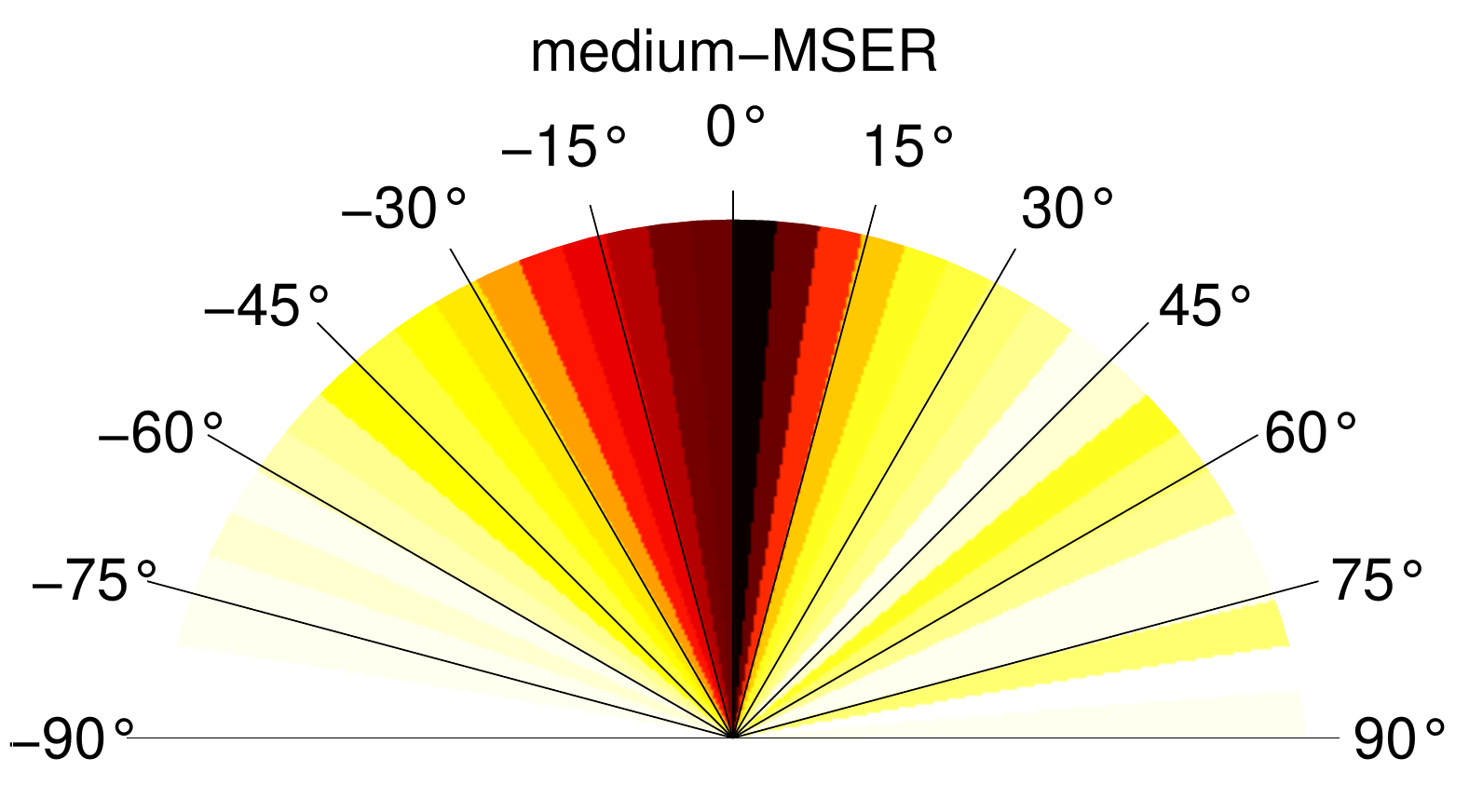}
\includegraphics[width=0.34\linewidth]{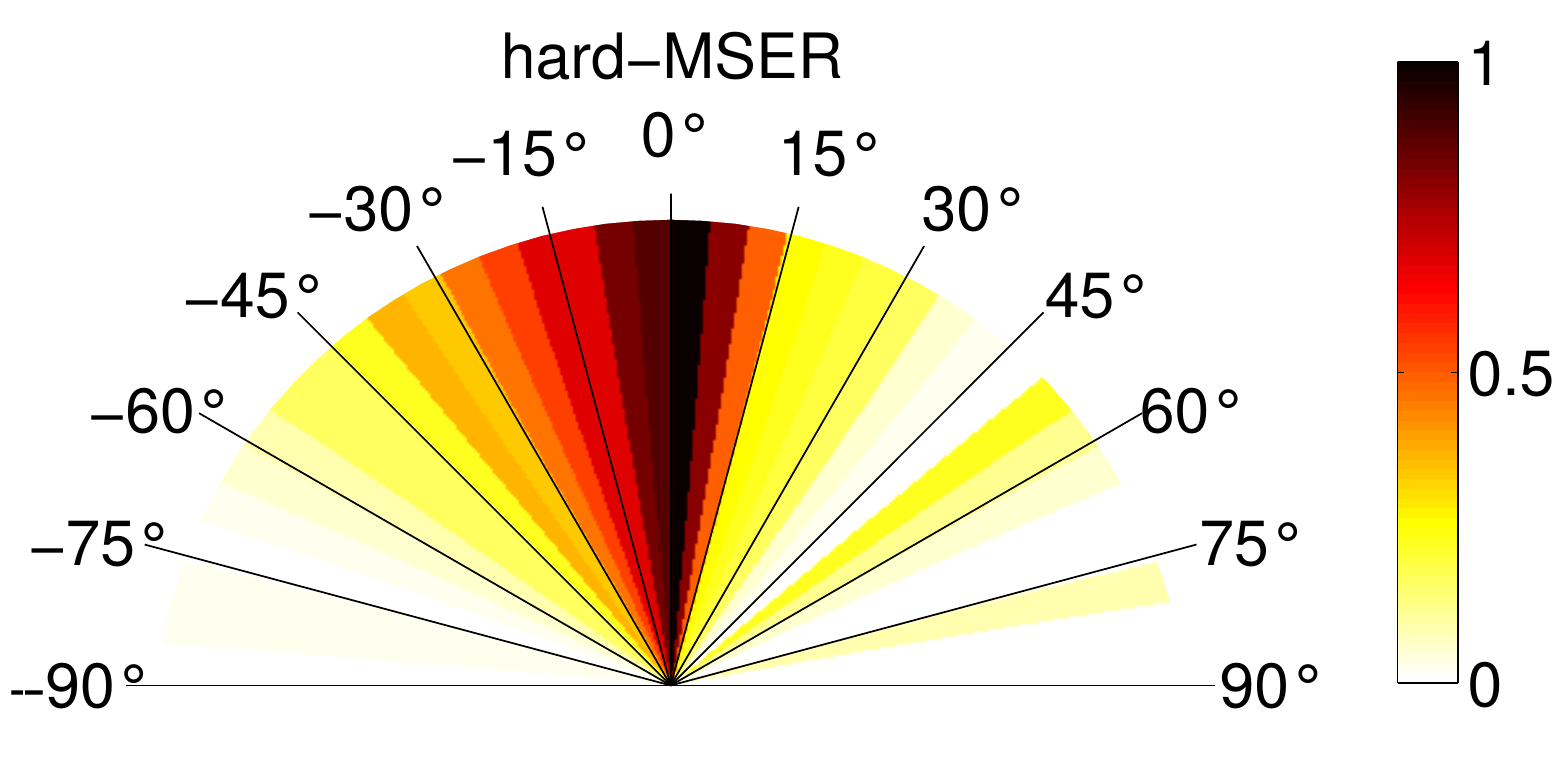}\\
\includegraphics[width=0.32\linewidth]{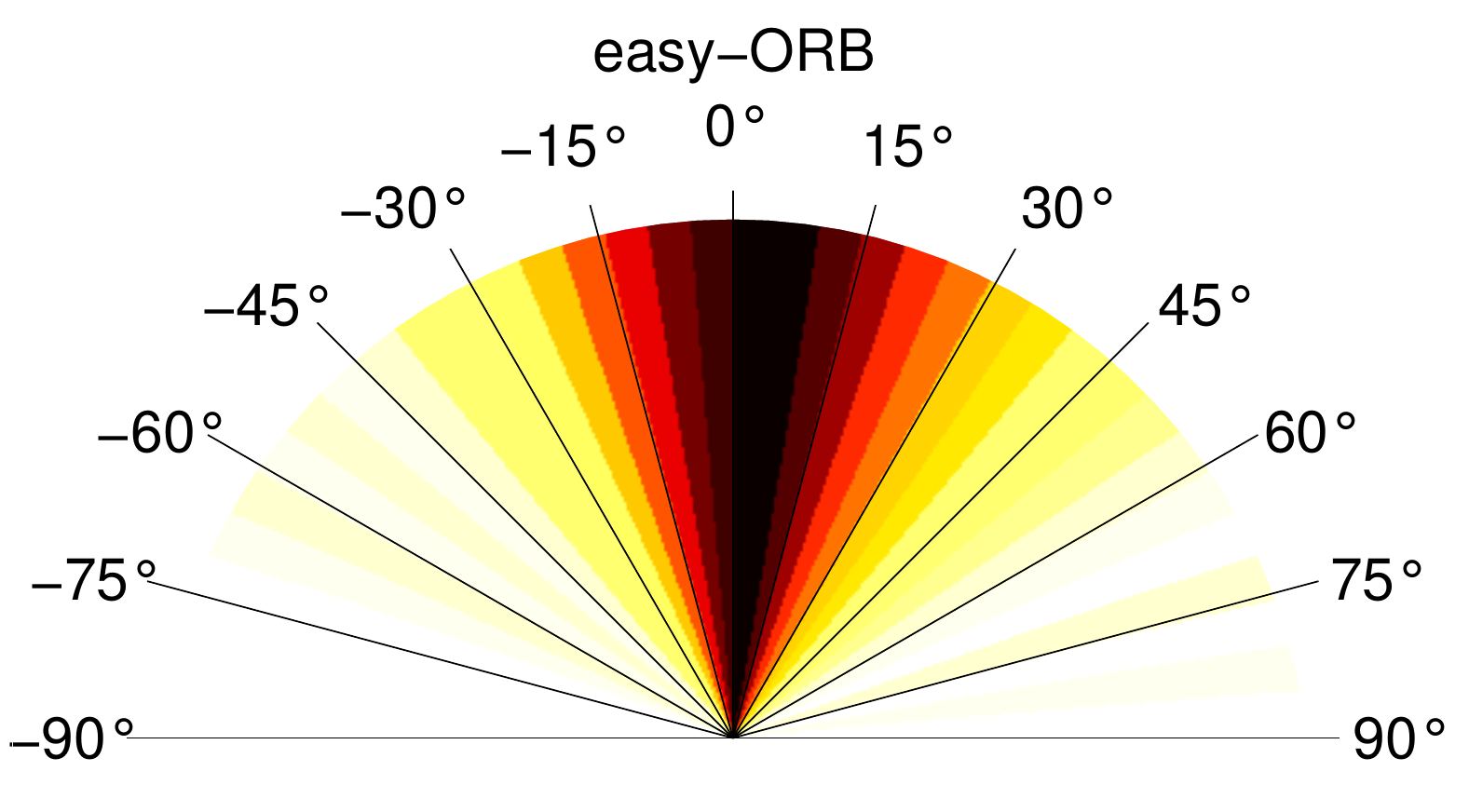}
\includegraphics[width=0.32\linewidth]{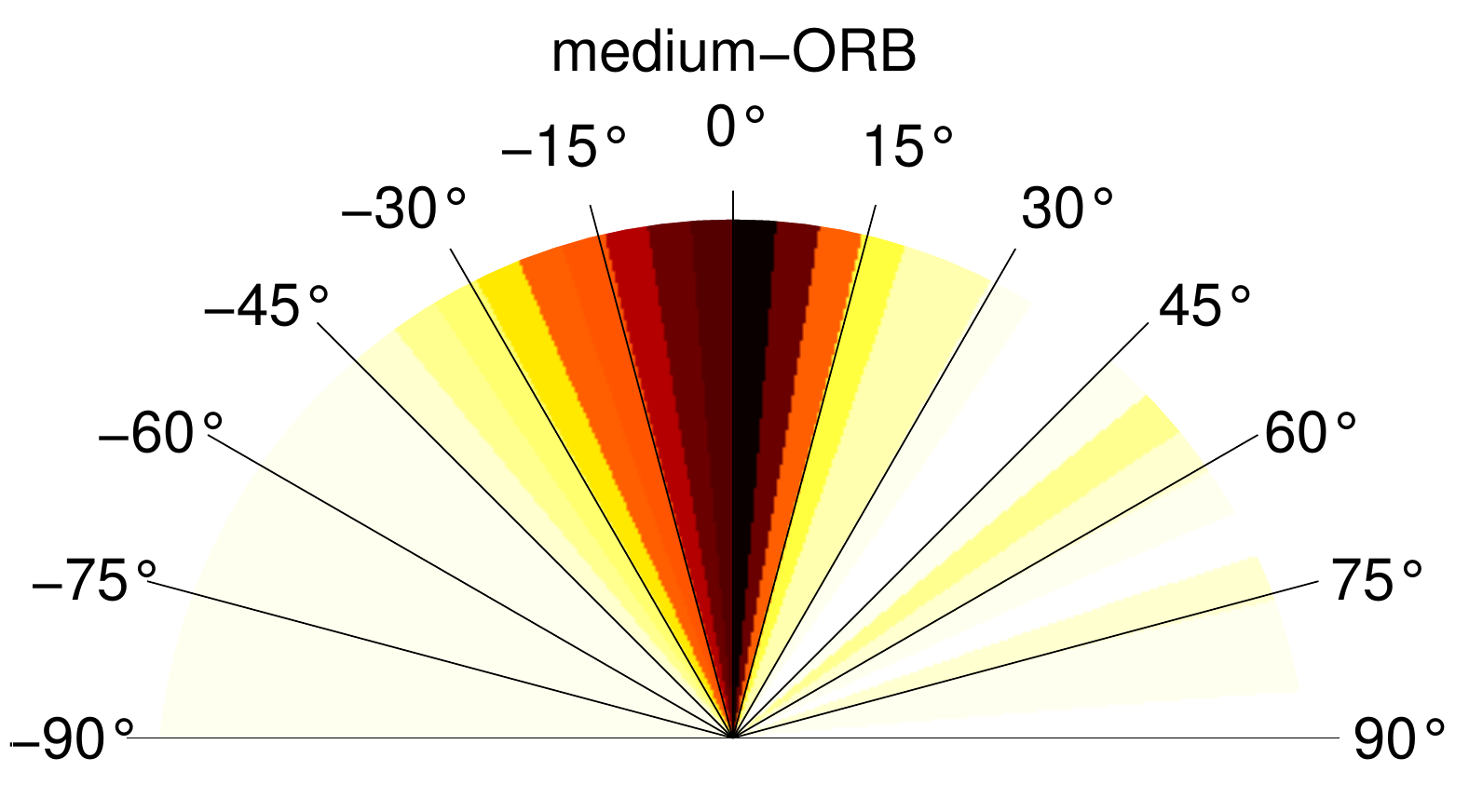}
\includegraphics[width=0.34\linewidth]{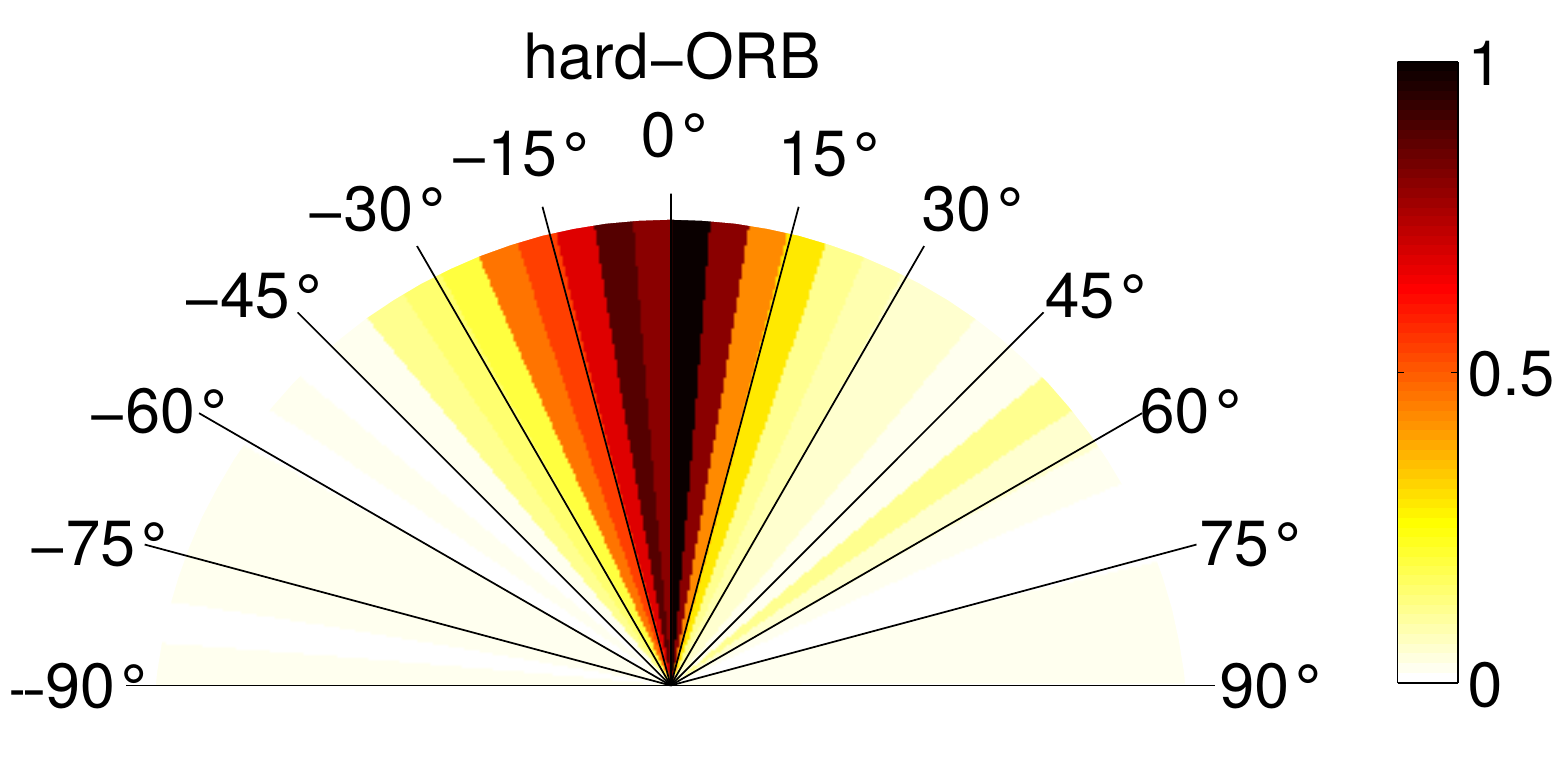}\\
\includegraphics[width=0.32\linewidth]{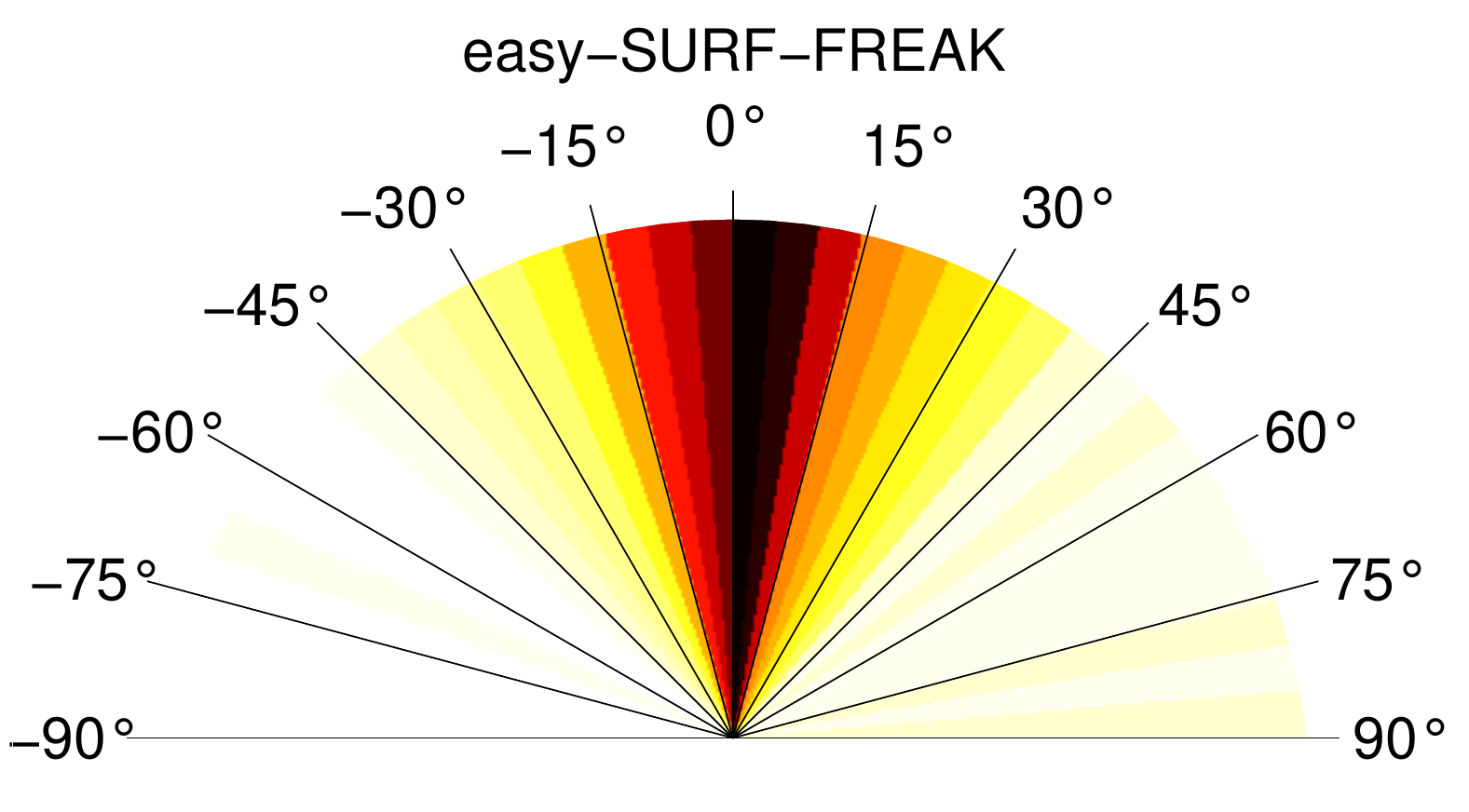}
\includegraphics[width=0.32\linewidth]{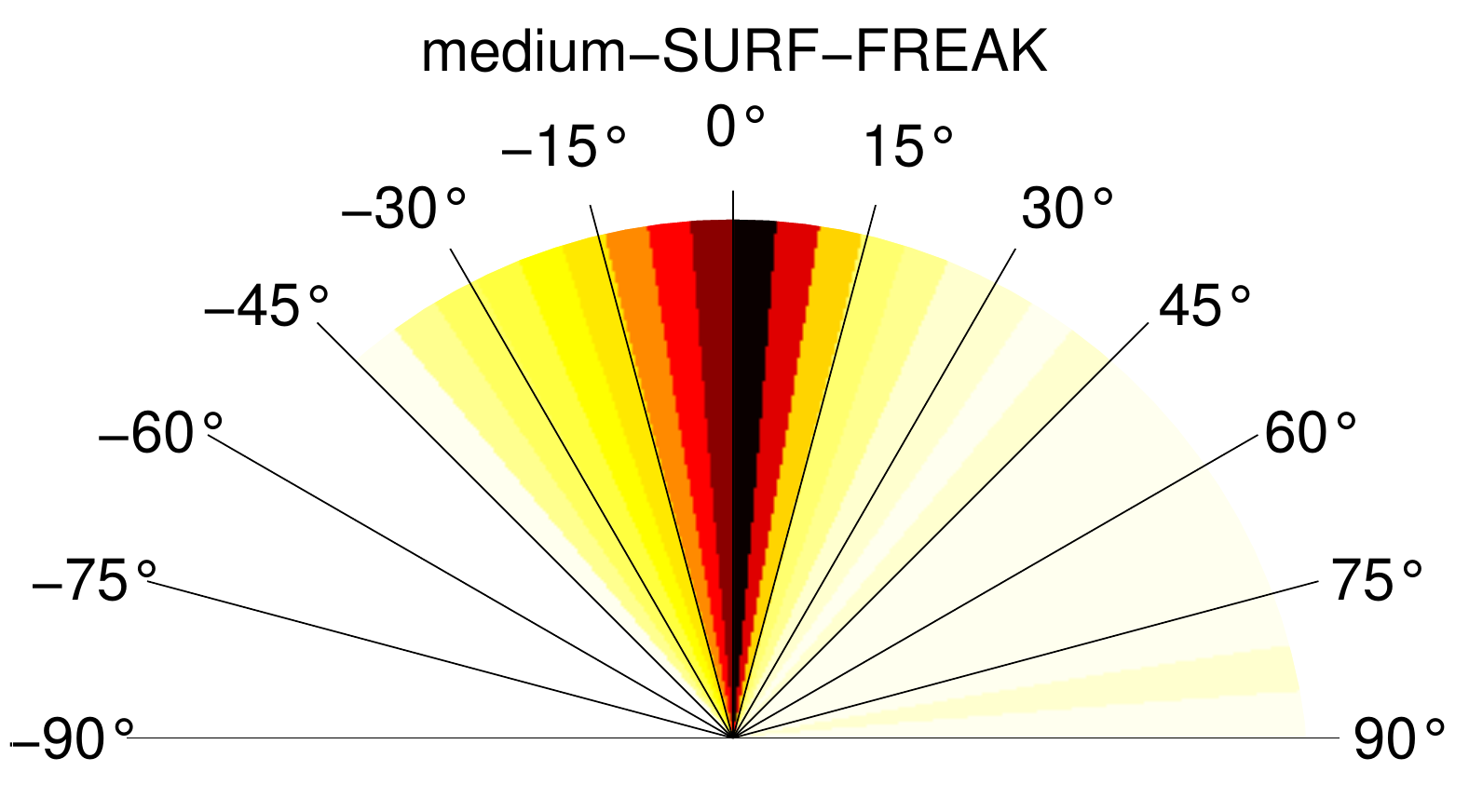}
\includegraphics[width=0.34\linewidth]{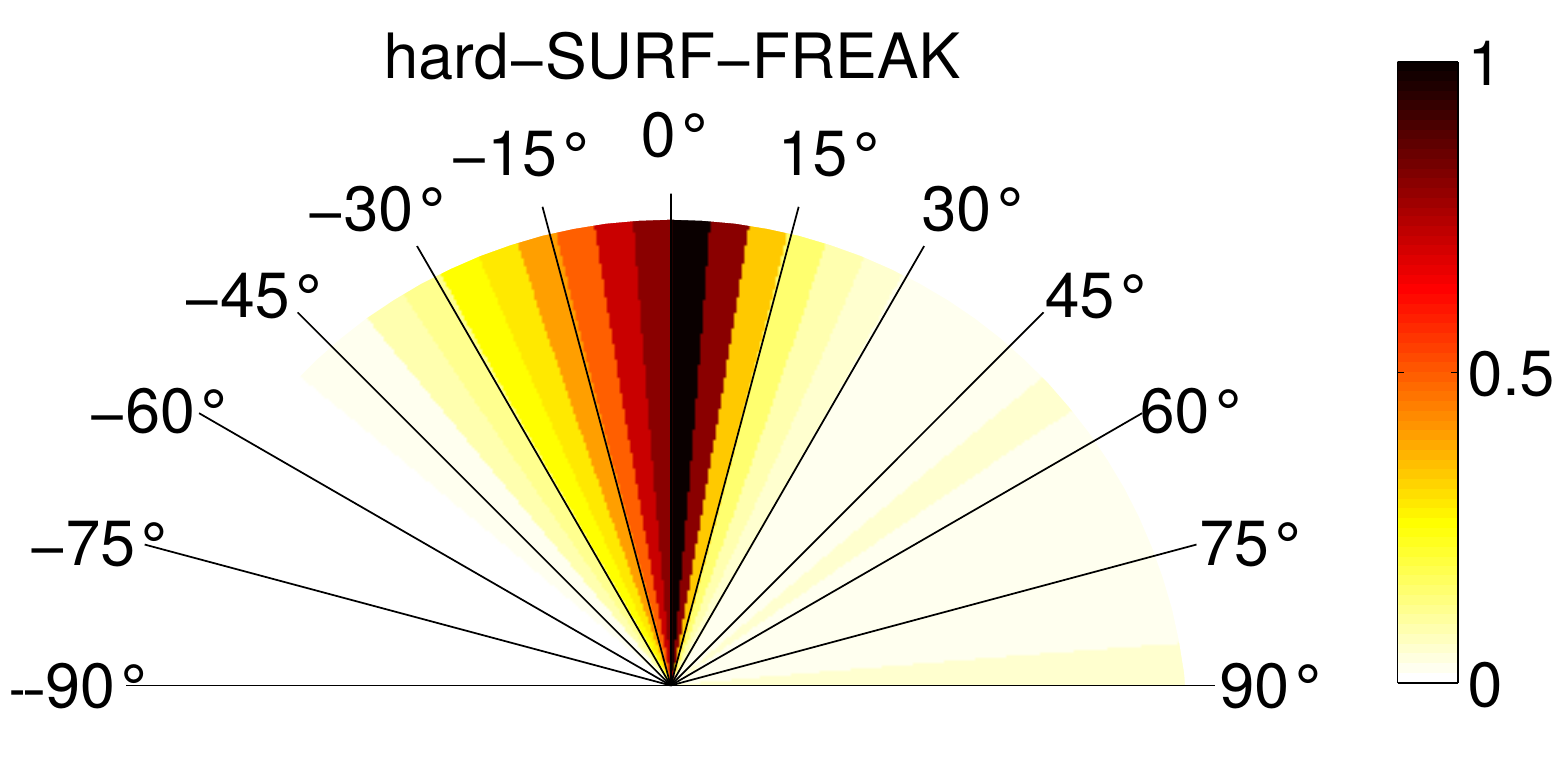}\\
\includegraphics[width=0.32\linewidth]{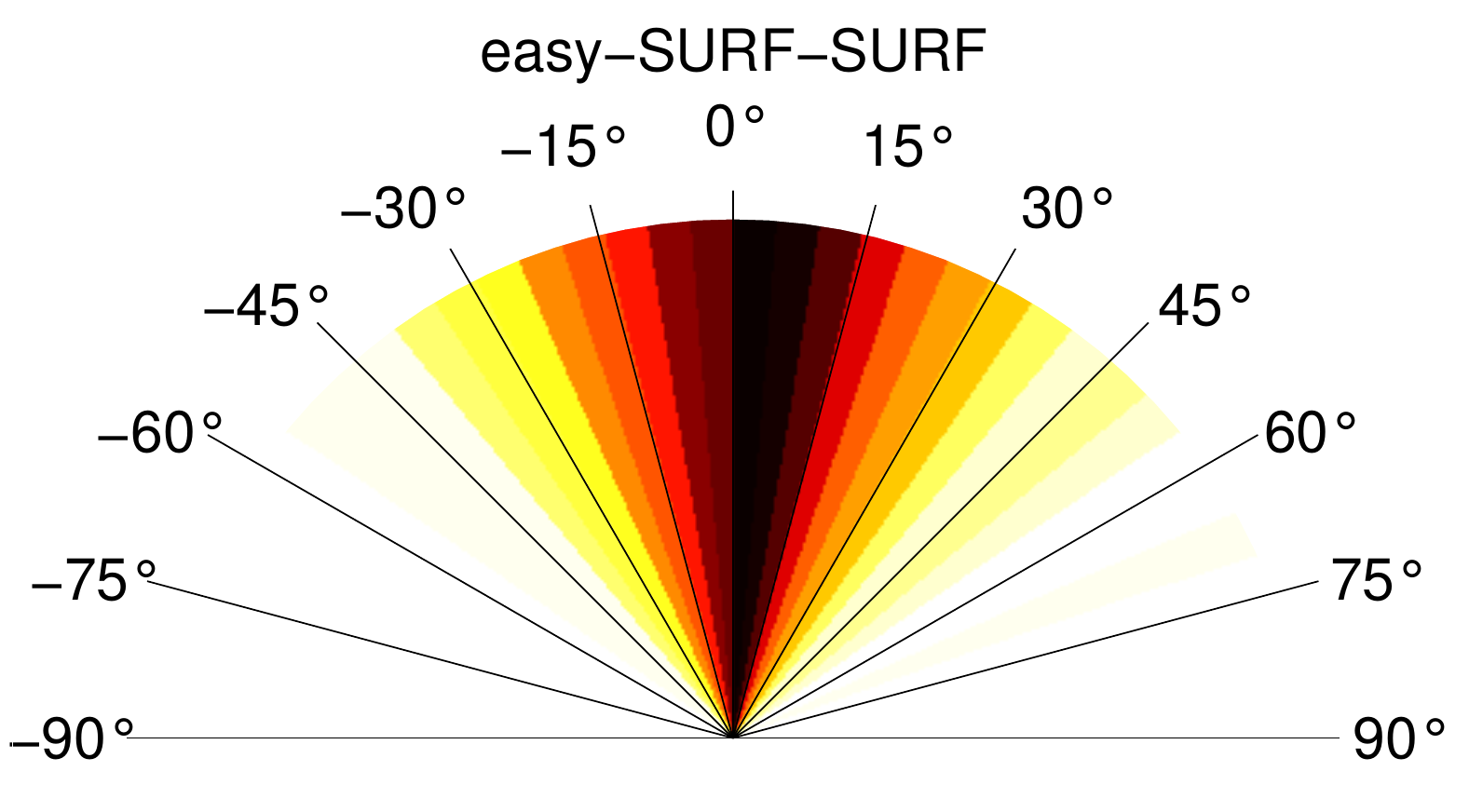}
\includegraphics[width=0.32\linewidth]{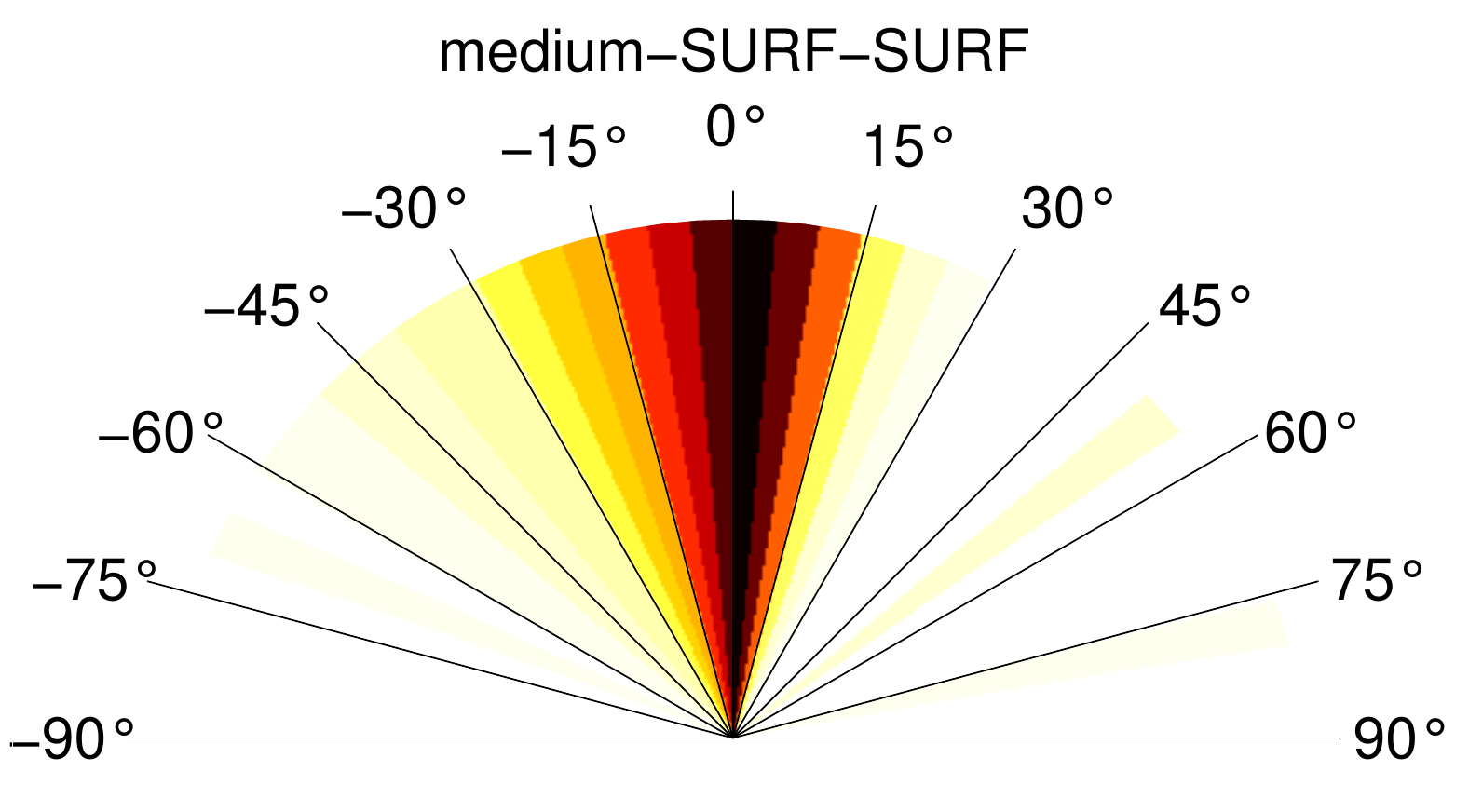}
\includegraphics[width=0.34\linewidth]{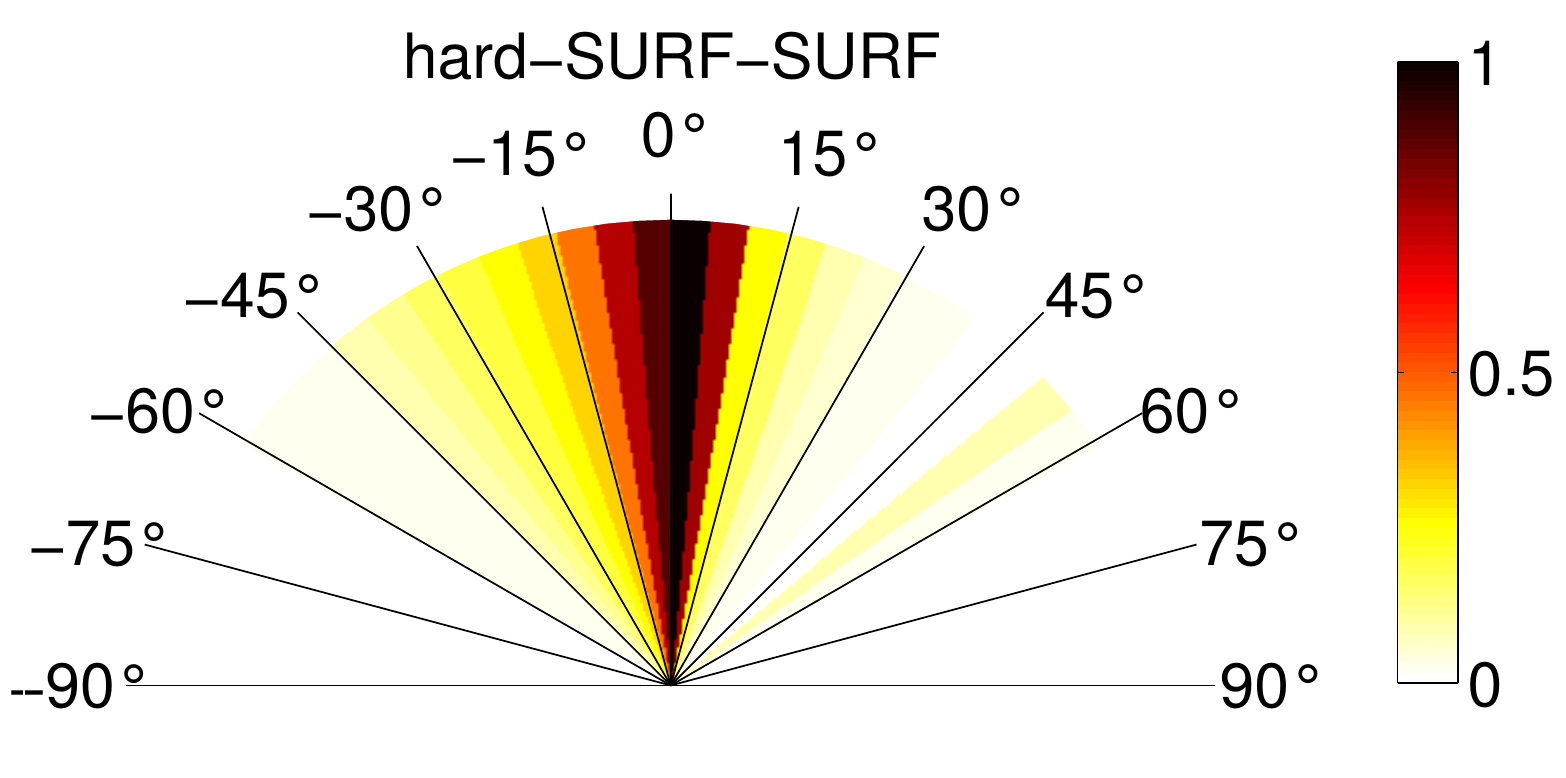}\\
\includegraphics[width=0.32\linewidth]{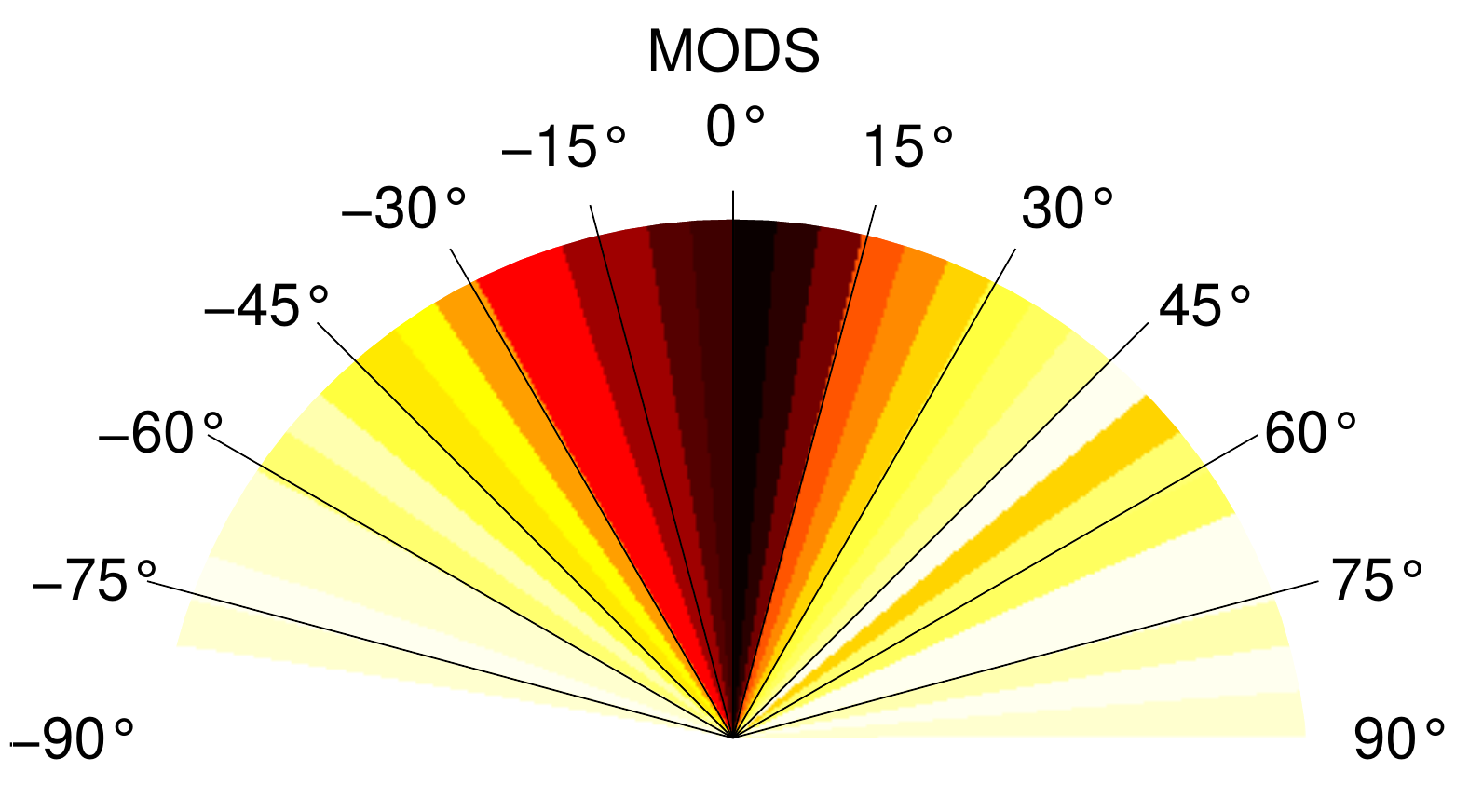}
\includegraphics[width=0.34\linewidth]{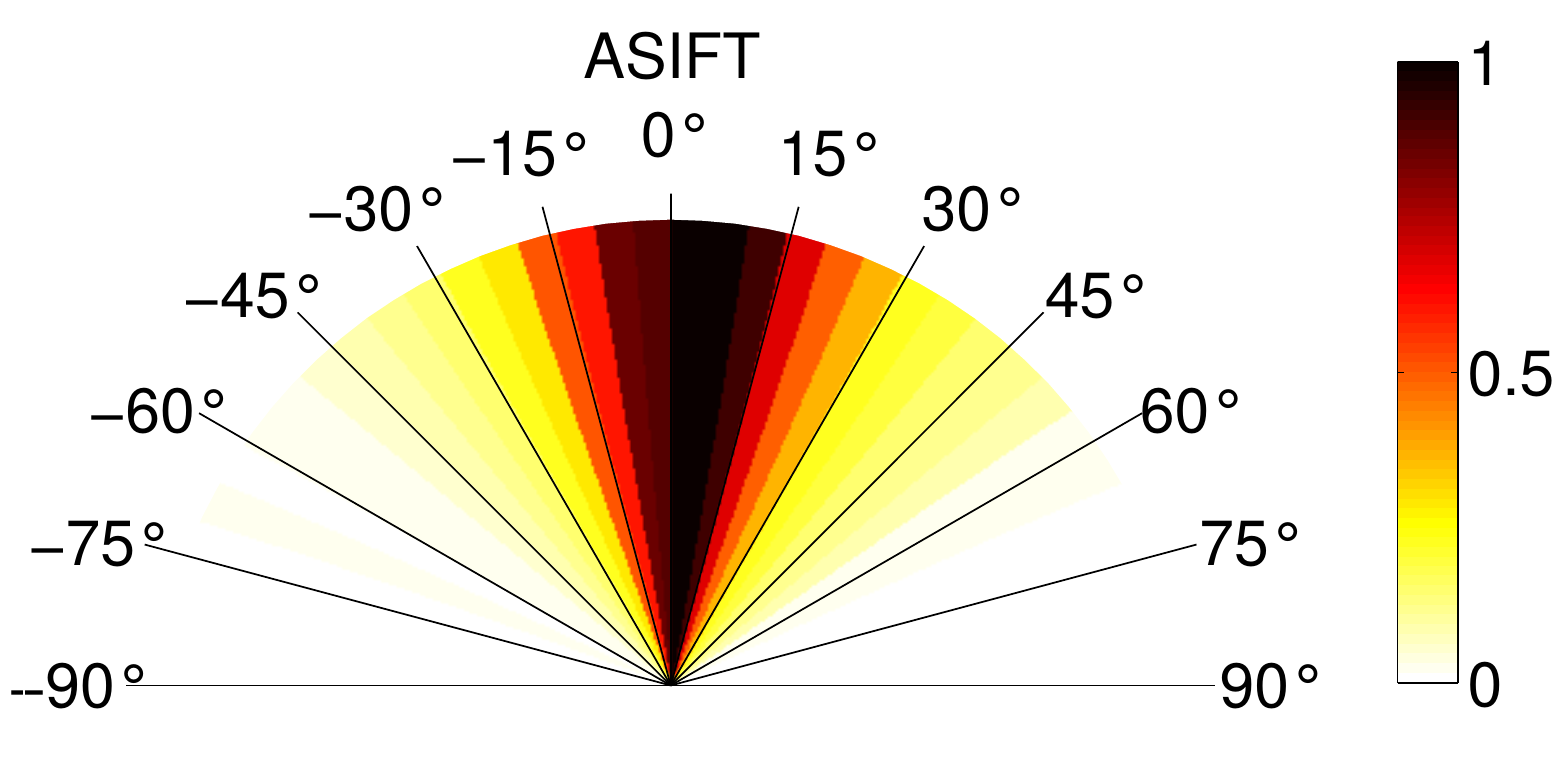}\\
\caption{A comparison of view synthesis configurations on the Turntable dataset~\cite{Moreels2007}. The fraction of correctly matched images for a given viewpoint difference.}
\label{fig:jet-angle}
\end{figure}
\begin{figure}[tbp] 
\centering
\includegraphics[width=0.32\linewidth]{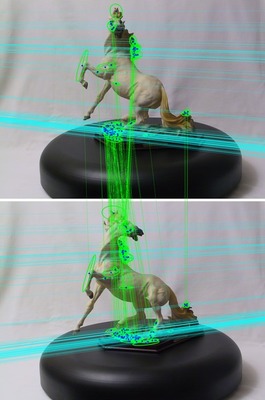}
\includegraphics[width=0.32\linewidth]{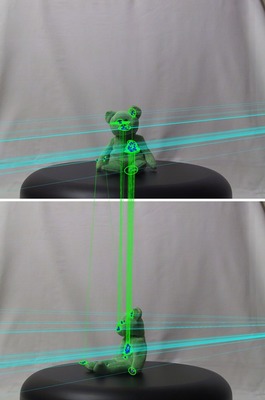}
\includegraphics[width=0.32\linewidth]{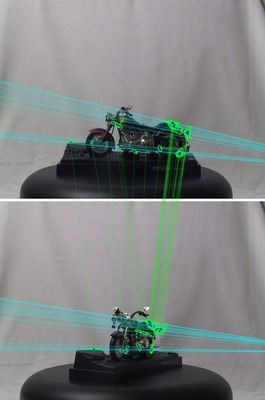}\\
\vspace{0.5ex}
\includegraphics[width=0.32\linewidth]{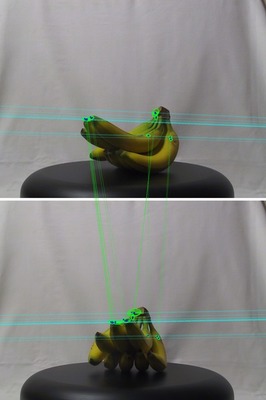}
\includegraphics[width=0.32\linewidth]{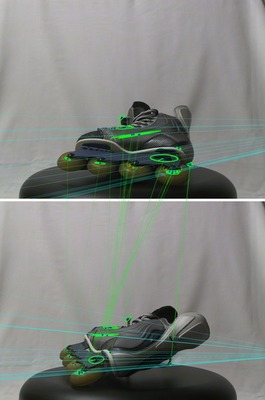}
\includegraphics[width=0.32\linewidth]{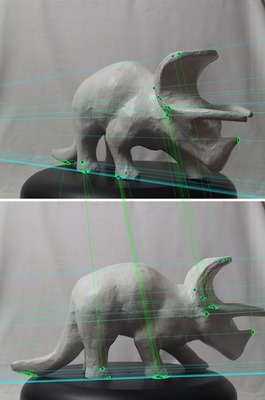}\\
\caption{Correspondences found by MODS. Green -- corresponding regions, cyan -- epipolar lines. Objects with significant self-occlusion and mostly homogenious texture were selected.}
\label{fig:mods-matches}
\end{figure} 
\begin{table}[tb]
\caption{Experiment on the 3D dataset. The configurations are defined in Table~\ref{tab:view-synth-configuration}. The number of correctly matched image pairs and the run-time per pair. Best results are \fbox{boxed}. Results within 90\% of the best are in \textbf{bold}. The configurations are sorted by average time for image pair.}
\label{tab:results}
\footnotesize
\centering
\setlength{\tabcolsep}{.35em}
\begin{tabular}{|l|rrrrrrrrrc|r|}
\hline
&\multicolumn{10}{c|}{Image sets solved (out of 35)} & \\
\hline
&\multicolumn{10}{c|}{Viewpoint angular difference} & Time\\
Matcher& $0\degree$& $5\degree$& $10\degree$& $15\degree$& $20\degree$& $25\degree$& $30\degree$&$35\degree$ &$40\degree$ &$\geq45\degree$&  [s] \\
\hline
HessianAffine easy  &\textbf{35}& \textbf{32}&\textbf{29}&\textbf{26}&\fbox{22}  &17&13&\fbox{14}&  \textbf{9}     &6&0.9\\
SURF-SURF easy      &\textbf{35}& \textbf{31}&24&18&15&12&7&5&4&2&0.9\\
HessianAffine medium&\textbf{35}& \textbf{30}&26&24&18&17&12&10&7&\textbf{10} &1.0\\
ORB easy            &\textbf{35}& \textbf{31}&\textbf{27}&20&16&11&10&5&5&4&1.0\\
AGAST-FREAK easy    &\textbf{35}&28&\textbf{29}&\textbf{26}&19&17&12&9&7&6&1.3\\
DoG easy            &\textbf{35}& \fbox{33}  &\fbox{30}  &22&18&12&10&7&5&3&1.4\\
MSER easy           &\textbf{35}&28&22&21&15&13&10&8&7&5&1.5\\
SURF-SURF hard      &\textbf{35}&26&16&11&9&7&6&4&3&3&2.1\\
HarrisAffine easy   &\textbf{35}&29&22&12&8&6&5&3&2&1&2.7\\
AGAST-FREAK hard    &\textbf{35}& \textbf{31}&\textbf{28}&\textbf{25}&\textbf{20}&17&   \fbox{15}&10&8&5&4.7\\
HarrisAffine medium &\textbf{35}&27&20&11&7&6&5&4&3&2&4.7\\
HessianAffine hard  &\textbf{35}&29&24&20&\textbf{20}&17&   \fbox{15}&11&  \textbf{9}     &6&5.6\\
DoG medium          &\textbf{35}& \textbf{32}&26&21&16&11&10&9&6&5&6.8\\
AGAST-FREAK medium  &\textbf{35}&29&23&22&18&14&13&6&6&6&7.2\\
ORB hard            &\textbf{35}& \textbf{31}&24&19&16&7&5&4&1&4&7.3\\
SURF-FREAK easy     &\textbf{35}&25&21&13&10&8&6&3&2&2&7.3\\
ORB medium          &\textbf{35}& \textbf{30}&26&18&17&10&5&4&2&4&8.3\\
SURF-FREAK medium   &\textbf{35}&22&15&10&9&7&6&4&1&2&9.1\\
SURF-SURF medium    &\textbf{35}&25&20&13&11&7&3&3&2&2&10.7\\
SURF-FREAK hard     &\textbf{35}&25&17&14&10&9&4&3&1&2&10.9\\
MSER hard           &\textbf{35}&29&24&24&19&16&12&\textbf{13}&8&8&14.5\\
HarrisAffine hard   &\textbf{35}&27&16&8&5&5&5&3&2&2&15.2\\
DoG hard            &\textbf{35}& \textbf{32}&24&17&16&13&12&8&7&8&16.7\\
MSER medium         &\textbf{35}&29&26&23&\textbf{21}&14&10&9&7&9&17.5\\
\hline
ASIFT               &\textbf{35}& \textbf{32}&        24 &      18   &     13     &8  &    7      &5&     4     &3 & 27.6\\ 
\hline
MODS:1-7                &\textbf{35}& \textbf{31}&\textbf{27}&\fbox{27} &\fbox{22} &\fbox{22} &\textbf{14}& 9&\fbox{10}&\fbox{11} & 6.7\\
\hline
\end{tabular}
\end{table}
Figure~\ref{fig:jet-angle} and Table~\ref{tab:results} show the percentage and the number of image sequences respectively for which the reference and tested views for the given viewing angle difference were matched correctly. 

The difference between {\sc easy}, {\sc medium} and {\sc hard} configurations is small for structured scenes --- unlike planar ones. Difficulties in matching are caused not by the inability to detect distorted regions but by object self-occlusions. Therefore synthesis of the additional views does not bring more correspondences.

Experiments with view synthesis confirmed \cite{Moreels2007} results that the Hessian-Affine outperforms other detectors for matching of structured scenes and can be used alone in such scenes. MODS shows similar performance, but is slower than the Hessian-Affine configuration.

The computations were performed on Intel i5 3.0GHz (4 cores) desktop with 16Gb RAM. Examples of the matched images are shown in Fig.~\ref{fig:mods-matches}.
 \subsection{MODS testing on other datasets}
\label{main:other-datasets}
\subsubsection{Extreme zoom dataset}
\label{experiments:extrazoom}
\begin{figure}[tb]
\includegraphics[height=0.33\linewidth]{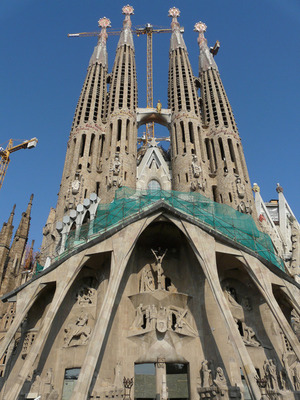}
\includegraphics[height=0.33\linewidth]{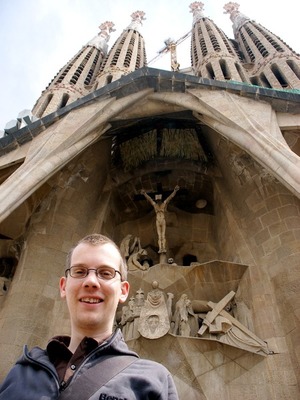}
\includegraphics[height=0.33\linewidth]{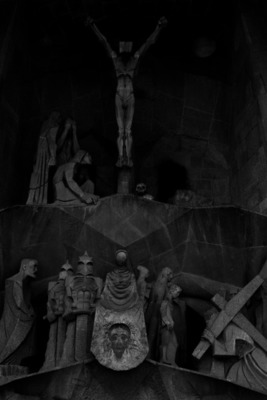}
\includegraphics[height=0.33\linewidth]{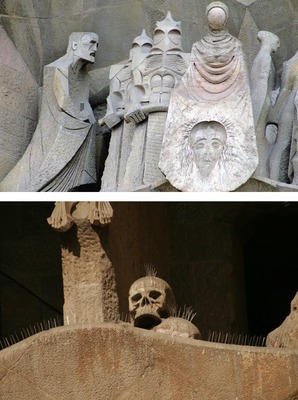}
\caption{Examples from the  Extreme Zoom Dataset}
\label{fig:extra-zoom-examples}
\centering
\end{figure}

We introduce the  Extreme Zoom dataset (EZD), which is a small subset of the retrieval dataset used in~\cite{Mikulik2014}. It consists of six sets of images with an increasing level of zoom (see examples in Figure~\ref{fig:extra-zoom-examples}).
The state-of-the art matcher -- ASIFT~\cite{Morel2009} and registration algorithm DualBootstrap~\cite{Yang2007} as well a results for MSER, ORB and Hessian-Affine matchers without view synthesis were compared to MODS. Image pairs are matched with tested algorithms. An image pair is considered solved when at least 10 output correspondences are correct.  Results are shown in Table~\ref{tab:extreme-zoom-results}.
\begin{table}[htb]
\caption{Results on the Extreme Zoom Dataset.}
\label{tab:extreme-zoom-results}
\footnotesize
\centering
\setlength{\tabcolsep}{.3em}
\begin{tabular}{lcccc}
\hline
Matcher& \multicolumn{4}{c}{Zoom level,  matched}\\
& I (max 6)&II (max 6) & III (max 6) & IV (max 4) \\ 
\hline
ORB & 0 & 0 & 0& 0\\
MSER & 2 & 2 & 1 & 0\\
DBstrap & 3 & 1 & 0 & 0\\
HessAff & 5& \textbf{3} & 2 & 0  \\
ASIFT & 3 & \textbf{3} & 2 & 0\\
\hline
MODS & \textbf{6} & \textbf{3} & \textbf{3} & 0\\
\hline
\end{tabular}
\end{table}
\subsubsection{Ultra wide-baseline dataset}
\label{experiments:ultrawide}
\begin{figure}[htbp]
\includegraphics[width=0.24\linewidth]{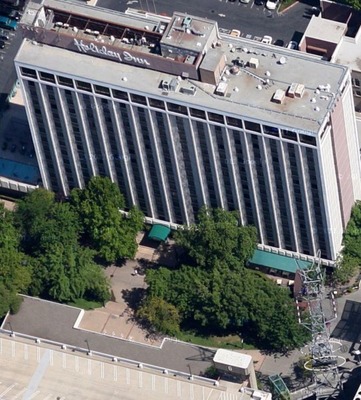}
\includegraphics[width=0.24\linewidth]{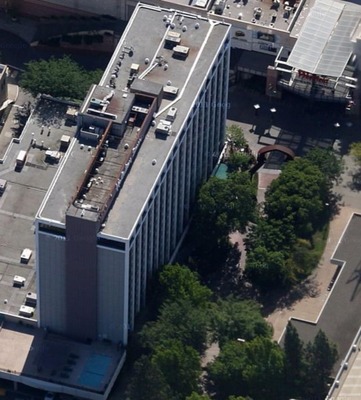}
\includegraphics[width=0.24\linewidth]{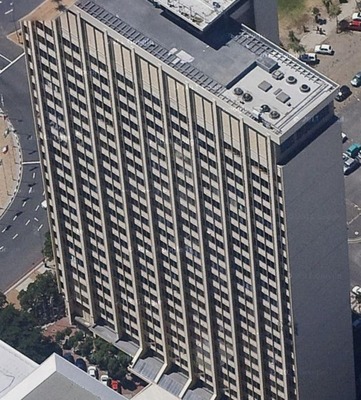}
\includegraphics[width=0.24\linewidth]{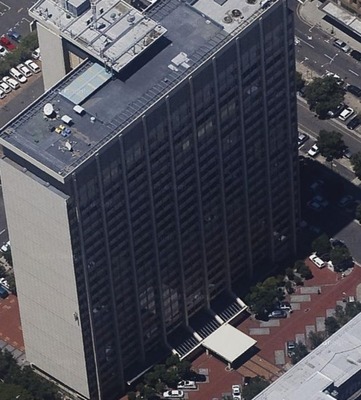}
\caption{Examples of image pairs from the Ultra wide-baseline dataset~\cite{Altwaijry2013}}
\label{fig:ultra-wide-examples}
\centering
\end{figure}

The MODS performance  was evaluated on the city-from-air dataset from \cite{Altwaijry2013}. The dataset comprises 30 pairs (examples shown in Figure~\ref{fig:ultra-wide-examples}) of photographs of buildings taken from the air. The view points difference is quite large, the images contain repeated structures, illumination differs. The authors proposed a matcher based on HoG~\cite{Dalal2005} descriptor with view synthesis and compare it to ASIFT and D-Nets~\cite{Hundelshausen2012} for skyscraper frontal face matching. We follow the evaluation protocol which considers a pair matched correct only if the facade plane is matched ($\geq 75\%$ correct inliers). If the output homography was ground/roof, it is considered incorrect. The results are shown in Table~\ref{tab:ultra-wide-results}.

Note that no special adjustment is done in MODS for homography selection, so the reported performance is a lower bound.
\begin{table}[htb]
\caption{Results on Ultra wide-baseline dataset (all results except MODS taken from ~\cite{Altwaijry2013})}
\label{tab:ultra-wide-results}
\footnotesize
\centering
\setlength{\tabcolsep}{.3em}
\begin{tabular}{llll}
\hline
Method& Correct (or shifted)& Different plane & Failure/ground plane\\
\hline
Altwaijry and Belongie~\cite{Altwaijry2013} & \textbf{9} &  1& 20 \\ 
ASIFT-homography & 1& 5 & 24 \\ 
D-Nets & 7& 2& 21 \\ 
\hline
MODS & 8& 1&21 \\ 
\hline
\end{tabular}
\end{table}
\subsubsection{Datasets with other than geometric changes}
\label{experiments:other}
Despite being  designed for (extreme) wide baseline stereo problems, MODS performance was evaluated on other datasets: GDB-ICP~\cite{Kelman2007} (modality, viewpoint and photometry changes), SymBench~\cite{Hauagge2012} (photometrical changes and photo-vs-painting pairs),  and MMS~\cite{Aguilera2012} (infrared-vs-visible pairs) -- see Table~\ref{tab:other-datasets}. The state-of-the art matcher -- ASIFT~\cite{Morel2009} and registration algorithm DualBootstrap~\cite{Yang2007} as well a results for MSER, ORB and Hessian-Affine matchers without view synthesis were compared to MODS.

\begin{table}[htb]
\caption{Evaluation Datasets}
\label{tab:other-datasets}
\footnotesize
\centering
\setlength{\tabcolsep}{.3em}
\begin{tabular}{llll}
\hline
Short name& Proposed by& \#images & Nuisance\\
\hline
GDB-ICP&Kelman~et.al.~\cite{Kelman2007}, 2007&22 pairs& Illumination, modality\\
SymBench&Hauagge and Snavely~\cite{Hauagge2012}, 2012&46 pairs&Illumination, modality\\
MMS& Aguilera~et.al.~\cite{Aguilera2012}, 2012&100 pairs&Modality\\
EVD&Mishkin~et.al.~\cite{Mishkin2013}, 2013&15 pairs&Viewpoint change\\
\end{tabular}
\end{table}
Image pairs are matched with the tested algorithms. Output keypoints correspondences were checked against the ground truth homographies. An image pair is considered  solved when at least 10 output correspondences are correct
Our primary evaluation criterion is the ability to find sufficiently correct geometric transformations in a reasonable time;  accurate geometry can be found in  consecutive step.

The computations were performed on Intel i3 3.0GHz desktop (4 cores) with 4Gb RAM.

Results are shown in the Table~\ref{tab:other-results}. MODS is the fastest method and it is able to match the most image pairs in GDB-ICP and SymBench datasets without using  symmetrical parts or other problem-specific features. Images from the MMS dataset (as well as other thermal images) produce a small number of features as they do not contain many textured surfaces, and have very short geometrical baseline. Those are the main reasons why area-based method -- Dual-Bootstrap -- work significantly better than the feature-based methods.

After lowering the threshold for detectors allowing to detect more feature points and using the orientation restricted SIFT~\cite{Chen2010} in addition to the RootSIFT, MODS-IR solved 83 out of 100 image pairs from MMS dataset. 
\begin{table}[htb]
\caption{Performance on non-WBS datasets. For comparison, results of MSER, ORB and Hessian-Affine matchers without view synthesis are added. Best results are in \textbf{bold}, average time per image pair is shown.}
\label{tab:other-results}
\footnotesize
\centering
\setlength{\tabcolsep}{.35em}
\begin{tabular}{|l|rr|rr|rr|rr|}
\hline
Matcher
& \multicolumn{2}{c|}{GDB-ICP}
& \multicolumn{2}{c|}{SymBench}
& \multicolumn{2}{c|}{MMS}
& \multicolumn{2}{c|}{EVD}\\
\hline
& \shortstack{pairs \\ solved}& \shortstack{time \\ $[s]$}
& \shortstack{pairs \\ solved}& \shortstack{time \\ $[s]$}
& \shortstack{pairs \\ solved}& \shortstack{time \\ $[s]$}
& \shortstack{pairs \\ solved}& \shortstack{time \\ $[s]$} \\
\hline
ASIFT &         15/22 &         41.5&          27/46&        14.7&          8/100&          3.2&           5/15& 12.4\\
MODS:1-7  & \textbf{17/22}& \textbf{2.8}&  \textbf{38/46}&\textbf{3.7}&        12/100&  \textbf{2.0}& \textbf{15/15}& \textbf{2.4}\\
DBstrap &       16/22 &         17.6&  \textbf{38/46}&        21.7&\textbf{79/100}&         9.3&           0/15 &1.9\\
ORB &            0/22 &         0.2&   0/46&         0.4&          0/100&          0.1&           0/15 &0.3\\
MSER &           8/22 &         1.7&          21/46 &         0.8&           0/100&         0.2&           4/15 &0.6\\
HessAff &        11/22 &        1.9&          29/46 &         1.5&          2/100&          0.4&           2/15 &1.1\\
MODS-IR:1-7  & \textbf{21/22}&      7.6&  \textbf{39/46}&\textbf{15.1}& \textbf{83/100}&  \textbf{9.1}& 14/15& 8.6\\
\hline
\end{tabular}
\end{table}
\section{Conclusions}
\label{sec:conclusions}
An algorithm for two-view matching called  Matching On Demand with view Synthesis algorithm (MODS) was introduced. 
The most important contributions of the algorithm are its ability to adjust its complexity to the problem at hand, and its robustness, i.e. the ability to solve a broader range of wide-baseline problems than the state of the art. This is achieved while being fast on simple problems. 

The apparent robustness vs. speed trade-off is finessed by the use of progressively more time-consuming feature detectors, and by on-demand generation of synthesized images that is performed until a reliable estimate of geometry is obtained. The MODS method demonstrates that the answer to the question "which detector is the best?" depends on the problem at hand, and that it is fruitful to focus on the "how to combine detectors" problem.

We are the first to propose view synthesis for two-view wide-baseline matching with affine-covariant detectors, which is superficially counter-intuitive, and we show that matching with the Hessian-Affine or MSER detectors outperforms the state-of-the-art ASIFT.
View synthesis performs well when used with simple and very fast detectors like ORB, which obtains results similar to ASIFT but in orders of magnitude shorter time.

Minor contributions include an improved method for tentative correspondence selection, applicable both with and without view synthesis and a modification of the standard first to second nearest distance rule  increases the number of correct matches by 5-20\% at no additional computational cost.

The evaluation of the MODS algorithm was carried out both on standard publicly available datasets as well as a new set of geometrically challenging wide baseline problems that we collected and will make public. The experiments show that the MODS algorithm solves matching problems beyond the state-of-the-art and yet is comparable in speed to standard wide-baseline matchers on easy problems. 
Moreover, MODS performs well on other classes of difficult two-view problems like matching of images from different modalities, with large difference of acquisition times or with  significant lighting changes.

\section*{Acknowledgements}
\label{sec:acknowledgement}
The authors were supported  by The Czech Science Foundation Project GACR P103/12/G084.

\section*{References}
\bibliographystyle{elsarticle-num}
\bibliography{main}   
\end{document}